\documentclass{article}
\usepackage{ijcai17}
\usepackage{times}

\usepackage{epsfig}
\usepackage[small]{caption}
\usepackage{graphicx}
\usepackage{amsmath}
\usepackage{amssymb}
\usepackage{color}
\usepackage{stmaryrd}
\usepackage{caption}
\usepackage{booktabs}
\usepackage{enumerate}
\usepackage{sidecap}

\def\bu{{\bf u}}

\def\bbf{{\bf f}}
\def\bh{{\bf h}}
\def\bq{{\bf q}}
\def\bx{{\bf x}}
\def\by{{\bf y}}

\begin{document}

\title{Learning deep structured network for weakly supervised change detection}
\author{Salman Khan$^{1,2}$\thanks{Corresponding author: \texttt{salman.khan@anu.edu.au}}, Xuming He$^{3}$\thanks{This work was done when the author was at Data61/ANU.}, Fatih Porikli$^{2}$, Mohammed Bennamoun$^{4}$ \\
\textbf{Ferdous Sohel$^{5}$ and Roberto Togneri$^{4}$  }  \\
$^1$Data61-CSIRO, Canberra, Australia \\
$^2$Australian National University, Canberra, Australia \\
$^3$ShanghaiTech University, Shanghai, China \\
$^4$The University of Western Australia, Perth, Australia \\
$^{5}$Murdoch University, Perth, Australia
}
\maketitle

\begin{abstract}
Conventional change detection methods require a large number of images to learn background models or depend on tedious pixel-level labeling by humans. In this paper, we present a weakly supervised approach that needs only image-level labels to simultaneously detect and localize changes in a pair of images. To this end, we employ a deep neural network with DAG topology to learn patterns of change from image-level labeled training data. On top of the initial CNN activations, we define a CRF model to incorporate the local differences and context with the dense connections between individual pixels. We apply a constrained mean-field algorithm to estimate the pixel-level labels, and use the estimated labels to update the parameters of the CNN in an iterative EM framework. This enables imposing global constraints on the observed foreground probability mass function. Our evaluations on four benchmark datasets demonstrate superior detection and localization performance. 
\end{abstract}

\section{Introduction}
Identifying changes of interest in a given set of images is a fundamental task in computer vision with numerous applications in fault detection, disaster management, crop monitoring, visual surveillance, and scene analysis in general. When there are only two images available, existing approaches mostly resort to strong supervision, thus require large amounts of training data with accurate pixel-level annotations to perform pixel-level analysis. To comprehend the significant amount of effort needed for such a formidable task, we consider the example of CDnet-2014 \cite{wang2014cdnet}, which is the largest dataset for video based change detection. 
This dataset required manual annotations for $\sim$8 billion pixel locations. Although sophisticated methods have been investigated to reduce the human effort, e.g., by expert feedback in case of ambiguity \cite{jain2013predicting,gueguen2015large}, semi-automatic propagation of annotations \cite{badrinarayanan2013semi}, and point-wise supervision \cite{russakovsky2015s}, acquisition of accurate and dense pixel-wise labels still remains {a daunting task} \cite{lin2014microsoft,song2015sun}.

\begin{figure}[t]
\centering
\includegraphics[width=0.8\columnwidth]{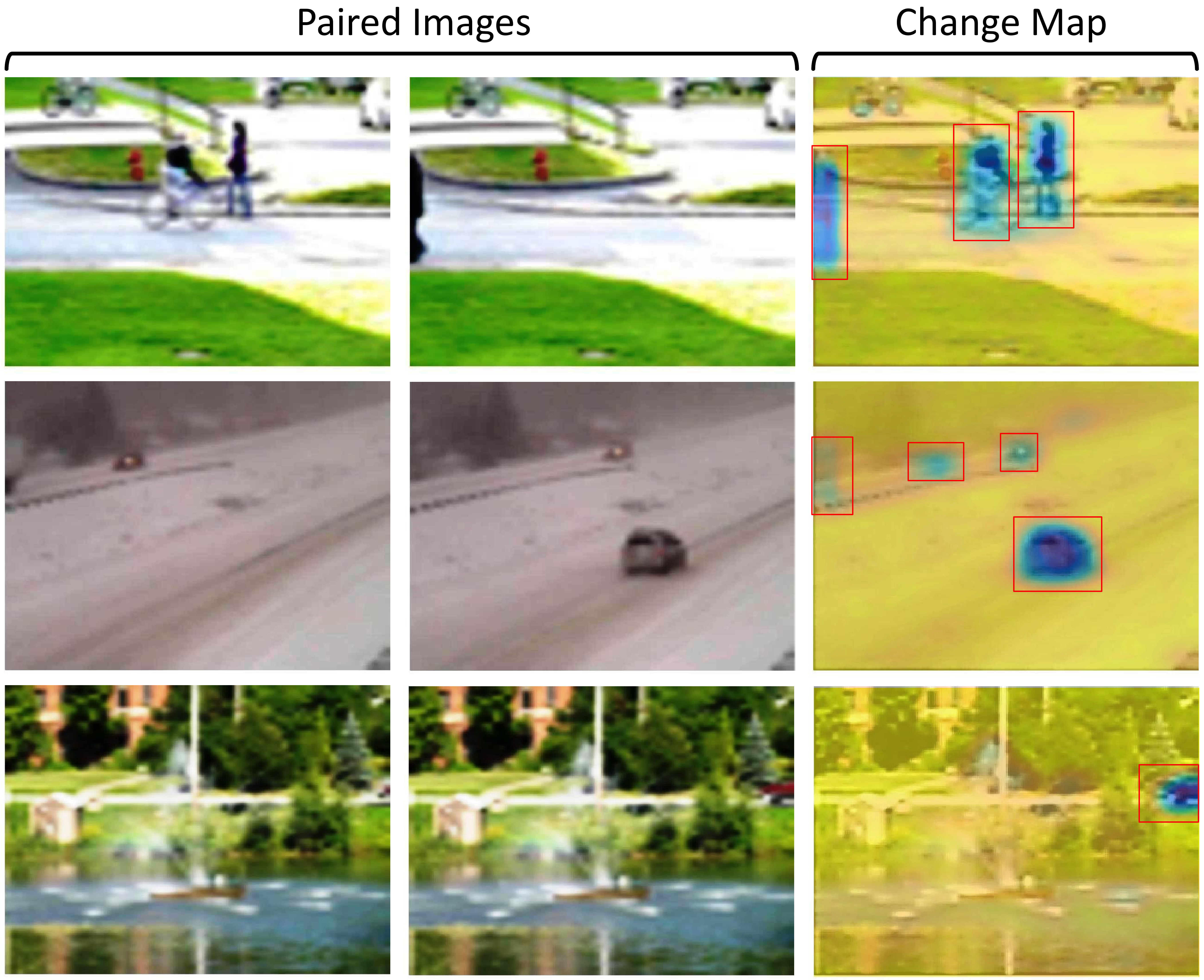}
\caption{Change Detection in a pair of images. Our approach uses only image-level labels to detect and localize changed pixels. \textbf{Left:} pair of images. \textbf{Right-most} our change localization results ({\color{blue} blue} denotes a high change region, enclosed in a {\color{red} {red}} box for clarity). Note that the paired images have rich backgrounds with different motion patterns (e.g., fountain in the \emph{third row}) and subtle changes (e.g., small vehicles in the \emph{second} row), which make the detection task very challenging. \vspace{-1.5em}}
\label{fig:IntroFig}
\end{figure}

Here, we address the problem of change detection within a pair of images and present a solution that uses only {\it image-level} labels to detect and localize changed regions (Fig.~\ref{fig:IntroFig}). Our method drastically reduces the effort required to collect annotations and provides an alternative to video change detection that requires a large number of consecutive frames to model the background scene. In many real-world applications, a continuous stream of images may not be always available due to a number of reasons such as challenging acquisition conditions, limited data storage, latency in processing, and long intervals before changes happen. For example, the analysis of aerial images for change detection, in particular for damage detection, often is formulated for a pair of images acquired at different times. Other examples where only a pair images might be available include structural defect identification, face rejuvenation tracking, and updating city street-view models.

Our algorithm jointly predicts the image-level change label and a segmentation map indicating the location of changes for a given pair of images. The central component of our method is a novel two-stream deep network model with structured outputs (Sec.~\ref{sec:model}). This model operates on a pair of images and does not need the images to be registered precisely. It can be trained with only weak image-level labels (Sec.~\ref{sec:dataNpro}). The network has a Directed Acyclic Graph (DAG) architecture where the initial layers are shared, while the {latter} part splits into two branches that make separate (but coupled) predictions for change detection and localization. In this manner, our deep network is different from the popular single-stream convolutional neural networks (CNN) for object classification \cite{khan2015cost}, detection \cite{girshick2014rich,khan2016automatic} and semantic labeling tasks \cite{papandreou15weak,long2015fully,pinheiro2015image}.

In order to jointly predict the image-level and pixel-level labels, we introduce a constrained mean-field inference algorithm (Sec.~\ref{sec:inf}), that employs a factorizable approximate posterior distribution with {global linear constraints}. Using a global constraint on the foreground (changed pixels) probability mass function, we suppress the bias towards the background (no-change labeled pixels) and encourage the assignment of change labels to nonidentical regions. Such global constraints enable us to derive an efficient mean-field {inference} procedure, while eliminating the need of approximate biases \cite{papandreou15weak} and object based priors \cite{pinheiro2015image,russakovsky2015s}. Furthermore, based on {the} novel inference algorithm, we apply a variational Expectation-Maximization (EM) learning algorithm that maximizes the lower bound of the log-likelihood of image-level labels. We extensively evaluate our approach on three publicly available datasets (CDnet-2014, PCD-2015 and AICD-2012) and a custom built satellite image dataset (GASI-2015) (Sec.~\ref{sec:dataNpro}). Our experimental results demonstrate that the proposed approach outperforms the state-of-the-art by a large margin (Sec.~\ref{sec:results}). 
The key contributions of our work include:
\begin{itemize}
\item To the best of our knowledge, this is the first work to address the weakly supervised change detection problem.
\item Our proposed CNN model jointly detects and localizes changes in image pairs.
\item We present a modified mean-field algorithm with additional constraints to efficiently localize changes.
\item We introduce a new satellite image dataset (GASI-2015) for change detection. Furthermore, we perform a rigorous evaluation on three other relevant datasets. 
\end{itemize}

\section{Two-stream CNNs for Change Localization}\label{sec:model}
We address the problem of joint change detection and localization with only image-level weak supervision. To this end, we propose a two-stream deep convolutional neural network model with structured outputs, which can be learned with weakly labeled {image pairs}. We describe our model next. 

\subsection{Model Overview}
Given a pair of input data, which can be images or (short) video clips, our goal is to predict the categories of change events in the data pair and localize the change more precisely at the pixel level. For simplicity, we focus on the image pair scenario in the following and video clips can be processed in a similar manner.   

Specifically, let each input consists of a pair of images, $\mathbf{x} = \{\mathbf{I}^1 , \mathbf{I}^2 \}$. We associate an image-level output label vector $\mathbf{y}=\{y_1,y_2\}$ to indicate the occurred change events i.e., change, no-change and $y_{1},y_{2}\in\{0,1\}$.  
It is important to note here that the no-change category (i.e., $\by=\mathbf{0}$) refers to the static-background, irrelevant changes and the dynamic background change patterns while those change categories refer to changes of interest. 
In order to localize the change events at the pixel level, we introduce a set of binary variables $\mathbf{h}$ to denote the labels of individual pixel locations for each image pair $\mathbf{x}$. Assume the image has $m$ pixel locations, $\mathbf{h} = \{h_1,\cdots, h_m\}\in \{0,1\}^{m}$. 

We formulate the change detection and localization problem as the joint prediction of its image-level and pixel-level change variables. {To achieve this, we consider a deep structured model that defines a joint probabilistic model on $\mathbf{y}$ and $\mathbf{h}$ as $P(\by, \bh|\bx; \theta) = \frac{1}{Z(\bx)}\exp(-E(\by,\bh|\bx; \theta))$, where the Gibbs energy is defined as:
\begin{align} \label{eq:model}
 \resizebox{0.9\hsize}{!}{$
E(\by, \mathbf{h}|\mathbf{x}; \theta)  = \Phi_l(\by|\mathbf{x};\theta_l)+ \Phi_u(\mathbf{h}|\mathbf{x}; \theta_u)+ \Phi_p(\mathbf{h}, \by|\mathbf{x},\theta_p), $}
\end{align} 
where $\Phi_l(\by|\mathbf{x};\theta_l)$ is the unary term for image-level label $\by$, modeled by a CNN with parameter $\theta_l$, and $\Phi_u(\mathbf{h}|\mathbf{x}; \theta_u)$ is the unary term for pixel-level labels, modeled by a Fully Convolutional Network (FCN) with parameter $\theta_u$. The pairwise energy $\Phi_p(\mathbf{h}, \by)$ consists of two terms, $\psi_p$ and $\psi_u$, which enforces the spatial smoothness of pixel-level labels and captures the coupling between image- and pixel-level labels, respectively. The joint prediction can be formulated as inferring the MAP estimation of the model distribution, 
\begin{align} \label{eq:map}
{\by}^*, {\mathbf{h}}^* & = \arg\max_{\by,\mathbf{h}} P(\by, \mathbf{h}|\mathbf{x}; \theta).
\end{align}}
A graphical illustration of the model is shown in Fig.~\ref{fig:model}. 

\subsection{Deep Network Architecture}

\begin{figure*}
	\centering
		\begin{minipage}{0.245\textwidth}
	\vspace{2em}
	\includegraphics[width=1.15\columnwidth]{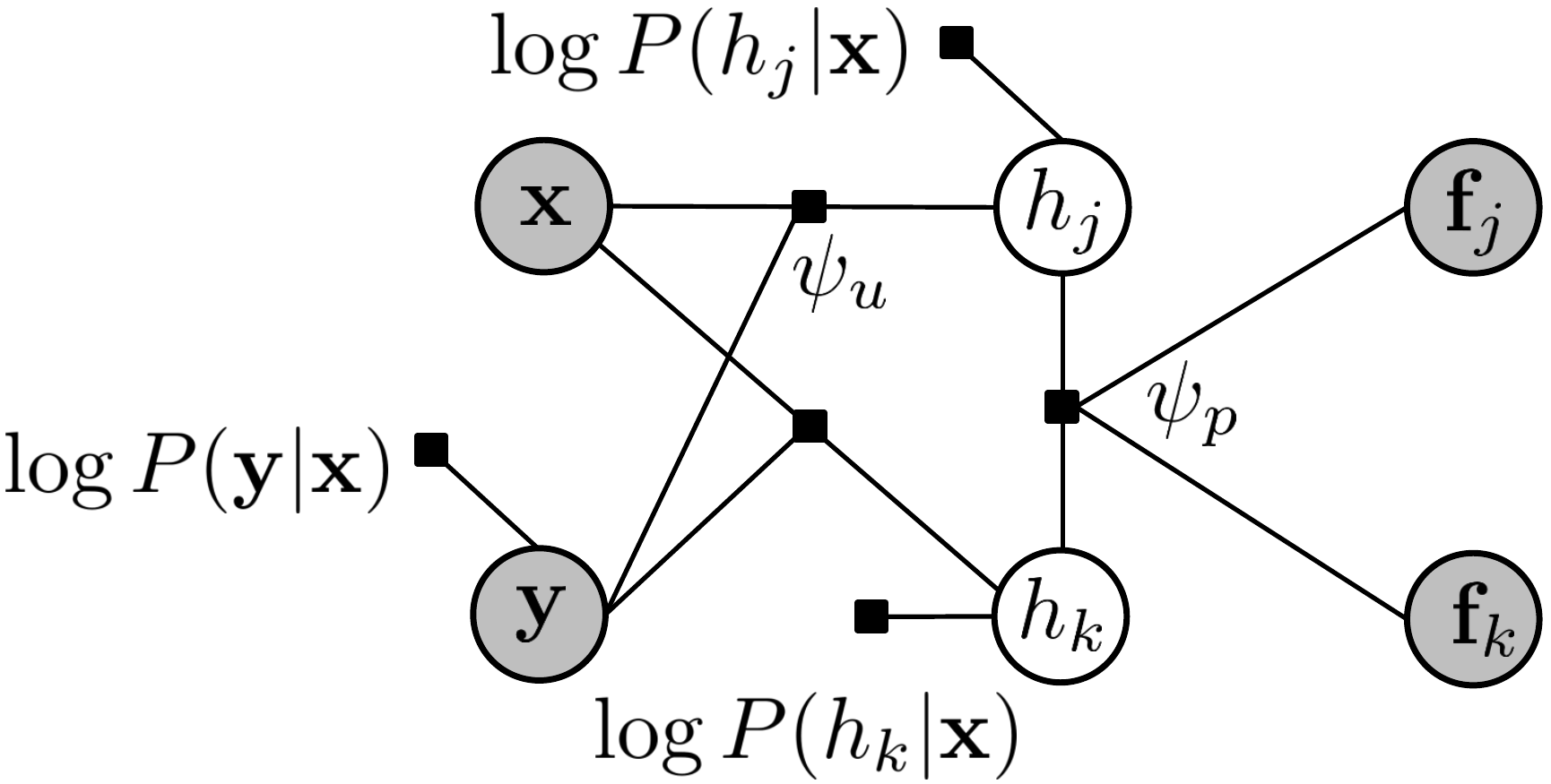}
	\caption{\textbf{Factor graph representation} of the weakly supervised change detection model. The shaded and non-shaded nodes represent the observable and hidden variables, respectively. The dense connections between hidden nodes are not shown here for clarity.  }
	\label{fig:model}
\end{minipage}
\hfill
    \begin{minipage}{0.70\textwidth}
	\includegraphics[width=1\textwidth]{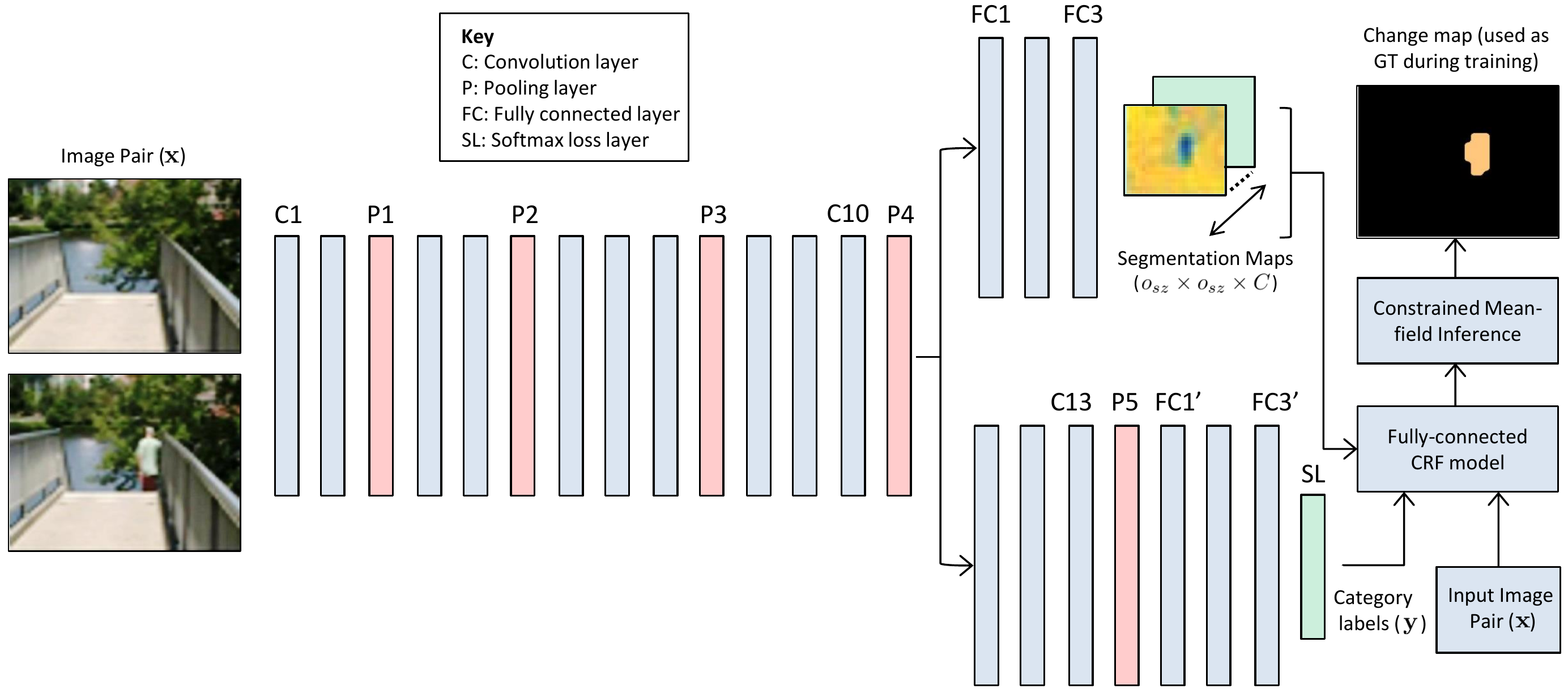}
	\caption{\textbf{CNN architecture:} The network operates on paired images and divides in two branches after the fourth pooling layer ($P4$). The classification branch (lower) is trained using the image-level labels ($y$). The hidden variables $h_j$ are estimated using the constrained mean-field algorithm, which are iteratively used to update the CNN parameters. }
	\label{fig:CNNarch}
	\end{minipage}
	\vspace{-0.3em}
\end{figure*}

{We build the deep structured model by first introducing a two-stream deep CNN for the unary terms as shown in Fig.~\ref{fig:CNNarch}.} The underlying architecture of the network is similar to the VGG-net (configuration-D, the winner of the classification and localization challenge, ILSVRC'14) \cite{simonyan2014very} but with several major differences. Most importantly, the network operates on multichannel inputs (6 channels for paired color images) and divides into two branches after the fourth pooling layer ($P4$). From our initial experiments (consistent with \cite{zagoruyko2015learning}), a multi-channel network performs better than a traditionally used Siamese network for paired images. The two branches compute the probability of the image-level and pixel-level labels, and therefore will be called as the classification and the segmentation branch, respectively. The segmentation branch in our architecture is similar to FCN-VGG16-16s network \cite{long2015fully} which demonstrated state-of-the-art performance on the Pascal VOC segmentation dataset. The initial shared layers in our architecture combine the initial (essentially similar) portions of VGG and FCN networks, which results in a significant decrease in trainable parameters without any drop in performance. We now describe the details of the two branches of the network architecture. 

\vspace{-0.5em}
\paragraph{\bf Image-level change unary energy:} 
The classification branch predicts the image-level label probability and has more layers to collapse the filter responses from the initial layers. Specifically, the classification branch output of the CNN architecture models $\Phi(y|\mathbf{x};\theta_l)$, predicting the image-level change energy as: 
\begin{align}\label{eq:lower}
\Phi_l(y|\mathbf{x};\theta_l) = -F_{l-cnn}(\mathbf{x};W_s,W_y),
\end{align}
where $F_{l-cnn}$ is the deep network feature before the final softmax operator, $W_s$ are the weight parameters shared with the segmentation branch, and $W_y$ are the weight parameters for the classification branch only.
 
\vspace{-0.5em}
\paragraph{\bf Pixel-level change unary energy:} 
The segmentation branch generates a down-sampled coarse segmentation map (of size $o_{sz} \times o_{sz}$) for each change category. After shared layers, the branch has three fully connected layers, which are implemented as convolution layers as in the FCN~\cite{long2015fully}.  
Formally, the segmentation branch of the CNN model generates the pixel-level change label energy as follows, 
\begin{align*}
 \resizebox{1.02\hsize}{!}{$
\Phi_u(\mathbf{h}|\mathbf{x};\theta_u) = \sum\limits_{j=1}^{m} \Phi({h}_{j}|\mathbf{x}),
\Phi({h}_{j}|\mathbf{x}) = -F_{u-cnn}(h_j, \mathbf{x}; W_s, W_f),$
} 
\end{align*}
where $F_{u-cnn}$ denotes the segmentation branch scores of the CNN architecture before the soft-max operator and $W_f$ are the weights for the fully connected layers.  

{We now describe the pairwise energy  
$\Phi(\mathbf{h}, y|\bx,\theta_p)$ that encodes the compatibility relations between the image-level and the pixel-level variables as well as the spatial smoothness. Specifically,} 
on the top of the fully connected layers, we add a densely connected Conditional Random Field (CRF) to impose the spatial smoothness of the pixel labeling. Unlike the previous models~\cite{papandreou15weak}, our dense CRF depends on the output label of the classification branch, and thus couples the image-level and pixel-level prediction.

Formally, we define the compatibility relations between the output variables $y$ and $\bh$ by the following energy functions, 
\begin{align}\label{eq:compat}
 \resizebox{0.9\hsize}{!}{$
\Phi_p(\mathbf{h}, y|\mathbf{x},\theta_p) = \sum\limits_{j}\psi_u(h_{j}, y,\bx_j) + \sum\limits_{j<k} \psi_p (h_{j}, h_{k},\mathbf{f}_{j},\bbf_k), $}
\end{align}
where $\psi_u$ enforces all hidden variables to be zero if the category label predicts no-change and encourages $h_j$ to take a change label otherwise: 
\begin{align*}
 \resizebox{1.0\hsize}{!}{$
\psi_u(h_{j}, y,\bx_j) = \llbracket y = 0 \rrbracket\llbracket h_j = 0 \rrbracket
+\llbracket y = 1 \rrbracket\big(1+e^{-\gamma\delta(\bx_{j})}\llbracket h_j=0\rrbracket\big), $}
\end{align*}
where $\gamma$ is a weight parameter and $\delta(\bx_{j})$ is the color difference between two images at pixel $j$. 
 The fully-connected pairwise term $\psi_p$ defines the smoothing term between the latent variables $\mathbf{h}$ given input features $\bbf_{j},\bbf_k$. These energies have a functional form of the weighted Potts model in which the weight is defined using Gaussian kernels of \cite{krahenbuhl2011efficient}:  
\begin{align}\label{eq:pairwise}
\psi_p(h_{j}, h_{k},\bbf_{j}, \bbf_k) =(\alpha_{ap}k_{ap}+\alpha_{sm}k_{sm}) \mu(h_{j}, h_{k}),
\end{align}
where $\alpha_{ap}$, $\alpha_{sm}$ are the kernel weights, $\mu(h_j,h_k)$ is the Potts compatibility while $k_{ap}(\bbf_j,\bbf_k)$, $k_{sm}(\bbf_j,\bbf_k)$ are the appearance and smoothness kernels \cite{krahenbuhl2011efficient}.

\subsection{Model Inference for Change Localization} \label{sec:inf}
Given the two-stream CNN+CRF model, we predict the image and pixel-level change labels by inferring the MAP estimation of the joint probability model in Eq.~\eqref{eq:model}. In order to compute the most likely configuration efficiently, we adopt a sequential prediction approach that first infers the image-level change label followed by the pixel-level change mask inference. 
Specifically, we compute the change label prediction approximately as follows, 
\begin{align*}
 \resizebox{1.0\hsize}{!}{$
y^* = \arg\min\limits_{y} \Phi_l(y|\mathbf{x};\theta_l),
\mathbf{h}^* = \arg\min\limits_{\mathbf{h}}\Phi_u(\mathbf{h}|\mathbf{x}; \theta_u) + \Phi_p(\bh,y^*|\bx; \theta_p).  $}
\end{align*}
This prioritized inference procedure allows us to compute the (more reliable) image-level label first  and to run an efficient mean-field inference for the pixel-level labels only once\footnote{{In general, we note that we can compute the MAP estimation jointly by enumerating $y$'s values and running mean-field inference multiple times, which is less efficient.}}.     

We now derive a constrained mean-field inference algorithm for inferring the pixel-level change labeling $\bh$. We note that the efficient mean-field algorithm~\cite{krahenbuhl2011efficient} usually leads to an over-smoothing of the pixel-level labeling and assigns most of the pixels to the `no-change' class. In this work, we incorporate an additional global constraint on the proportion of `change' label values in the image. Unlike previous methods (e.g.,~\cite{papandreou15weak,pinheiro2015image,russakovsky2015s}), we enforce such constraints on the approximate probability family which allows us to derive an efficient modified mean-field procedure.  

Formally, we assume the foreground label proportion to be $\tau$, which is fixed during training by cross-validation. For each test image pair, we find $K$ closely matching pairs  from the training set using a KNN search and average their foreground label proportion to estimate $\tau$ (details in Sec.~\ref{sec:results}). To enforce the proportion constraint, we introduce the following factorized approximate probability family with a global constraint:
\begin{align}\label{eq:q_func}
\resizebox{0.75\hsize}{!}{$
Q(\bh|\bx,y^*) = \prod_j \bq_j(h_j), \, \text{and} \, \sum_j [0,1]\bq_j=\tau, $}
\end{align} 
where, $\bq_j = [\bq_j(0),\bq_j(1)]^T$ and the constraint implies that the overall foreground probability mass $\sum_j\bq_j(1)$ is $\tau$. Following~\cite{krahenbuhl2013parameter}, we {minimize} the approximate KL-divergence, 
\begin{align}
\resizebox{0.1\hsize}{!}{$ D_{Q||P} $}= & \resizebox{0.85\hsize}{!}{$ \sum_j\big(\bq_j^T\log \bq_j + \bq_j^T\bu_j\big) + \frac{1}{2}\sum_{j,k}\bq_j^T\Psi_{jk}\bq_k+C, $}  \nonumber \allowdisplaybreaks \\
\text{with}\quad & \resizebox{0.48\hsize}{!}{$\mathbf{1}^T\bq_j = 1, \, \sum_j [0,1]\bq_j=\tau, $}
\end{align}  
where $\bu_j$ is the unary term vector ({including $P(h_j|\bx)$ and $\psi_u(h_j,y^*,\bx_j)$}) and $\Psi_{jk}$ is the compatibility matrix {computed from $\psi_p$}, and $C=\log Z$ is the log partition function. {We use the CCCP algorithm~\cite{yuille2003concave} to minimize $D_{Q||P}$ iteratively.}  

\vspace{-0.2em}
\section{EM Learning with Weak Supervision}\label{sec:learn}
We now consider a weakly supervised learning approach to estimate the parameters of the two-stream CNN+CRF model (Sec.~\ref{sec:model}). In particular, as the labeling of the pixel-level change pattern is tedious and impractical, we assume only image-level change annotations are available, which can be obtained with much less effort. Let us denote the dataset $\mathcal{D}$ comprising of $N$ labeled image pairs: $\mathcal{D} = \{\mathbf{x}^n,y^n\}^{1\times N}$. 

The learning objective is to maximize the log conditional likelihood and we consider a variational mean-field energy lower bound as follows, 
\begin{align} 
& \resizebox{1.05\hsize}{!}{$
 \sum_n\log P(y^n|\mathbf{x}^n;\theta) \geq \sum_n\sum_{\mathbf{h}^n}Q(\mathbf{h}^n|y^n,\mathbf{x}^n)\log\frac{P(y^n,\mathbf{h}^n|\mathbf{x}^n;\theta)}{Q(\mathbf{h}^n|y^n,\mathbf{x}^n)} $} \notag \allowdisplaybreaks\\
& \resizebox{0.75\hsize}{!}{$
=E_Q[\log P(y^n,\mathbf{h}^n|\mathbf{x}^n;\theta)] + H(Q(\mathbf{h}^n|\mathbf{x}^n, y^n)), $} \notag
\end{align} 
where, $E_Q[\cdot]$ and $H(\cdot)$ denote the expected value and the entropy function respectively,  and $Q(\mathbf{h}^n|\mathbf{x}^n,y^n)$ is an approximate posterior probability factorizing over $\{h^n_j\}$ as defined in Eq.~\eqref{eq:q_func}. 
In other words, the posterior probability can be expressed as the product of independent marginals: $Q(\mathbf{h}^n|\mathbf{x}^n,y^n) = \prod_jq^n_j(h^n_j)$. 
We then derive a variational expectation-maximization (EM) algorithm for learning our two-stream CNN+CRF in the following, which alternately maximizes the objective function above.  

\vspace{-0.2em}
\subsection{Mean-field E Step} 

We update the approximate $Q$ function by maximizing the objective w.r.t the $Q$ function given the model parameter $\theta$ from the previous iteration. Note that given the model structure, this leads to a mean-field updating equation to compute $q(h_j^n)$. 
The updating equation requires message passing between all the $h_j$ and $h_k$, which is computationally expensive. 
Efficient message passing is achieved using the high dimensional Gaussian filtering by considering the permutohedral lattice structure \cite{adams2010fast}.

Given the approximate posterior marginals, we can compute the (approximate) most likely configuration of the latent variables $\bh^n$, 
\begin{align}
\resizebox{0.55\hsize}{!}{$
h_j^{n*} \leftarrow  \underset{h_j^n}{\operatorname{argmax}} \; \prod\limits_{j = 1}^{m}  q(h^n_j|\mathbf{x}^n, y^n). $}
\end{align}
The marginal mode $\bh^{n*}$ will be used in the M step for the CNN+CRF learning. 

\vspace{-0.2em}
\subsection{M Step for CNN+CRF Training} 
Once we have the posterior marginal distribution $q(h_j^n)$ and its mode, we update the model parameters $\theta$ with the posterior mode configuration $\{\bh^{n*}\}$ and ground-truth $\{y^n\}$. Specifically, we treat them as the ground-truth for the pixel and image-level labels, and learn the two-stream deep CNN+CRF in a stage-wise manner. {Our stage-wise learning first estimates the parameters in the unary terms, i.e., the two deep CNNs, and then validates the parameters in the pairwise term. This strategy is similar to the piece-wise learning in the CRF literature.}

We first use back-propagation to train the two branches of the deep CNN separately with the corresponding training data. 
More precisely, the averaged gradient from two streams is back-propagated to update the shared parameters ($W_s$), while the individual gradients are computed using $y^{n}$ and $\mathbf{h}^{n*}$ as ground-truths to update $W_y$ and $W_f$ for the classification and segmentation branches, respectively. 
Concretely, the model parameters are updated to maximize the data likelihood as follows, 
\begin{align}
& \resizebox{0.95\hsize}{!}{$
W_s^* \leftarrow \underset{W_s}{\operatorname{argmax}}\; \sum_n \Big(\log P(y^n|\bx^n; \theta_l) + \log P(\bh^{n*}|\bx^n;\theta_u)\Big), $}\notag \\
& \resizebox{0.6\hsize}{!}{$
W_y^* \leftarrow \underset{W_y}{\operatorname{argmax}} \; \sum_n \log P(y^n|\bx^n; \theta_l), $} \notag \\
& \resizebox{0.6\hsize}{!}{$
W_f^* \leftarrow \underset{W_f}{\operatorname{argmax}} \; \sum_{n} \log P(\bh^{n*}|\mathbf{x}^n;\theta_u). $} \notag
\end{align} 
After the two-stream deep network component is trained, we estimate the parameters $\theta_p$ in Eq.~\eqref{eq:compat} by cross-validation.
 
The overall EM procedure starts with an M step with an initial value of $\bh^n$. We assume the initial hidden variable states ($\mathbf{h}_{0}^n$) to be consistent with the image-level labels: 
$\mathbf{h}^n_{0} = y^n$. 
The model parameters are fine-tuned by training the two-stream CNN+CRF with those initial labels.  
This is important because the CNN is pre-trained for object recognition on ImageNet and therefore the estimation of change regions in the initial E-step does not generate reasonable ground-truths.

\vspace{-0.2em}
\section{Experiments}

\subsection{CNN Implementation}
\label{subsec:CNNimp}
{The network weights are initialized from a pre-trained VGG network (on ImageNet).} 
The network splits into two portions after the fourth pooling layer.
As we need a coarse segmentation map ($32\times 32$) at the output of the segmentation branch, enlarged paired images of size $512\times 512$ are fed to the CNN. 
Moreover, the convolution filter size in FC1 (segmentation branch) is kept to $1\times 1$ (in contrast to a $7\times 7$ filter size in FC1$'$) to avoid the additional decrease in resolution of the $32\times 32$ output map.

\begin{figure*}[t]
\centering
\scalebox{.75}{
\includegraphics[trim=16pt 8pt 0pt 0pt, clip=true,width=0.285\columnwidth]{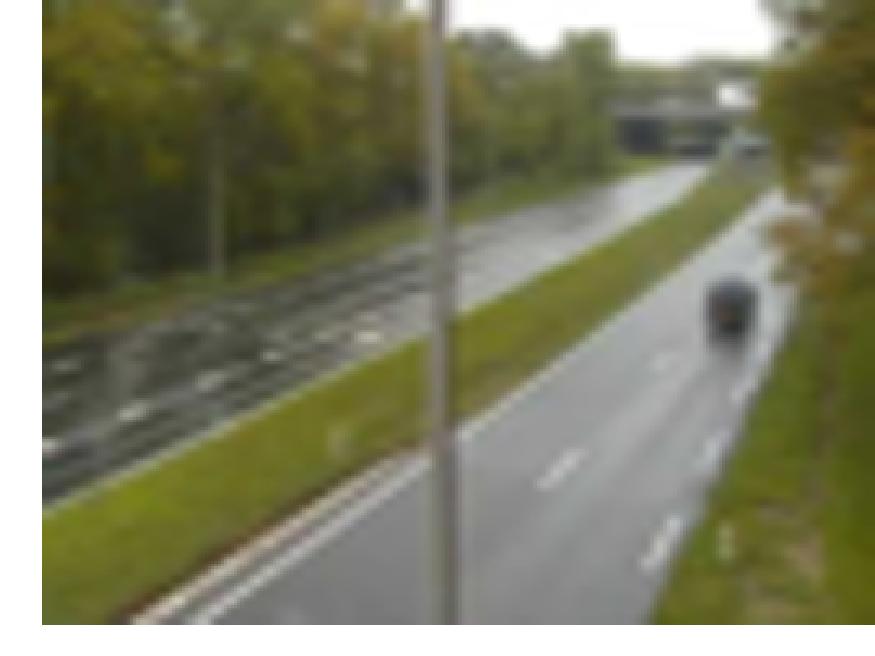} 
\includegraphics[trim=16pt 8pt 0pt 0pt, clip=true,width=0.285\columnwidth]{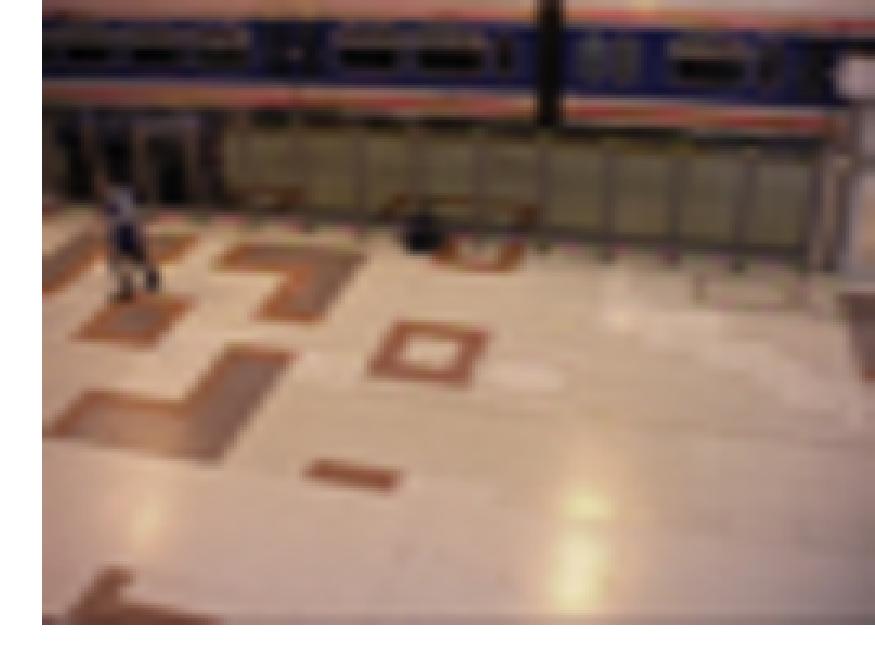}
\includegraphics[trim=16pt 8pt 0pt 0pt, clip=true,width=0.285\columnwidth]{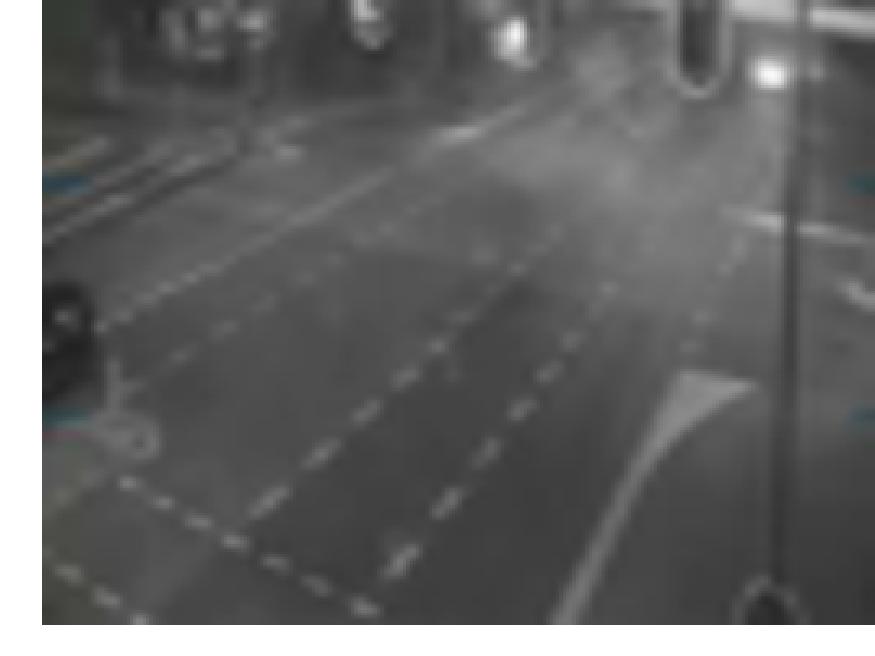}
\includegraphics[trim=16pt 8pt 0pt 0pt, clip=true,width=0.285\columnwidth]{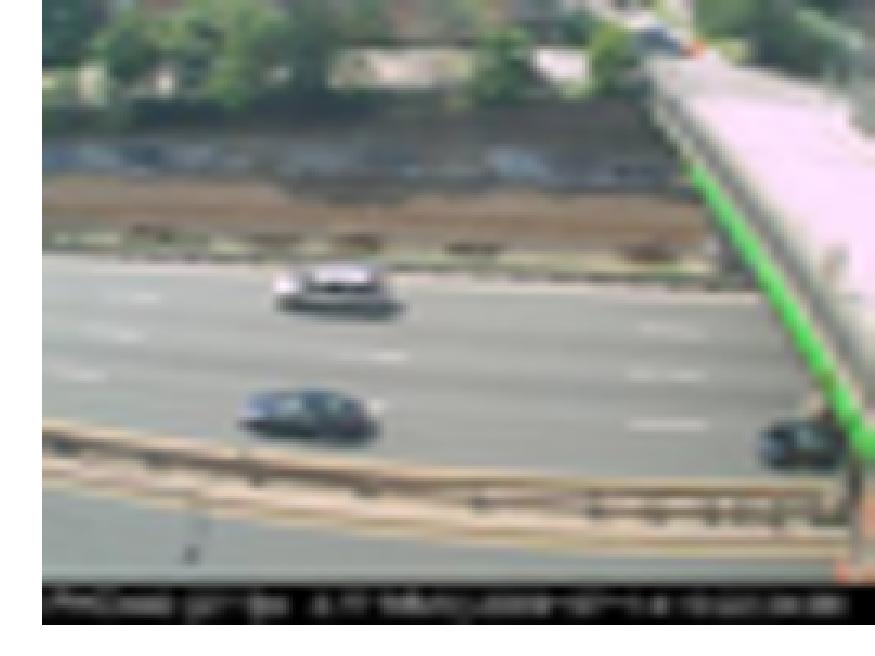}
\includegraphics[trim=16pt 8pt 0pt 0pt, clip=true,width=0.285\columnwidth]{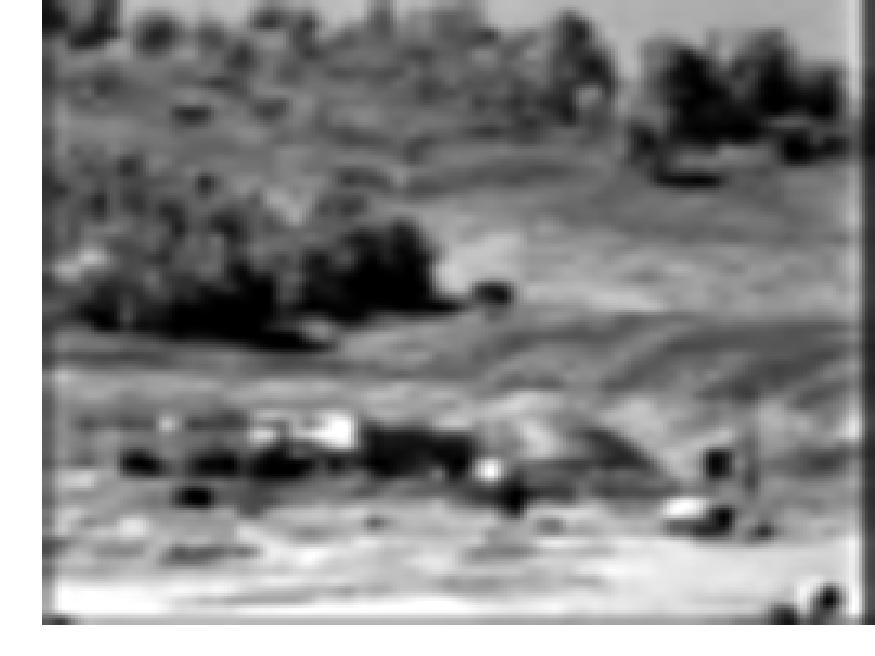}
\includegraphics[trim=16pt 8pt 0pt 0pt, clip=true,width=0.285\columnwidth]{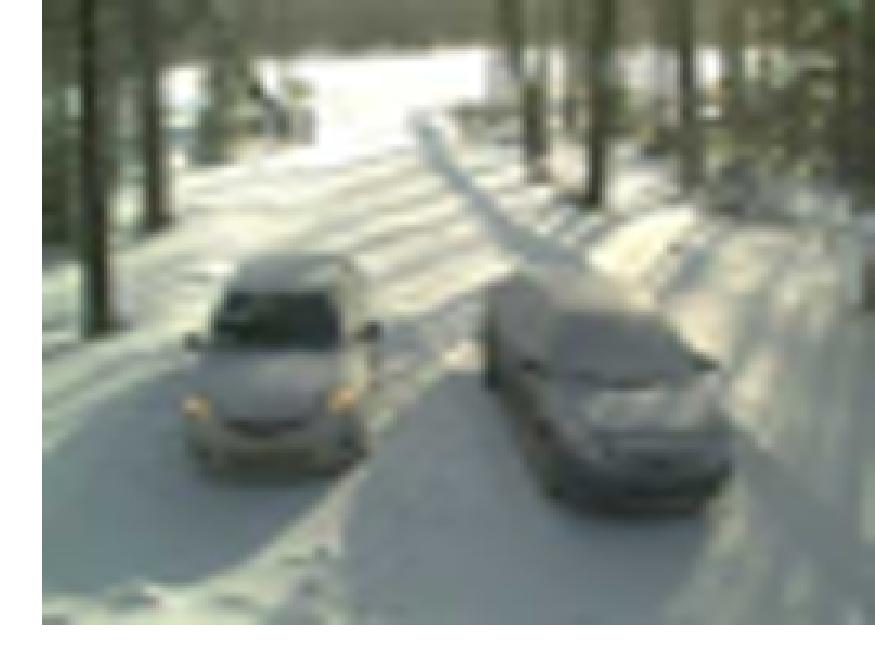}
\includegraphics[trim=16pt 8pt 0pt 0pt, clip=true,width=0.285\columnwidth]{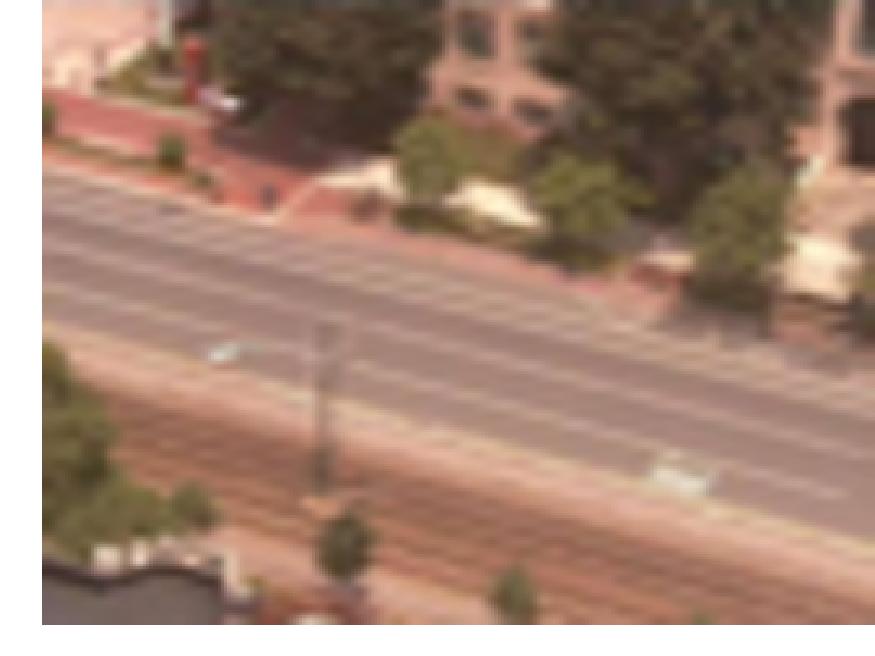}
\includegraphics[trim=16pt 8pt 0pt 0pt, clip=true,width=0.285\columnwidth]{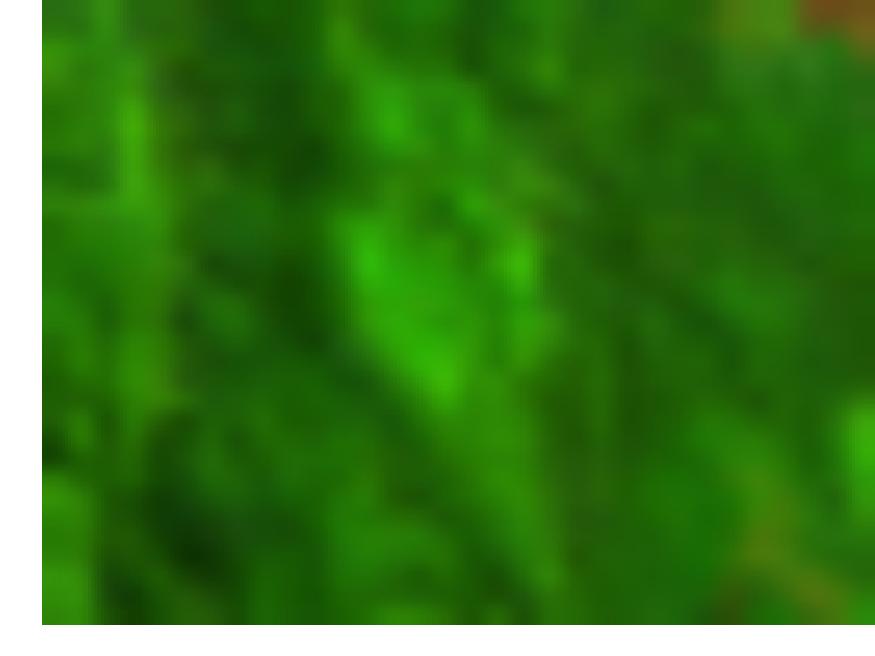}
\includegraphics[trim=16pt 8pt 0pt 0pt, clip=true,width=0.285\columnwidth]{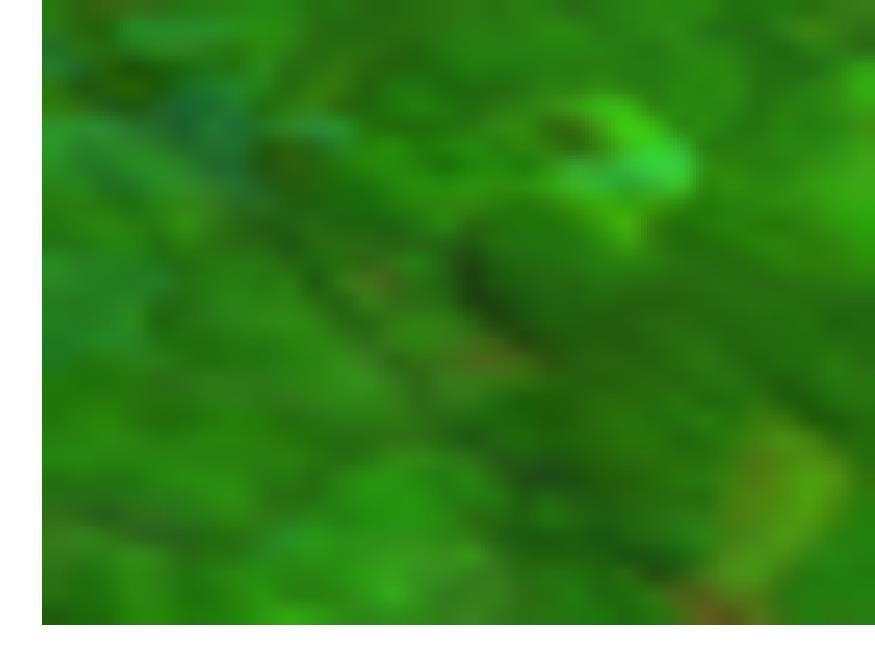}
}
\\
\scalebox{.75}{
\includegraphics[trim=16pt 8pt 0pt 0pt, clip=true,width=0.285\columnwidth]{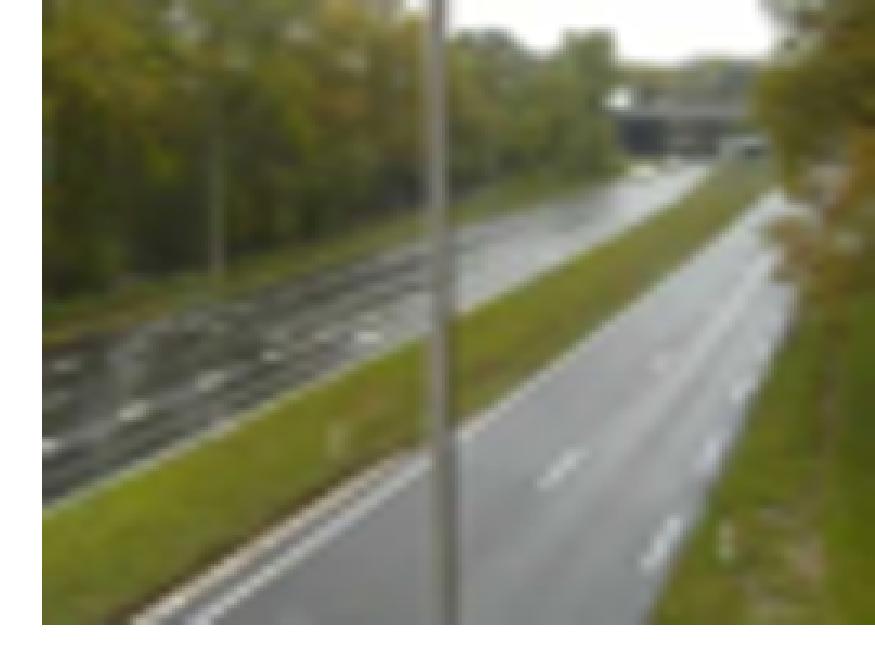}
\includegraphics[trim=16pt 8pt 0pt 0pt, clip=true,width=0.285\columnwidth]{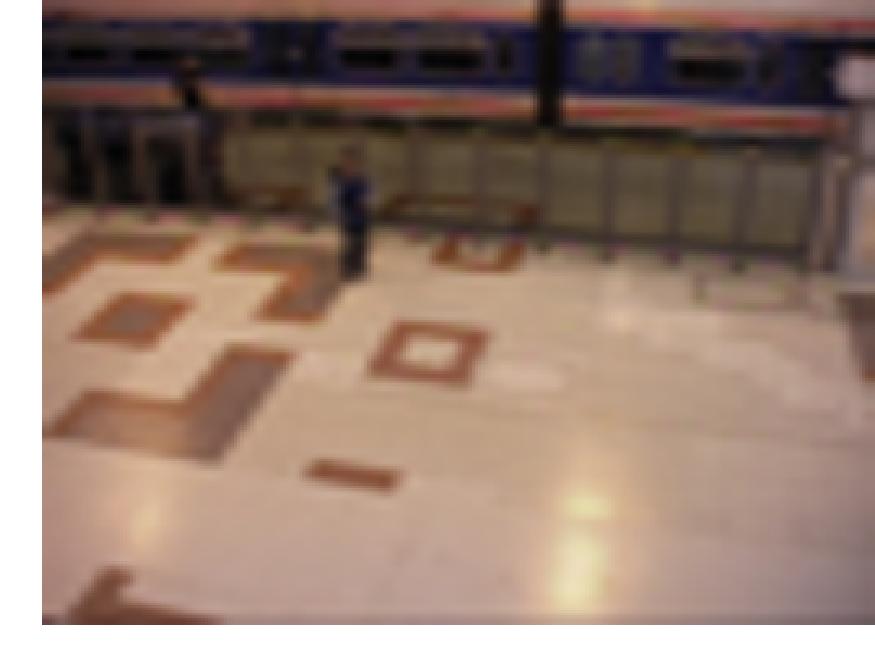}
\includegraphics[trim=16pt 8pt 0pt 0pt, clip=true,width=0.285\columnwidth]{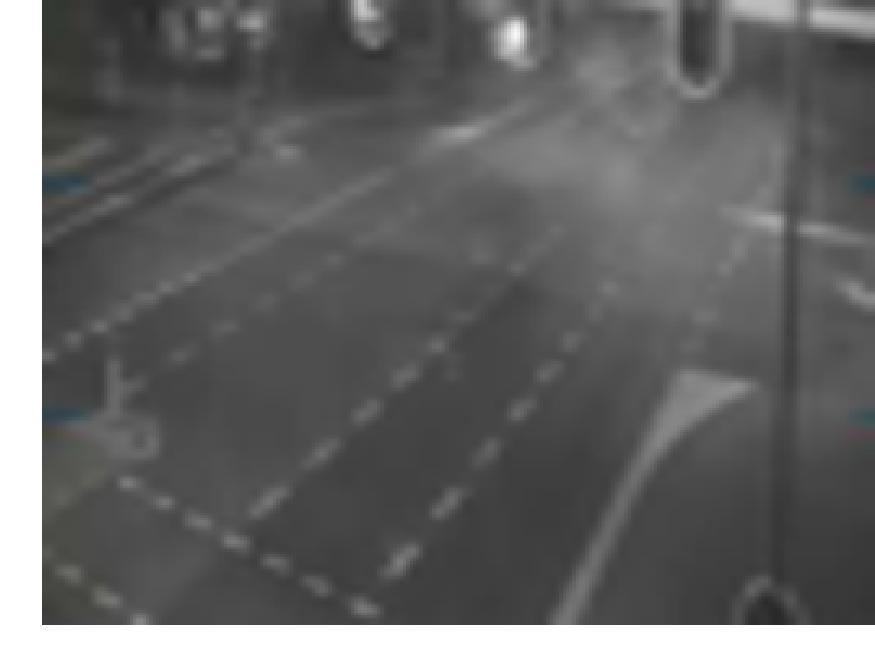}
\includegraphics[trim=16pt 8pt 0pt 0pt, clip=true,width=0.285\columnwidth]{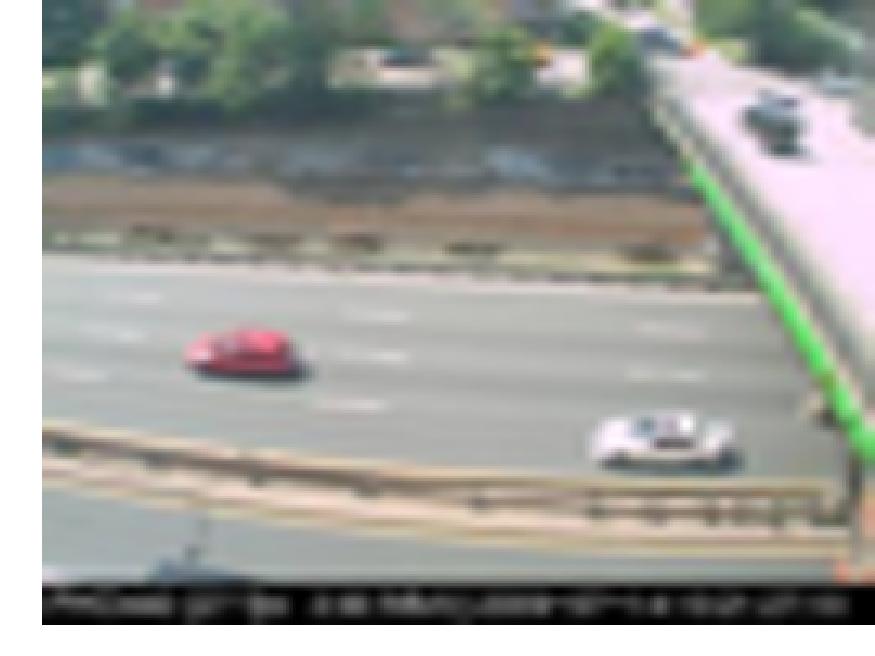}
\includegraphics[trim=16pt 8pt 0pt 0pt, clip=true,width=0.285\columnwidth]{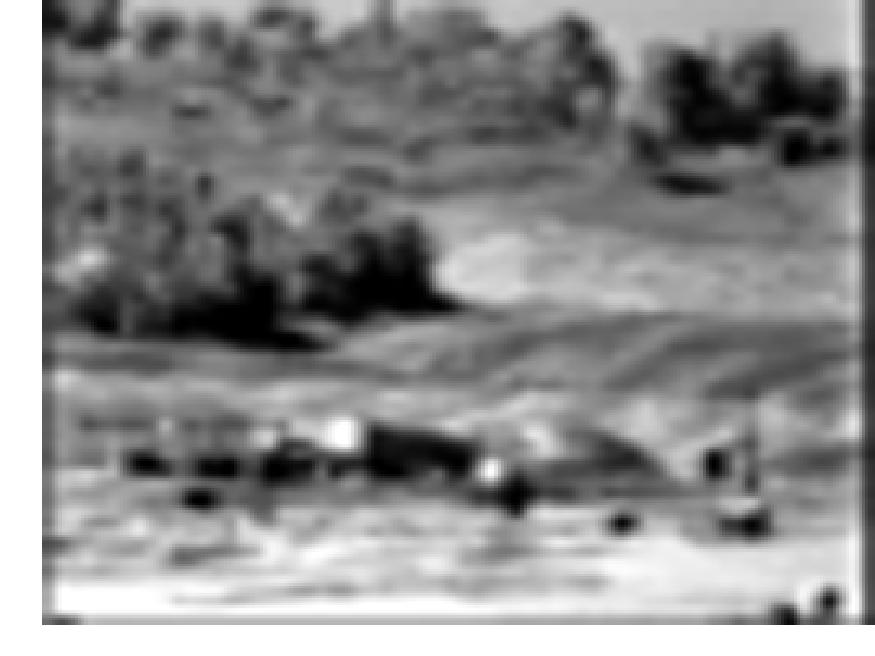}
\includegraphics[trim=16pt 8pt 0pt 0pt, clip=true,width=0.285\columnwidth]{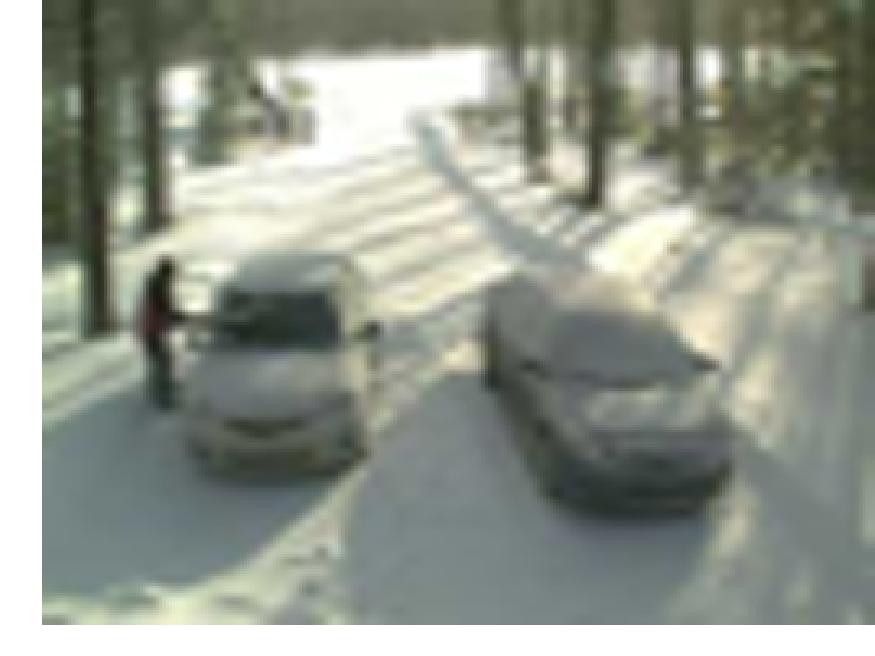}
\includegraphics[trim=16pt 8pt 0pt 0pt, clip=true,width=0.285\columnwidth]{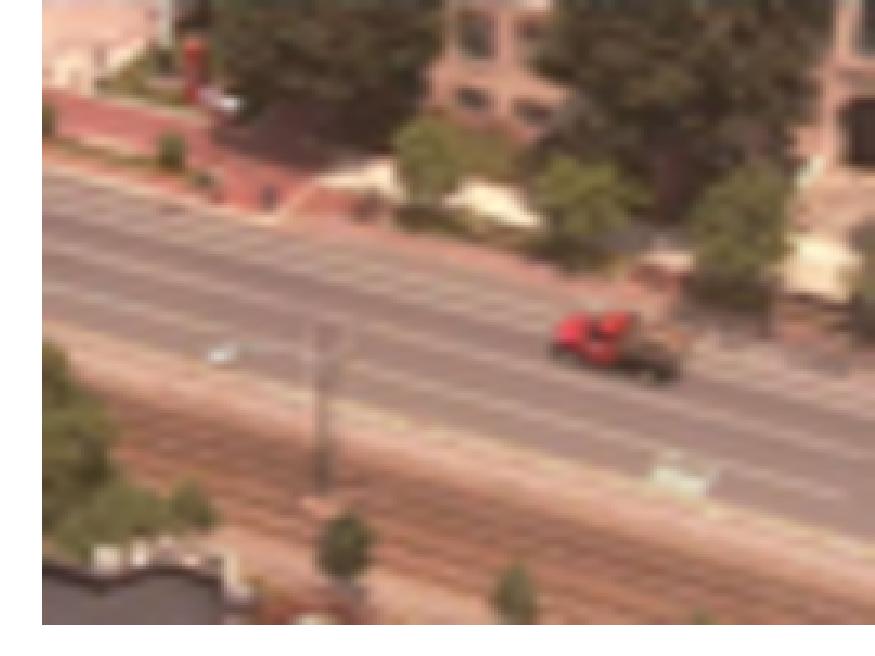}
\includegraphics[trim=16pt 8pt 0pt 0pt, clip=true,width=0.285\columnwidth]{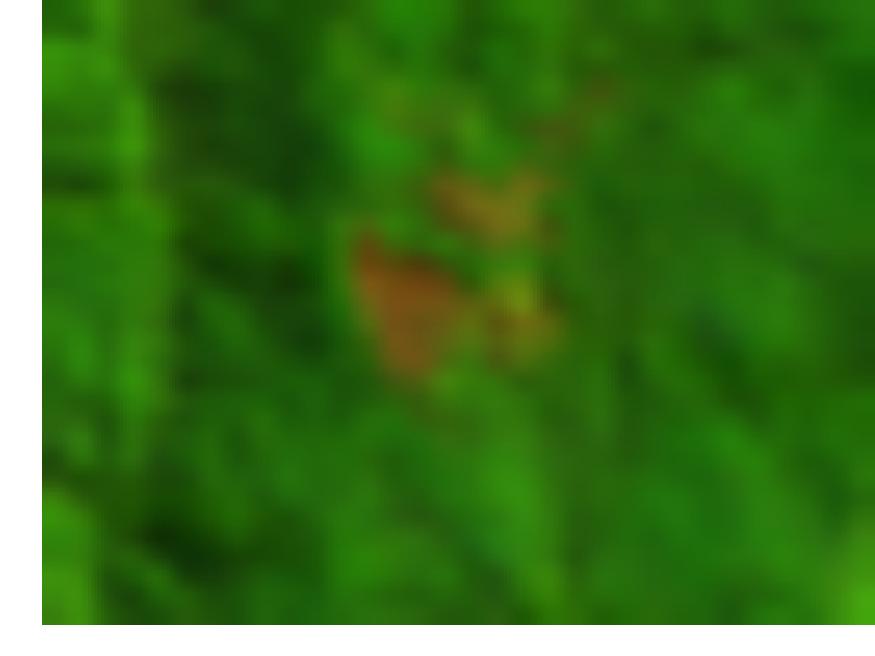}
\includegraphics[trim=16pt 8pt 0pt 0pt, clip=true,width=0.285\columnwidth]{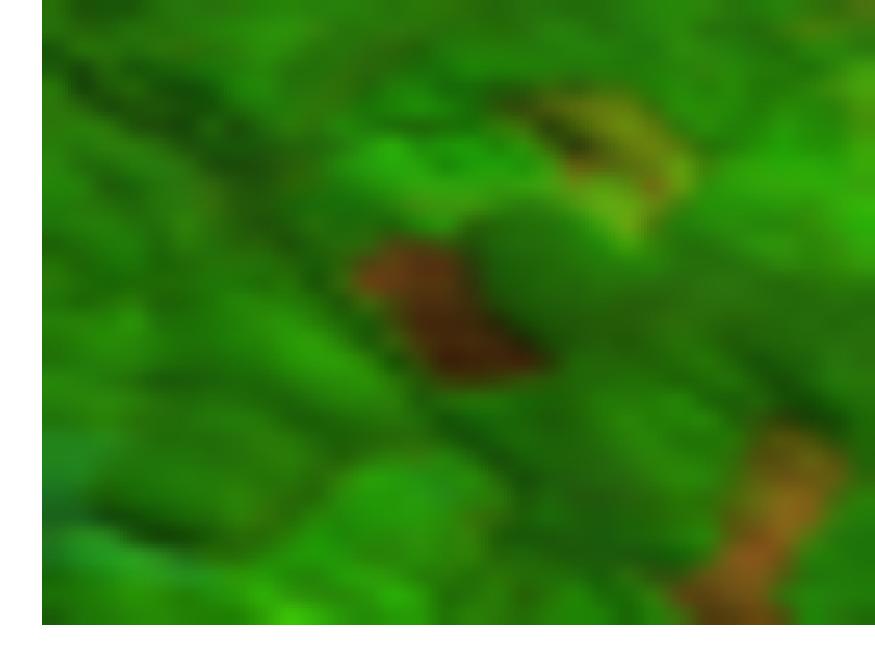}
}
\\
\scalebox{.75}{
\includegraphics[trim=16pt 8pt 0pt 0pt, clip=true,width=0.285\columnwidth]{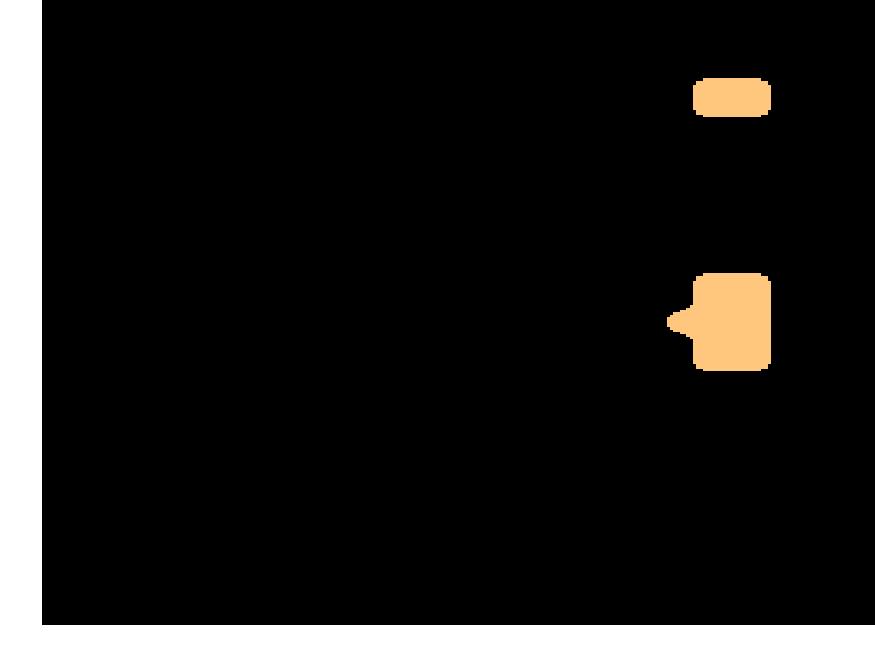}
\includegraphics[trim=16pt 8pt 0pt 0pt, clip=true,width=0.285\columnwidth]{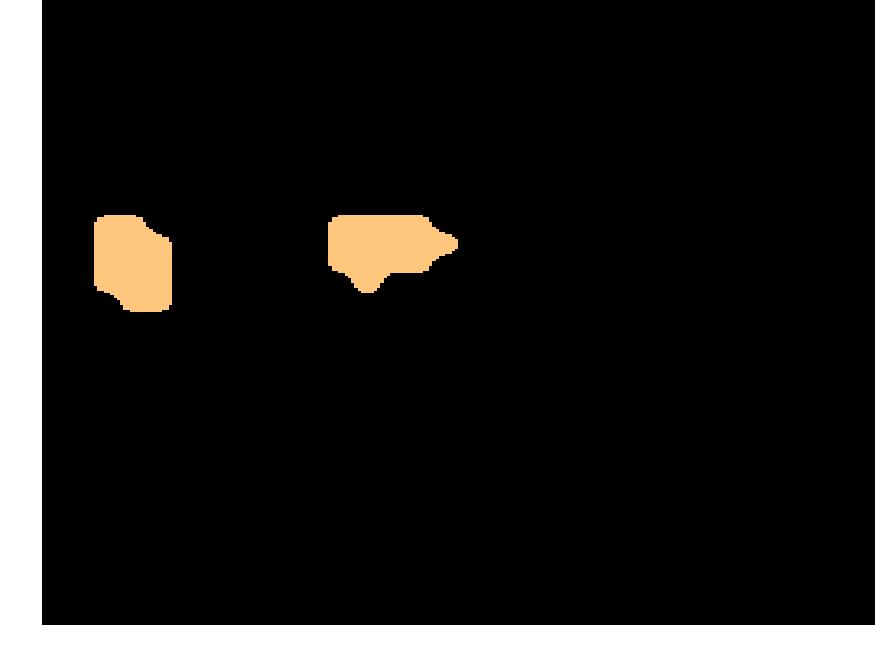}
\includegraphics[trim=16pt 8pt 0pt 0pt, clip=true,width=0.285\columnwidth]{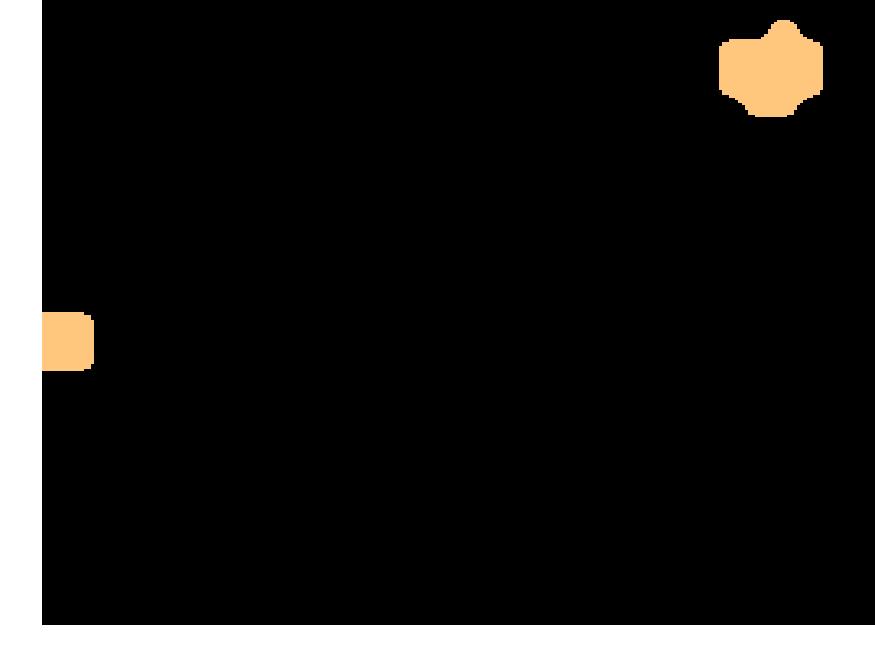}
\includegraphics[trim=16pt 8pt 0pt 0pt, clip=true,width=0.285\columnwidth]{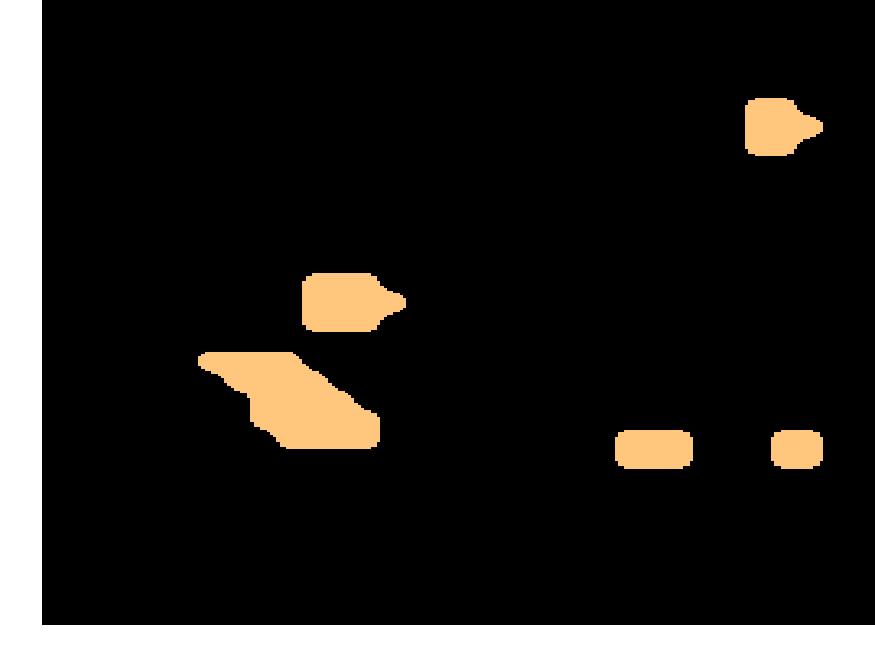}
\includegraphics[trim=16pt 8pt 0pt 0pt, clip=true,width=0.285\columnwidth]{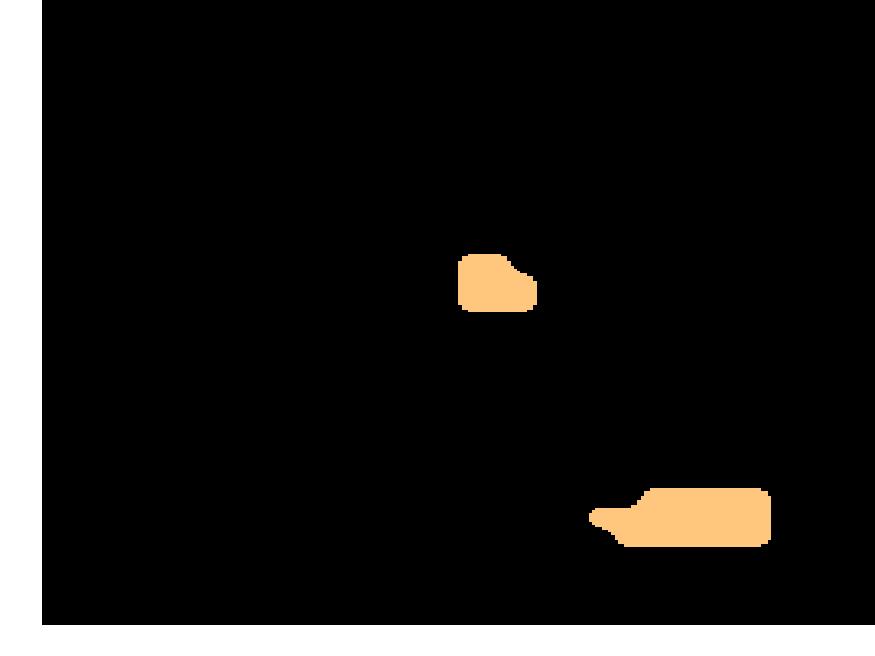}
\includegraphics[trim=16pt 8pt 0pt 0pt, clip=true,width=0.285\columnwidth]{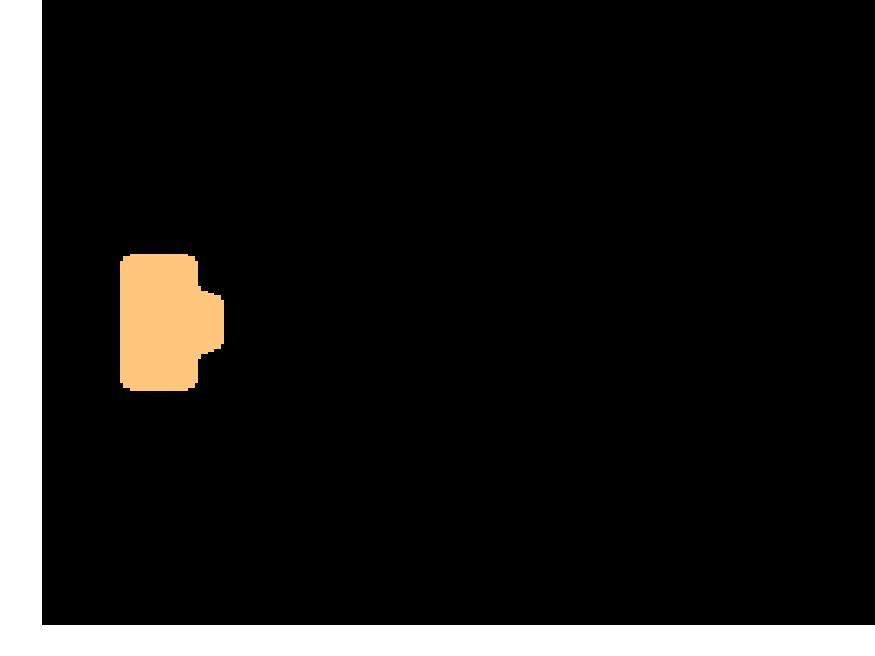}
\includegraphics[trim=16pt 8pt 0pt 0pt, clip=true,width=0.285\columnwidth]{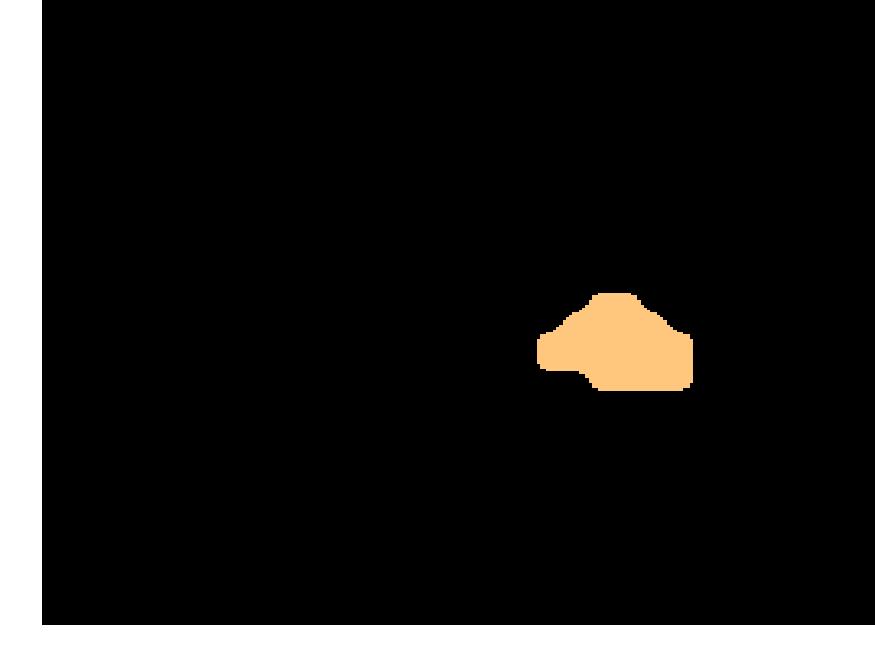}
\includegraphics[trim=16pt 8pt 0pt 0pt, clip=true,width=0.285\columnwidth]{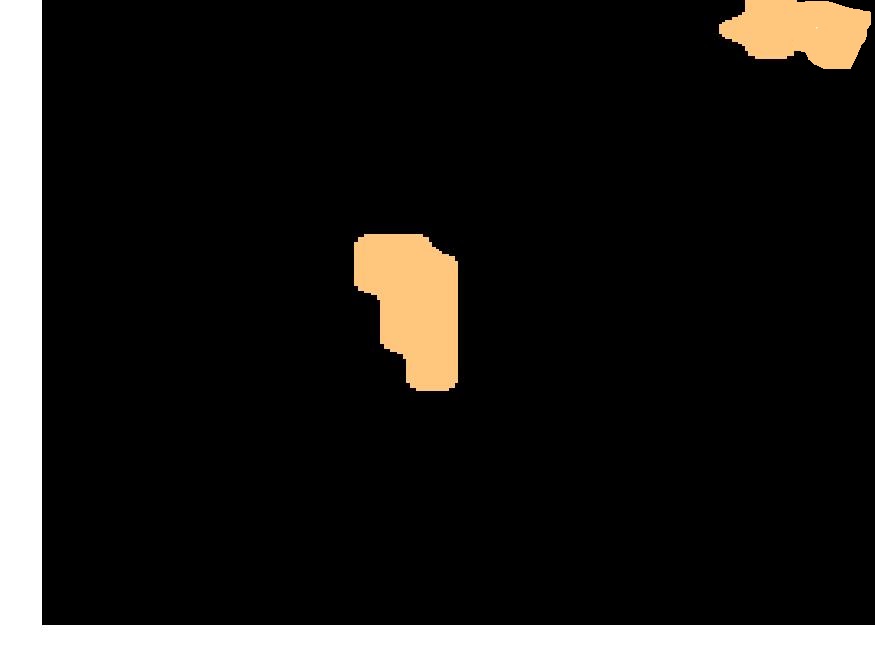}
\includegraphics[trim=16pt 8pt 0pt 0pt, clip=true,width=0.285\columnwidth]{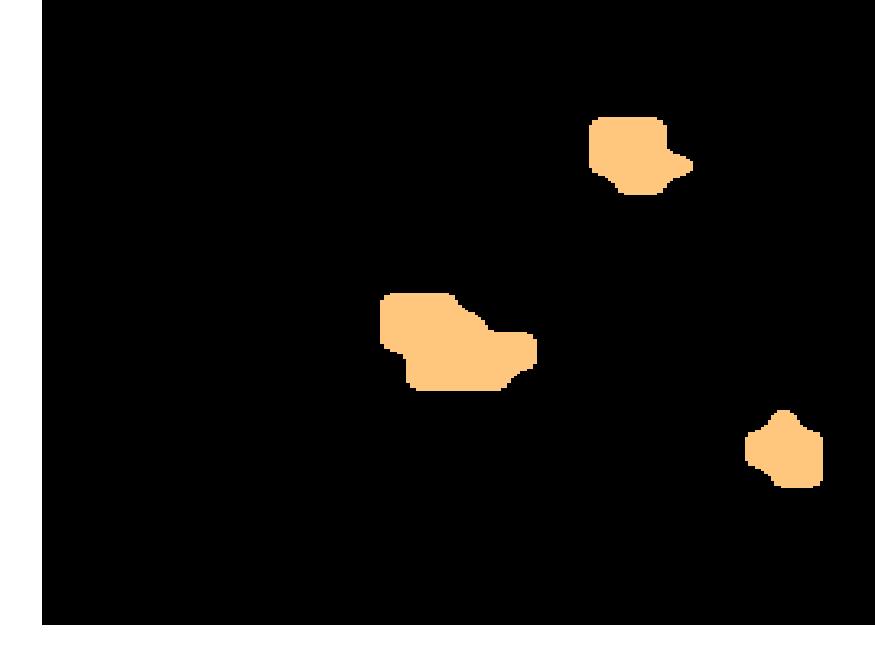}
}
\\
\scalebox{.75}{
\includegraphics[trim=16pt 8pt 0pt 0pt, clip=true,width=0.285\columnwidth]{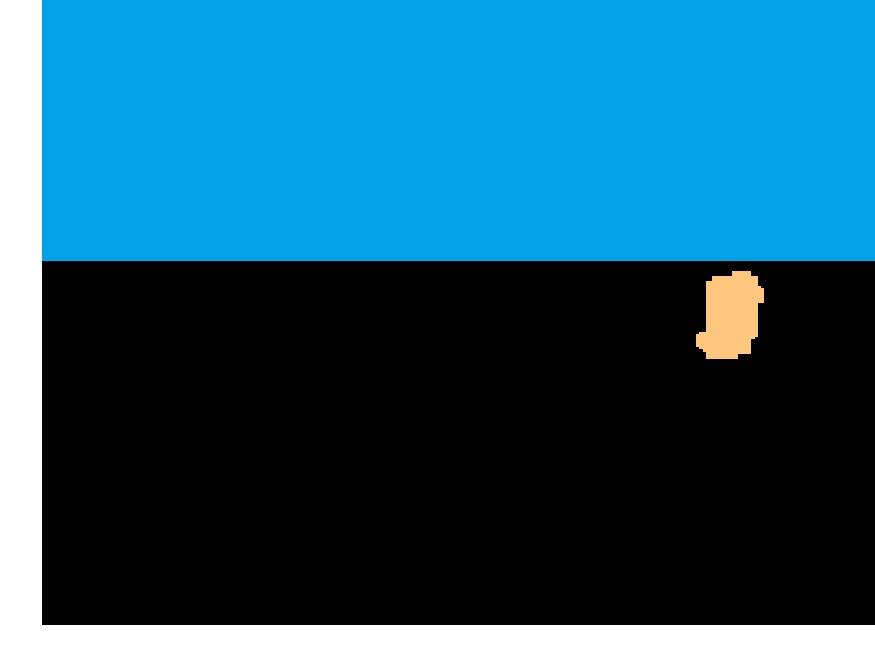}
\includegraphics[trim=16pt 8pt 0pt 0pt, clip=true,width=0.285\columnwidth]{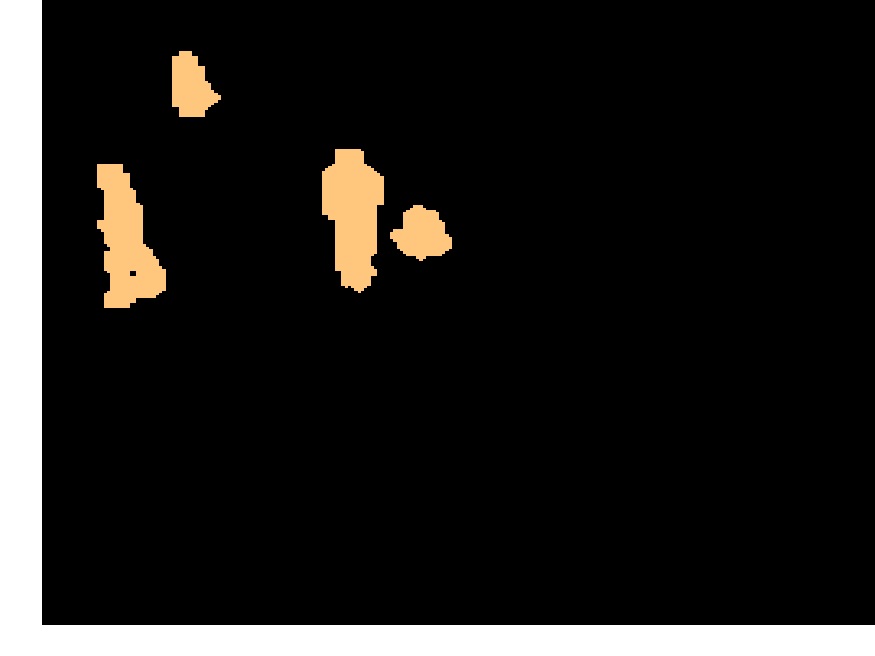}
\includegraphics[trim=16pt 8pt 0pt 0pt, clip=true,width=0.285\columnwidth]{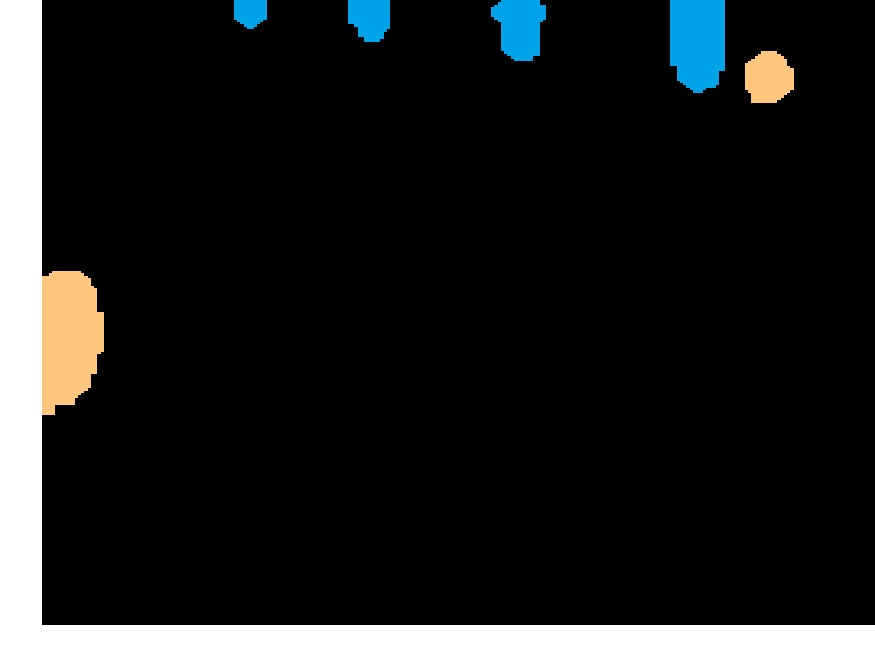}
\includegraphics[trim=16pt 8pt 0pt 0pt, clip=true,width=0.285\columnwidth]{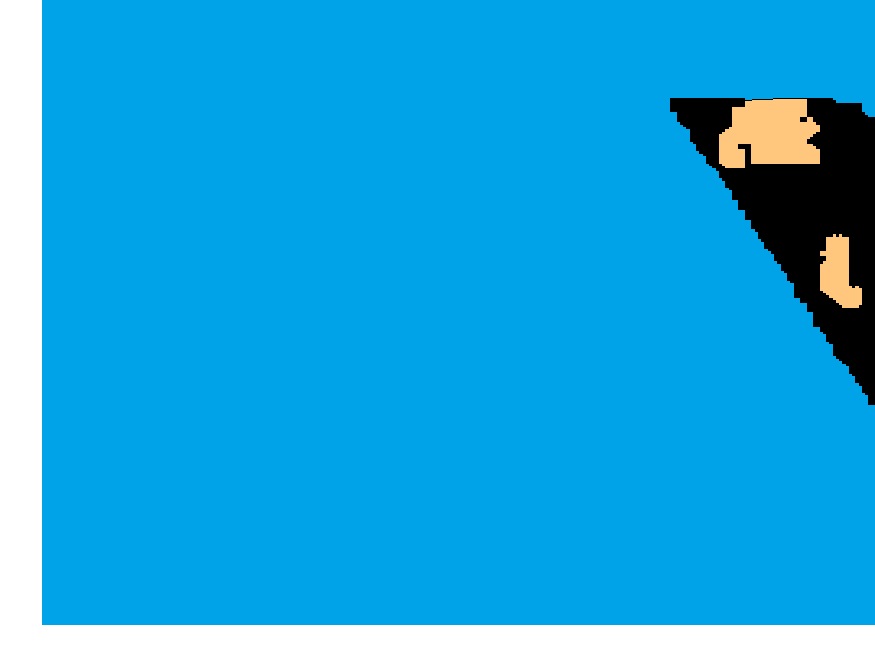}
\includegraphics[trim=16pt 8pt 0pt 0pt, clip=true,width=0.285\columnwidth]{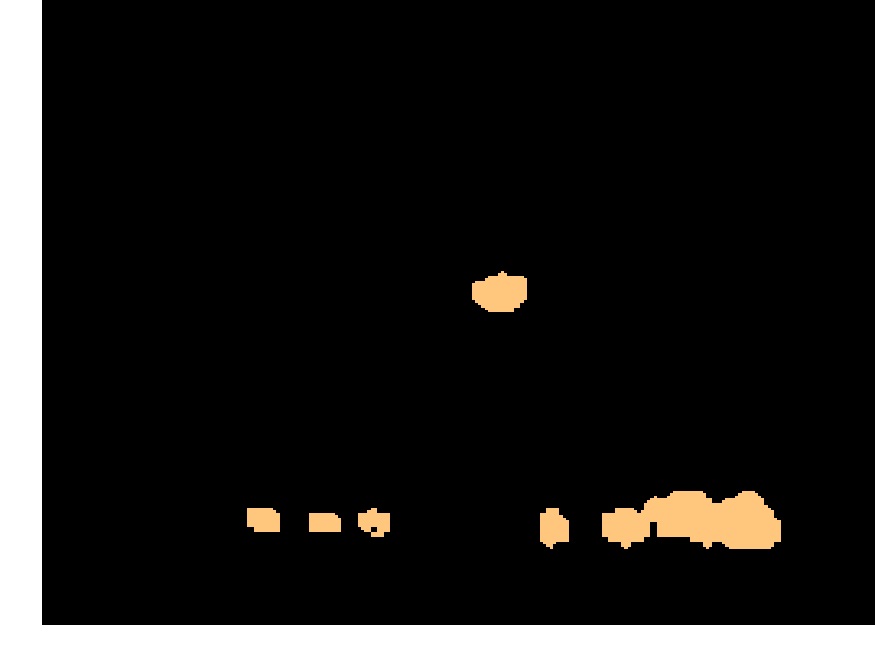}
\includegraphics[trim=16pt 8pt 0pt 0pt, clip=true,width=0.285\columnwidth]{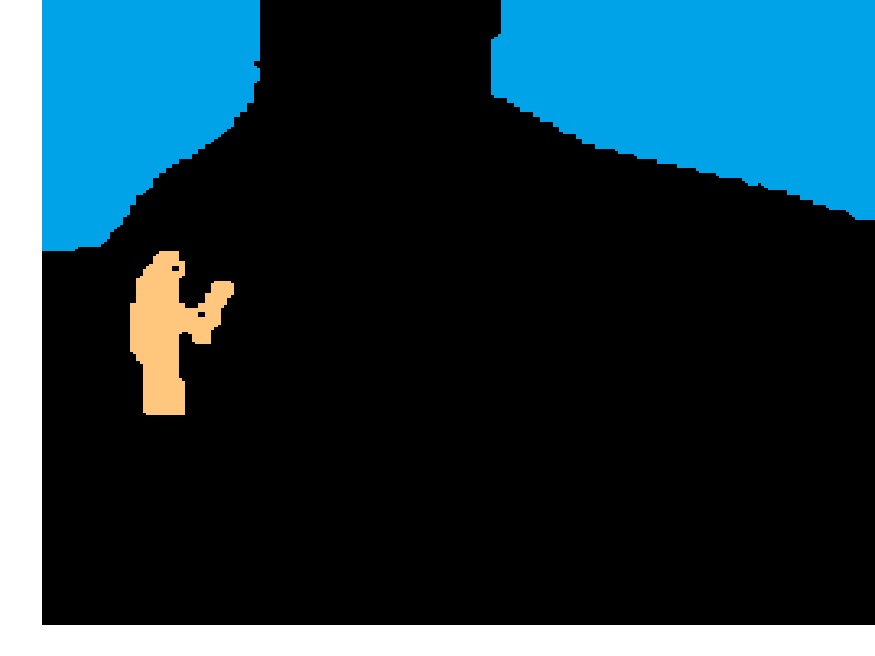}
\includegraphics[trim=16pt 8pt 0pt 0pt, clip=true,width=0.285\columnwidth]{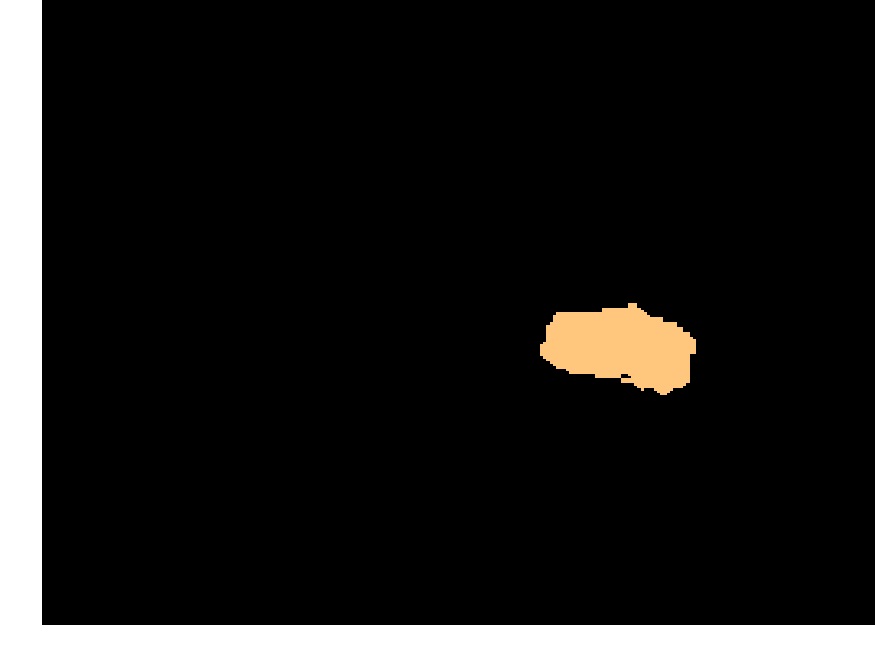}
\includegraphics[trim=16pt 8pt 0pt 0pt, clip=true,width=0.285\columnwidth]{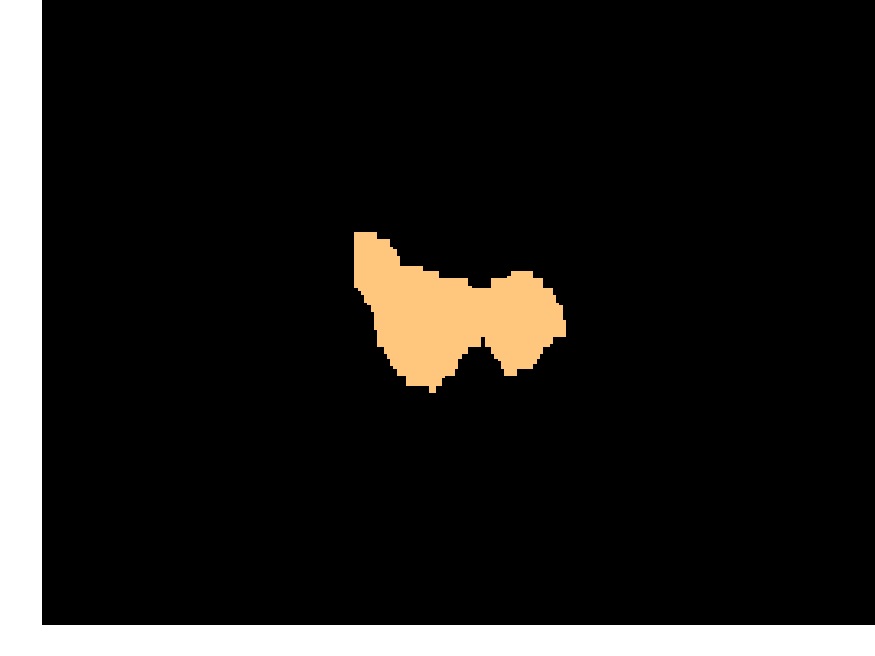}
\includegraphics[trim=16pt 8pt 0pt 0pt, clip=true,width=0.285\columnwidth]{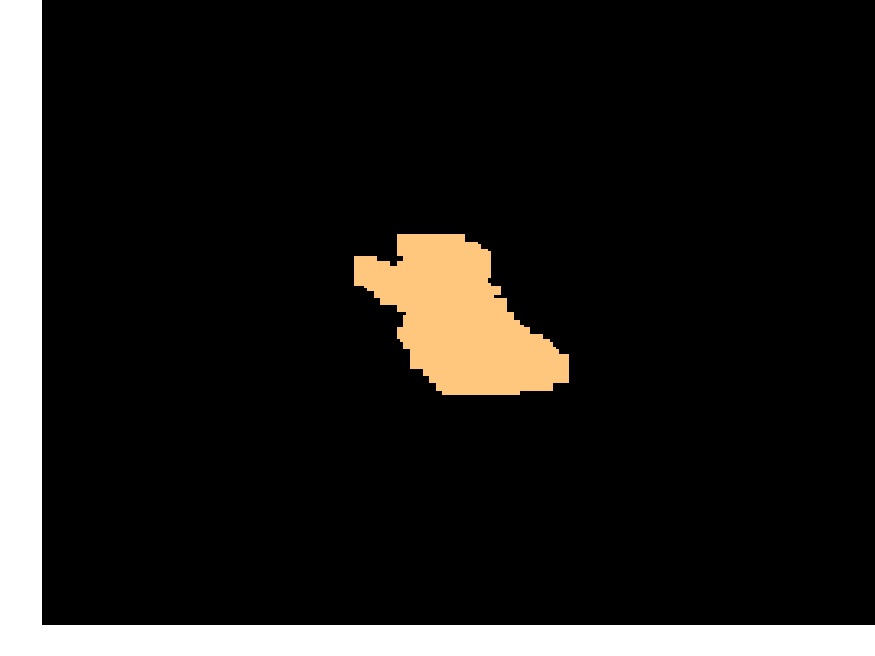}
}
\\
\scalebox{.75}{
\includegraphics[trim=16pt 8pt 0pt 0pt, clip=true, width=0.285\columnwidth]{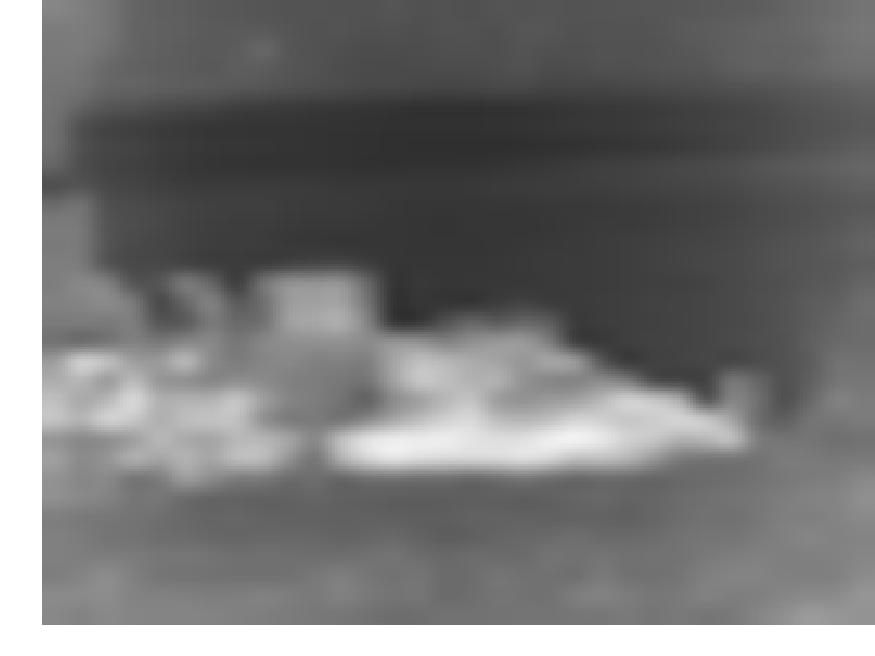}
\includegraphics[trim=16pt 8pt 0pt 0pt, clip=true, width=0.285\columnwidth]{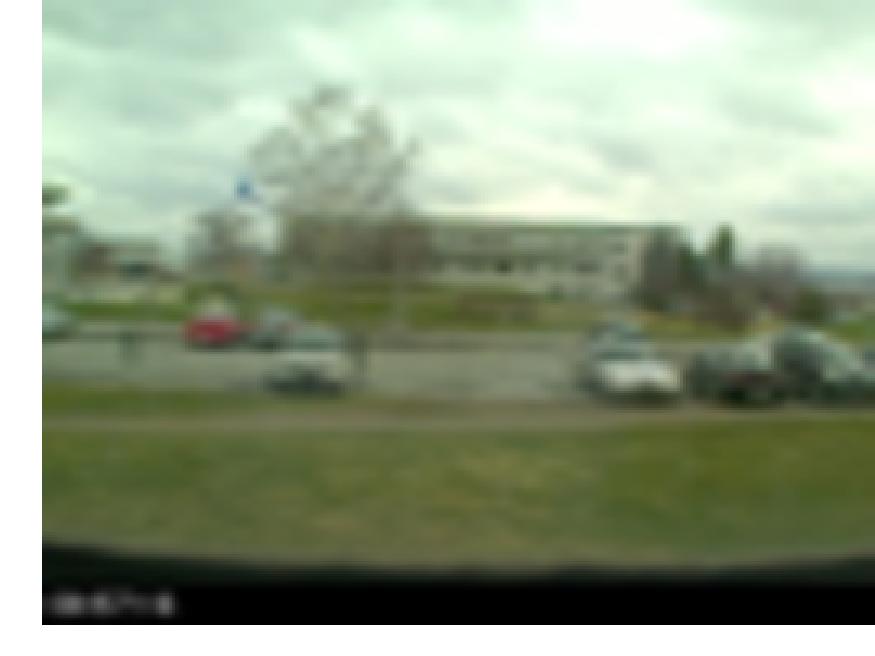}
\includegraphics[trim=16pt 8pt 0pt 0pt, clip=true, width=0.285\columnwidth]{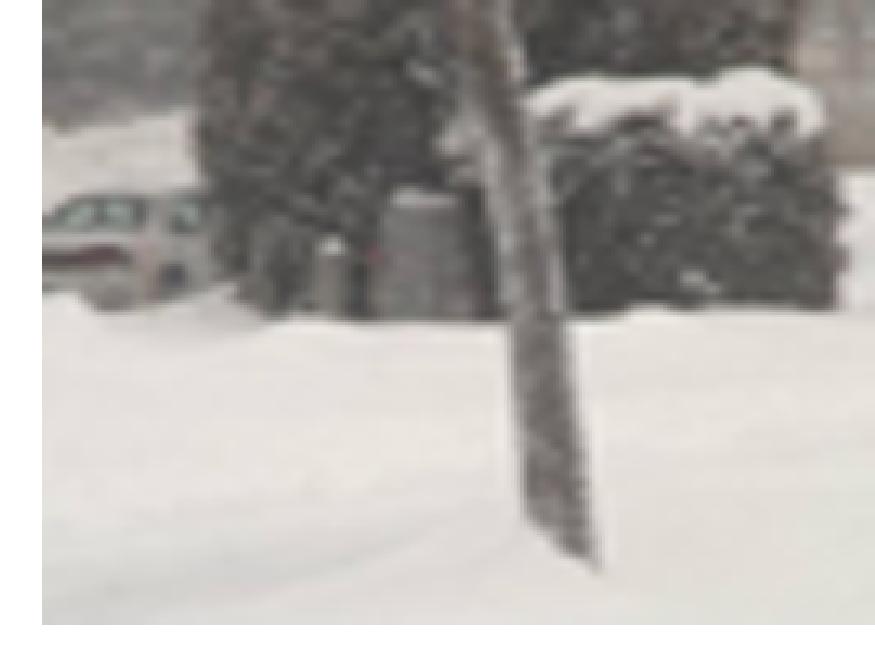}
\includegraphics[trim=16pt 8pt 0pt 0pt, clip=true, width=0.285\columnwidth]{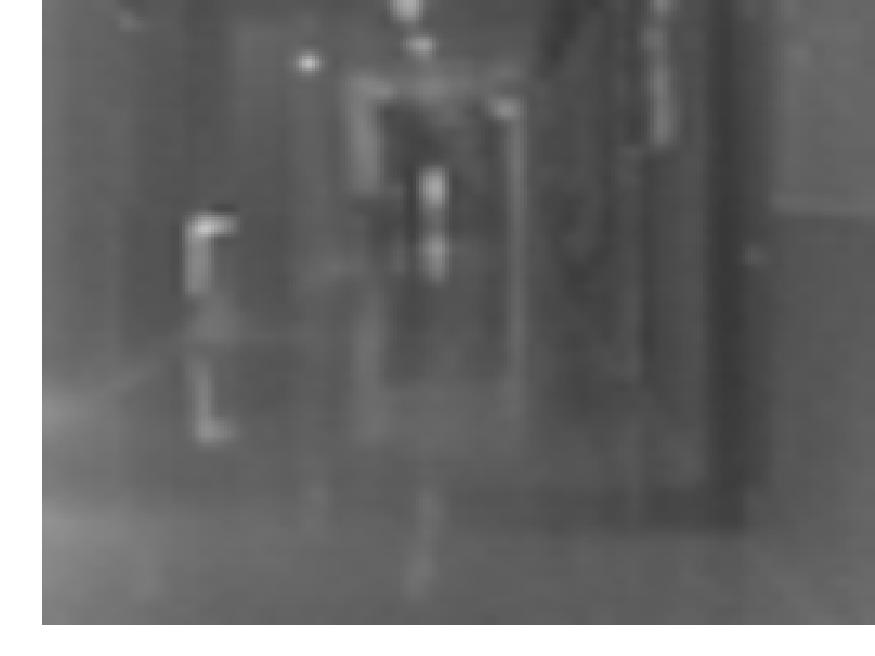}
\includegraphics[trim=16pt 8pt 0pt 0pt, clip=true, width=0.285\columnwidth]{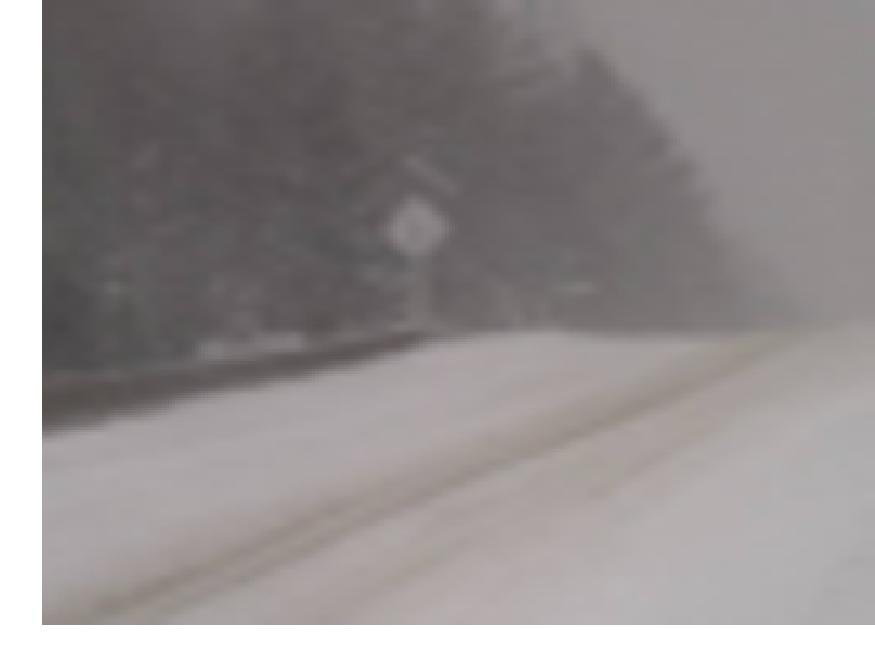}
\includegraphics[trim=16pt 8pt 0pt 0pt, clip=true, width=0.285\columnwidth]{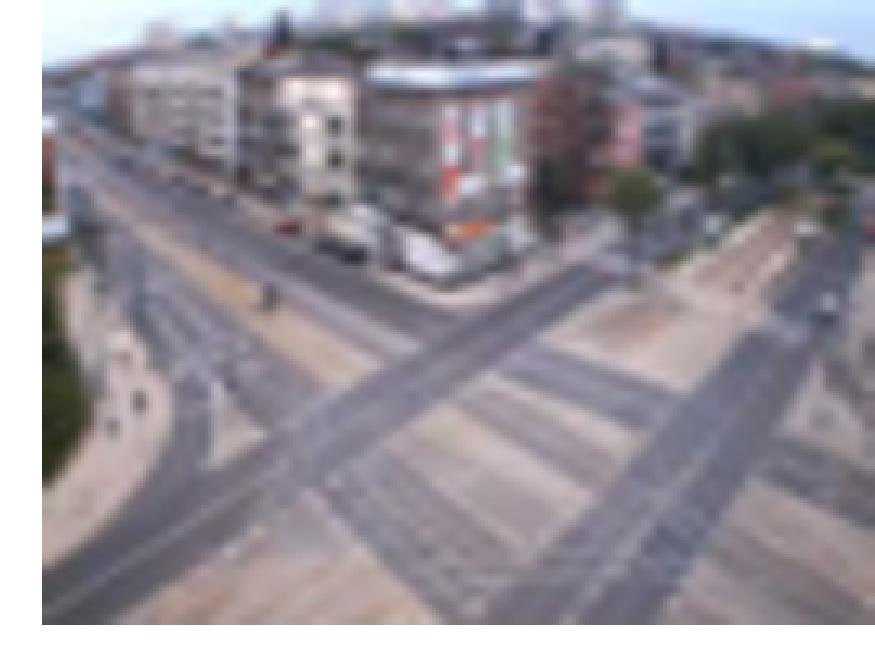}
\includegraphics[trim=16pt 8pt 0pt 0pt, clip=true, width=0.285\columnwidth]{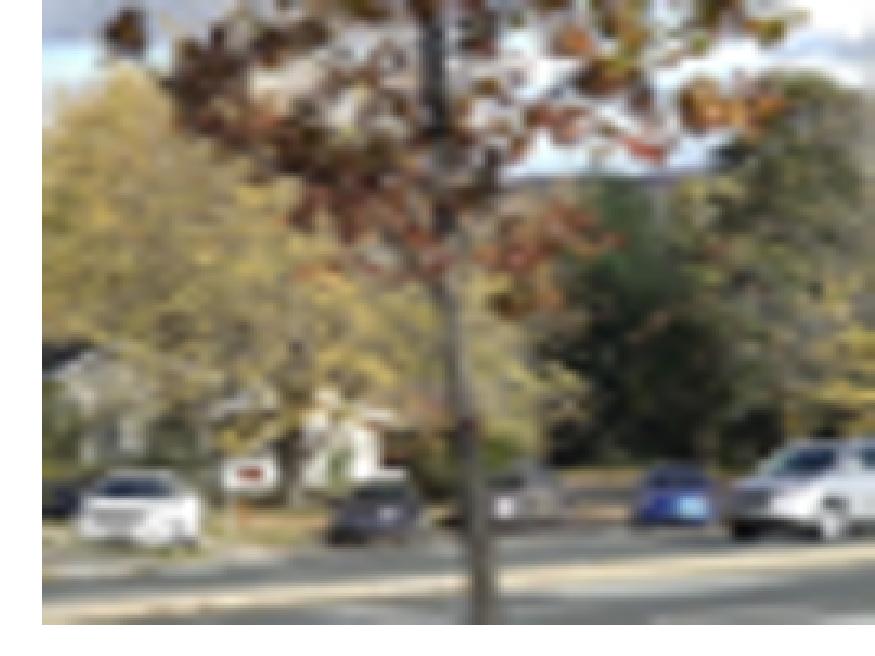}
\includegraphics[trim=16pt 8pt 0pt 0pt, clip=true,width=0.285\columnwidth]{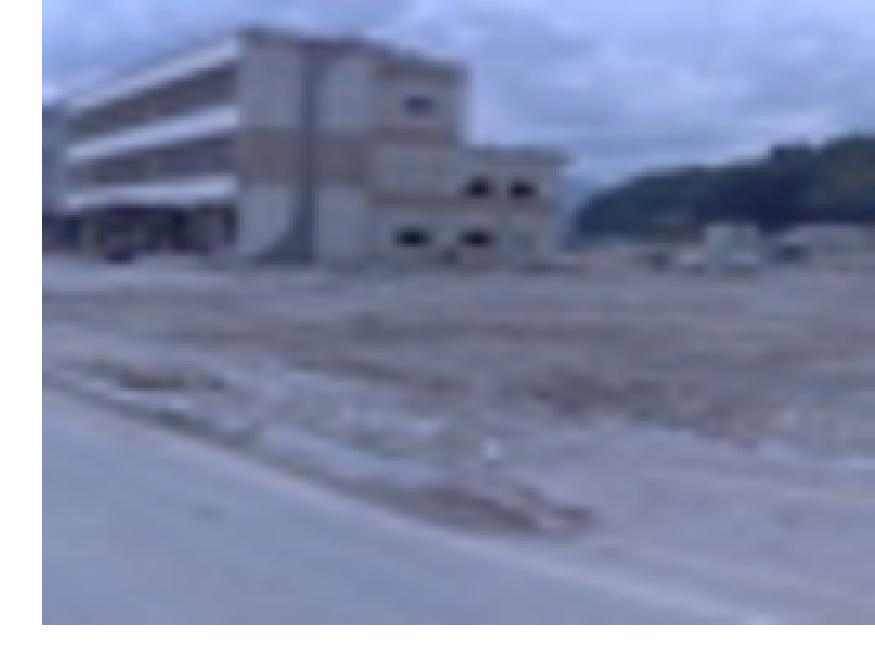}
\includegraphics[trim=16pt 8pt 0pt 0pt, clip=true,width=0.285\columnwidth]{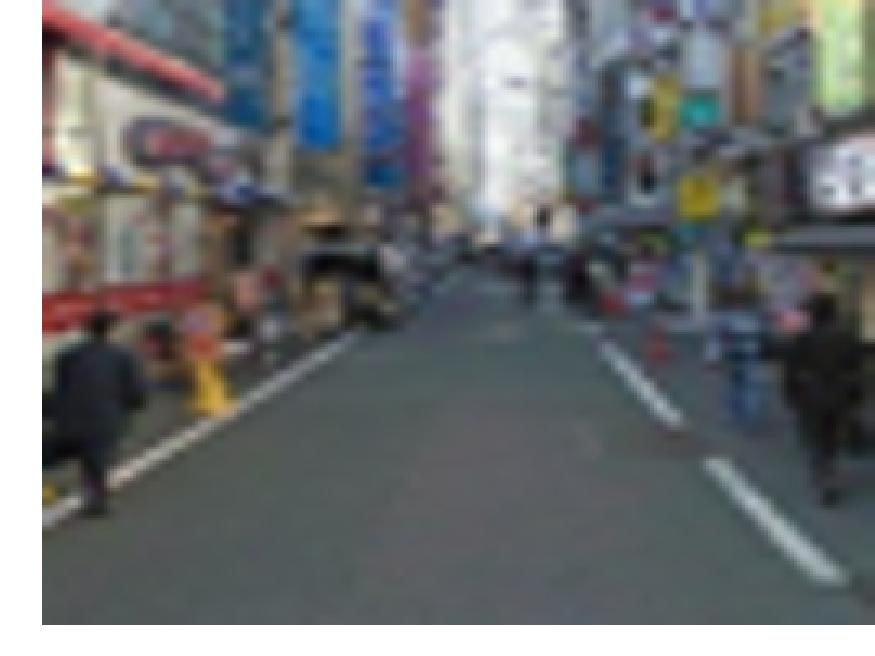}
}
\\
\scalebox{.75}{
\includegraphics[trim=16pt 8pt 0pt 0pt, clip=true, width=0.285\columnwidth]{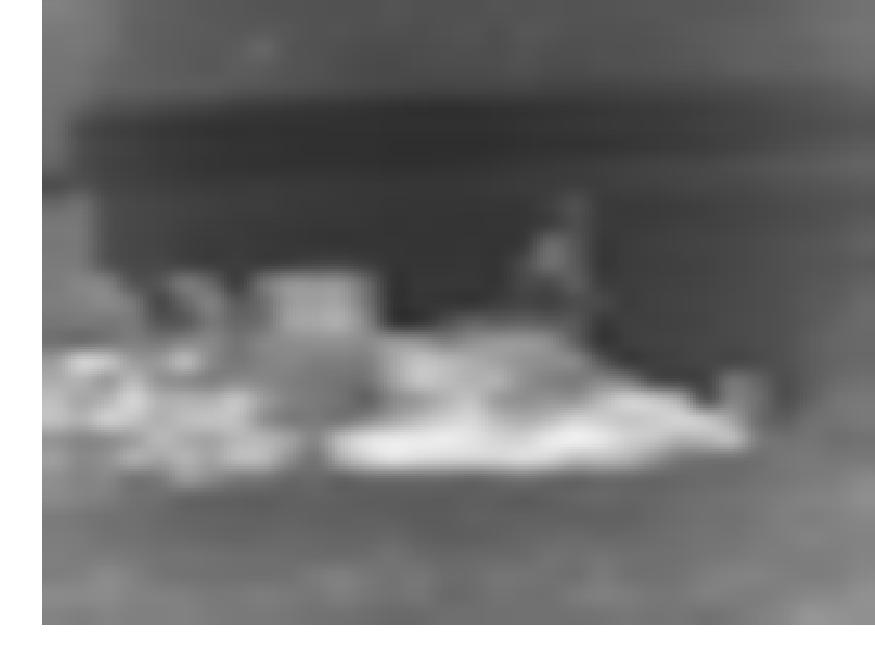}
\includegraphics[trim=16pt 8pt 0pt 0pt, clip=true, width=0.285\columnwidth]{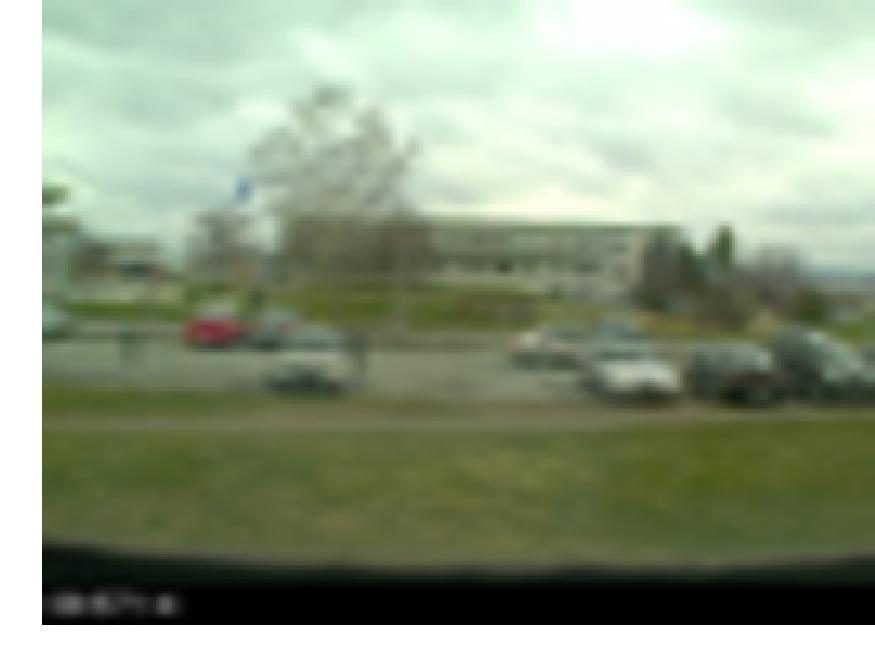}
\includegraphics[trim=16pt 8pt 0pt 0pt, clip=true, width=0.285\columnwidth]{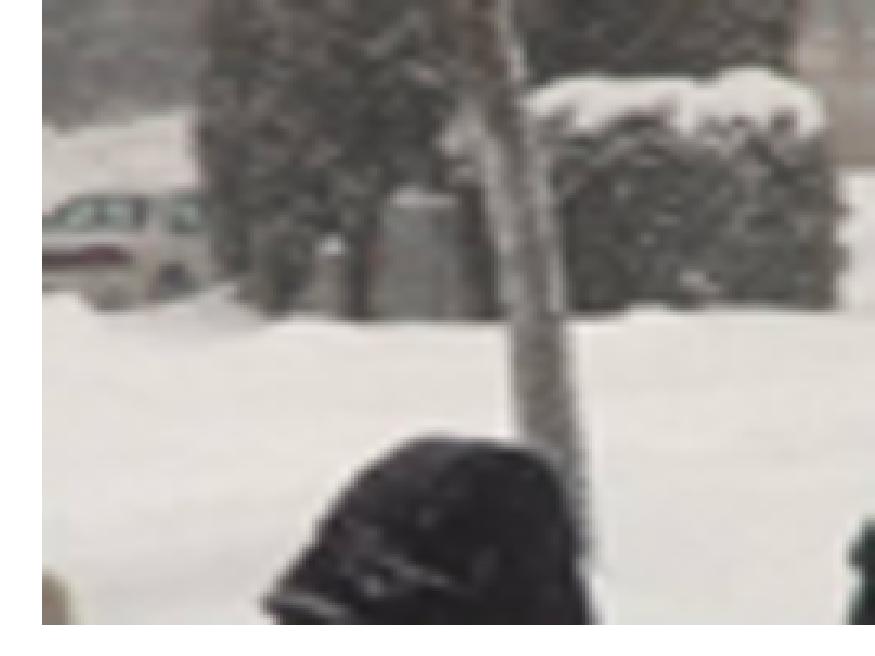}
\includegraphics[trim=16pt 8pt 0pt 0pt, clip=true, width=0.285\columnwidth]{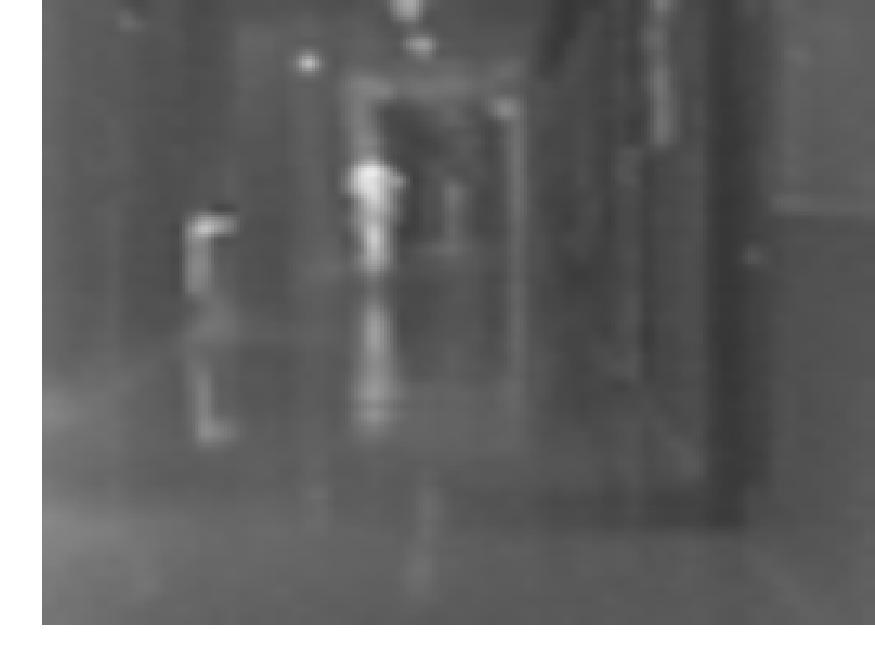}
\includegraphics[trim=16pt 8pt 0pt 0pt, clip=true, width=0.285\columnwidth]{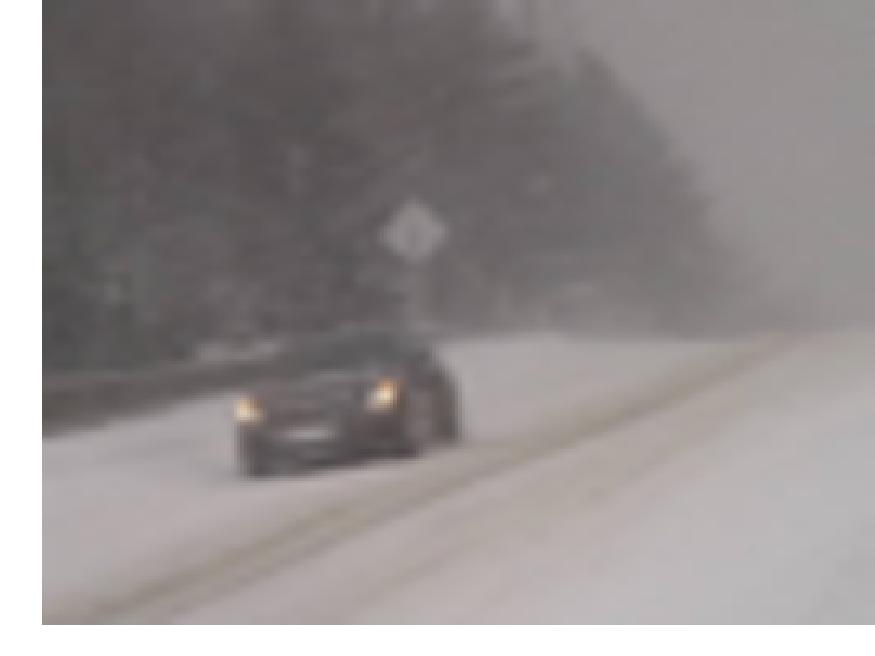}
\includegraphics[trim=16pt 8pt 0pt 0pt, clip=true, width=0.285\columnwidth]{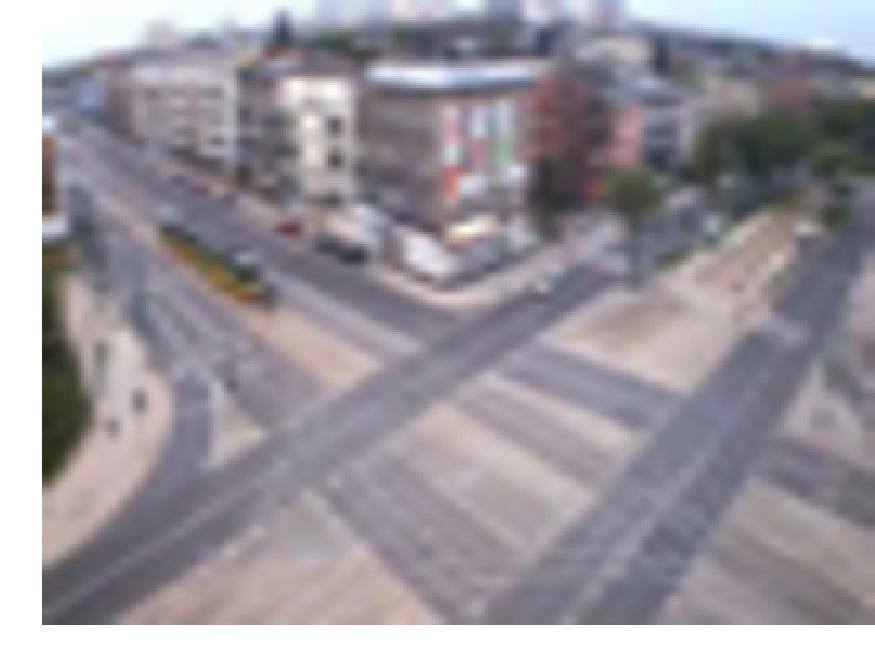}
\includegraphics[trim=16pt 8pt 0pt 0pt, clip=true, width=0.285\columnwidth]{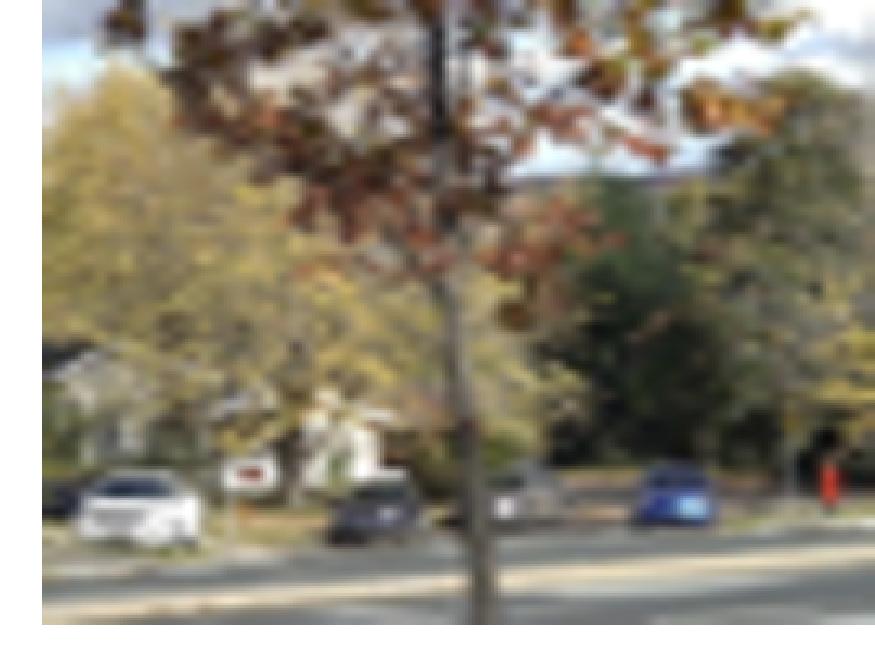}
\includegraphics[trim=16pt 8pt 0pt 0pt, clip=true,width=0.285\columnwidth]{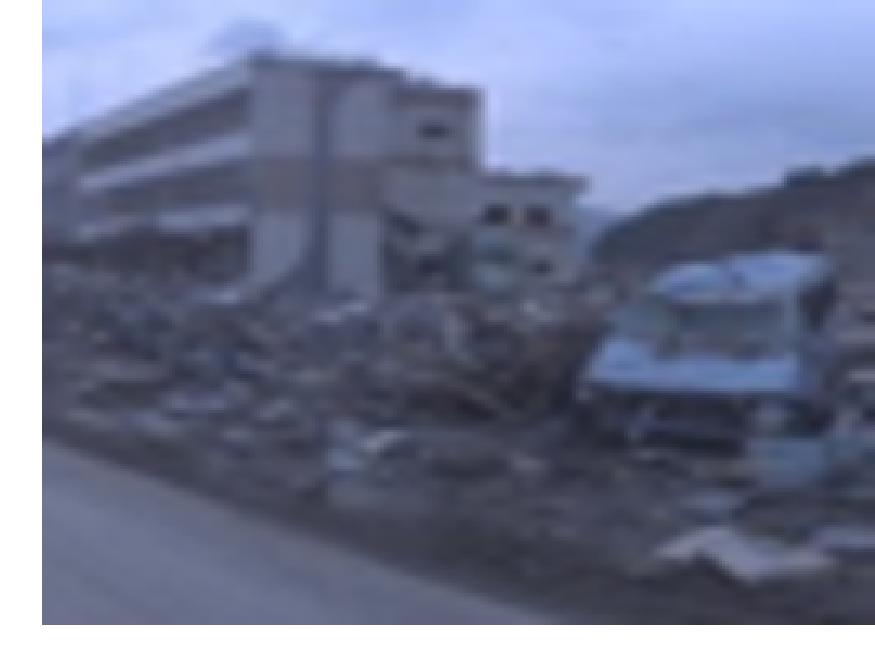}
\includegraphics[trim=16pt 8pt 0pt 0pt, clip=true,width=0.285\columnwidth]{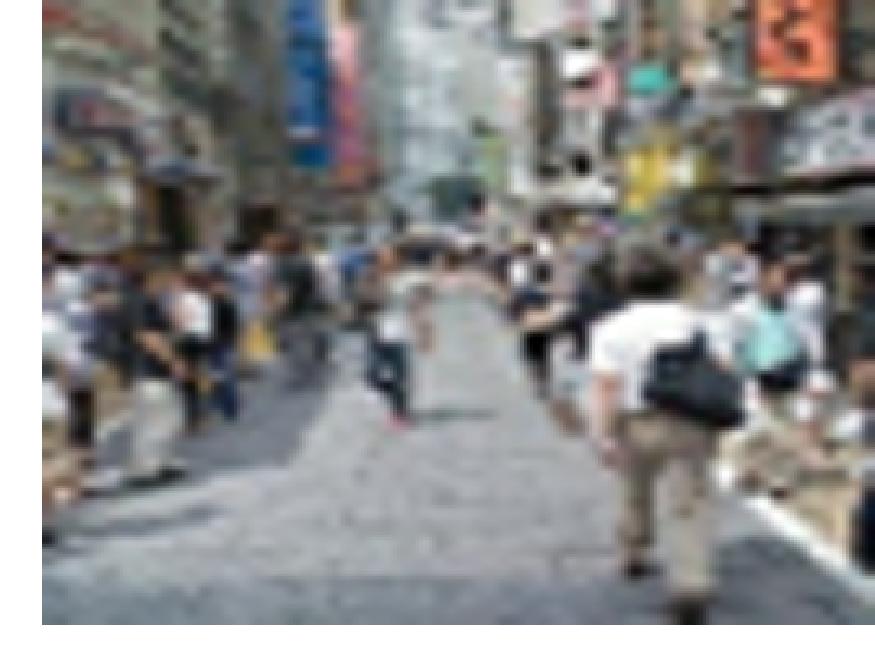}
}
\\
\scalebox{.75}{
\includegraphics[trim=16pt 8pt 0pt 0pt, clip=true, width=0.285\columnwidth]{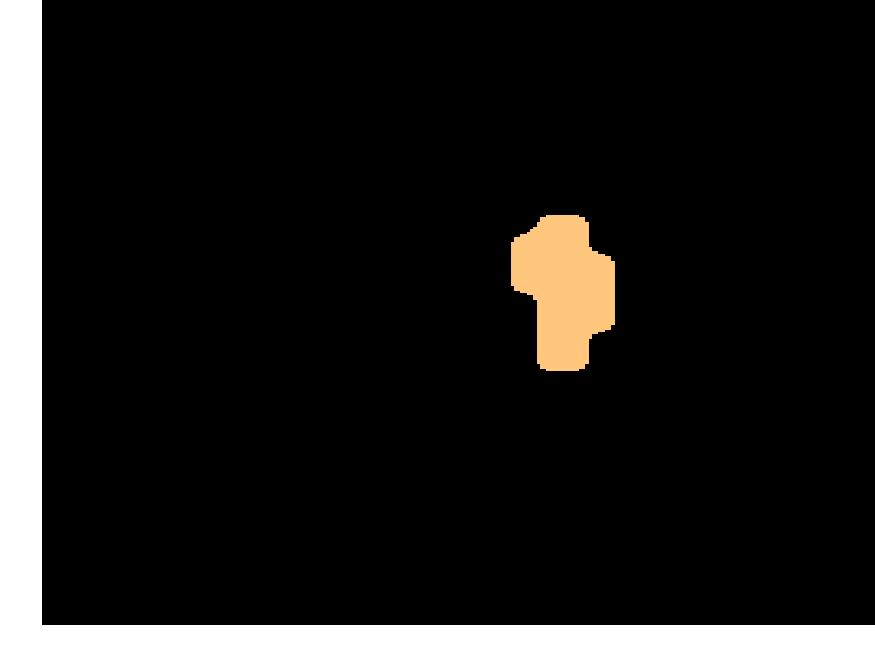}
\includegraphics[trim=16pt 8pt 0pt 0pt, clip=true, width=0.285\columnwidth]{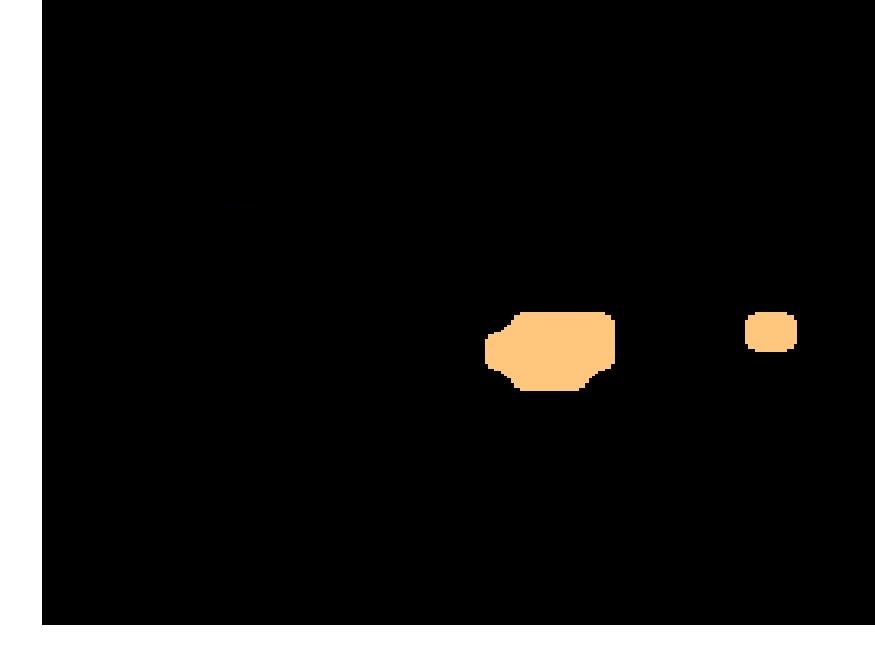}
\includegraphics[trim=16pt 8pt 0pt 0pt, clip=true, width=0.285\columnwidth]{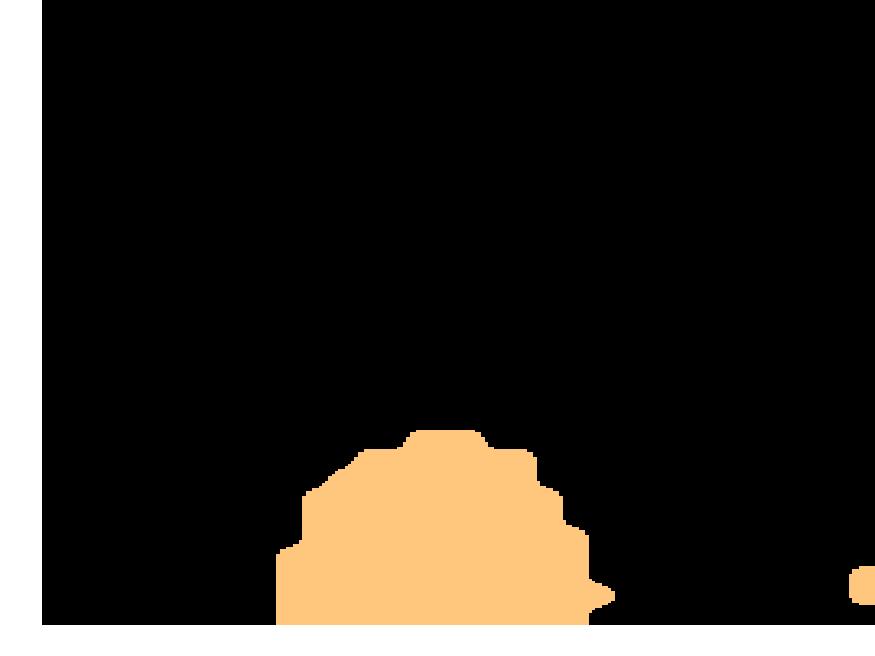}
\includegraphics[trim=16pt 8pt 0pt 0pt, clip=true, width=0.285\columnwidth]{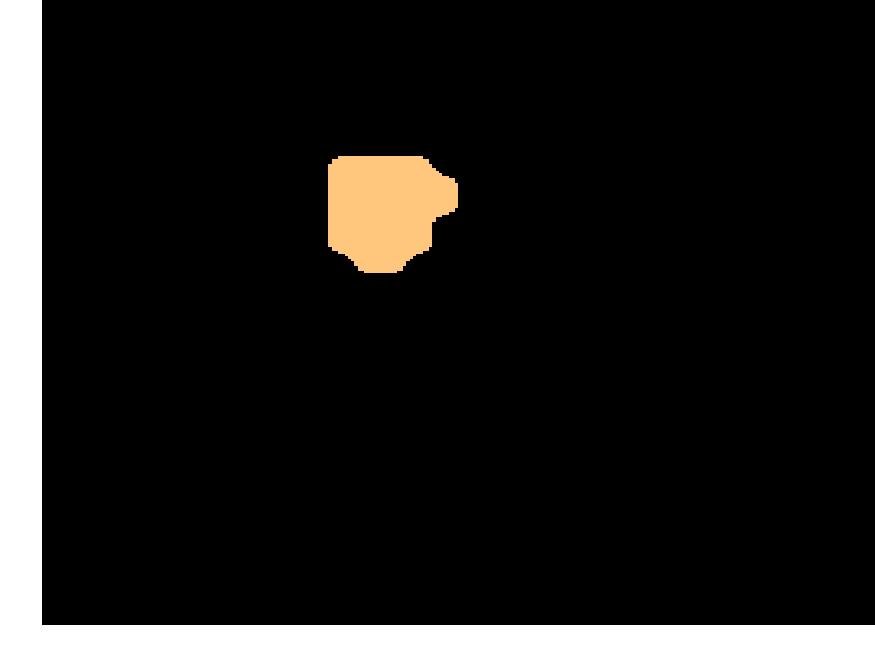}
\includegraphics[trim=16pt 8pt 0pt 0pt, clip=true, width=0.285\columnwidth]{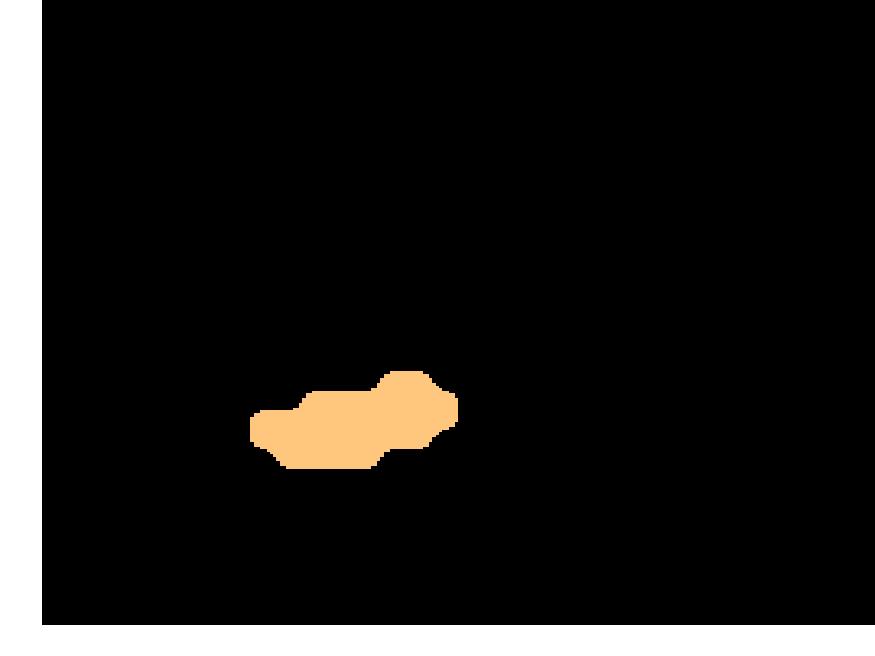}
\includegraphics[trim=16pt 8pt 0pt 0pt, clip=true, width=0.285\columnwidth]{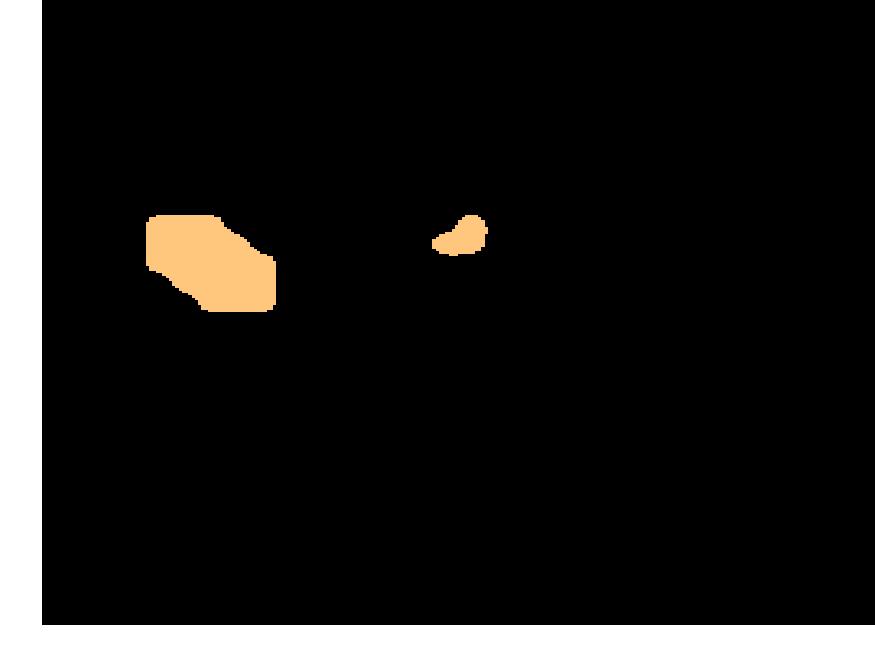}
\includegraphics[trim=16pt 8pt 0pt 0pt, clip=true, width=0.285\columnwidth]{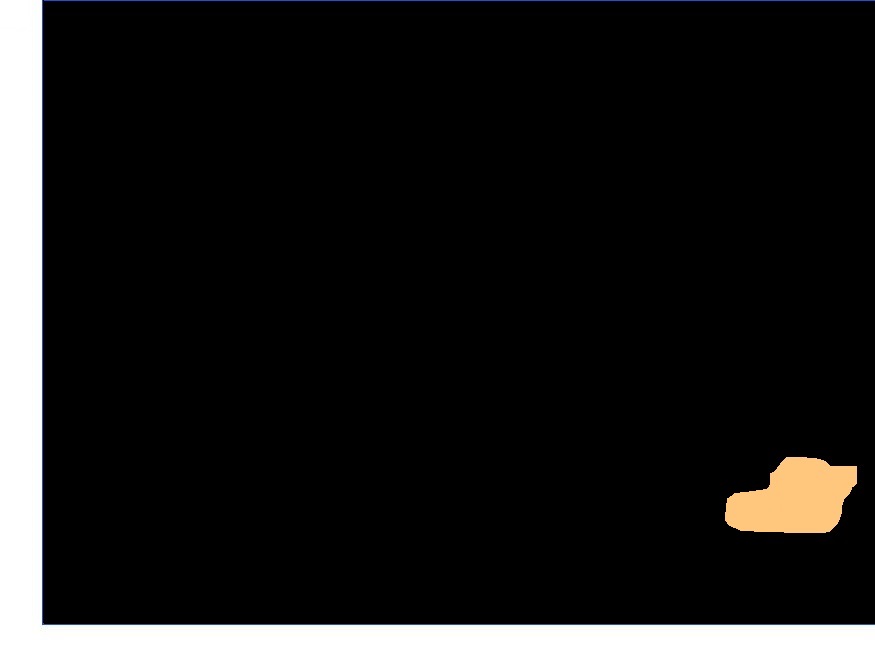}
\includegraphics[trim=16pt 8pt 0pt 0pt, clip=true,width=0.285\columnwidth]{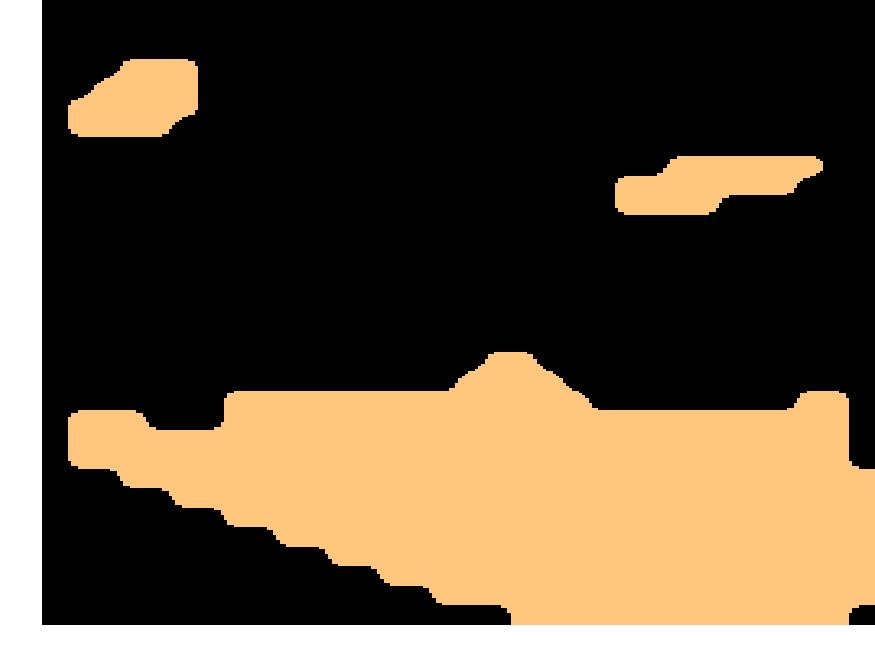}
\includegraphics[trim=16pt 8pt 0pt 0pt, clip=true,width=0.285\columnwidth]{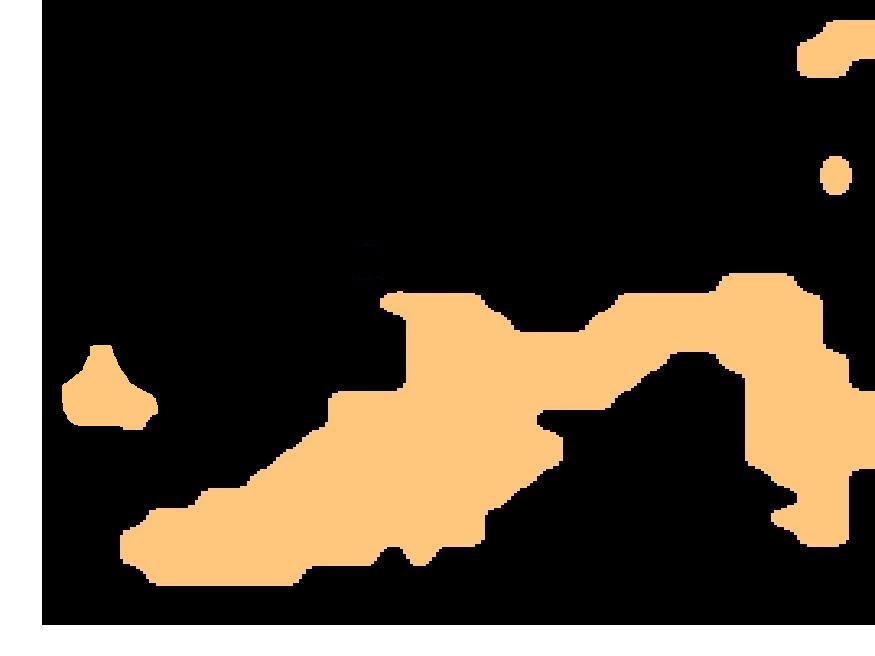}
}
\\
\scalebox{.75}{
\includegraphics[trim=16pt 8pt 0pt 0pt, clip=true, width=0.285\columnwidth]{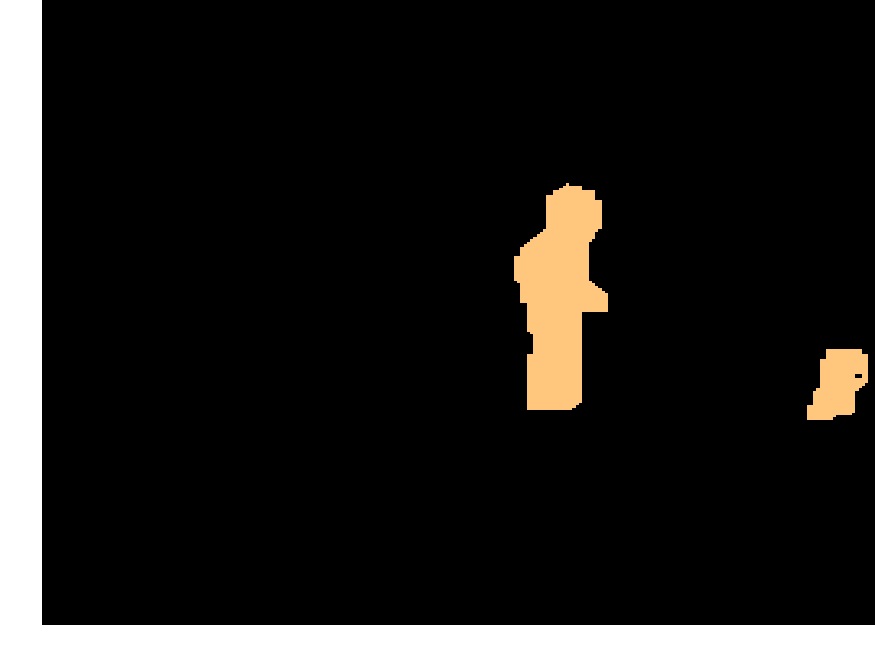}
\includegraphics[trim=16pt 8pt 0pt 0pt, clip=true, width=0.285\columnwidth]{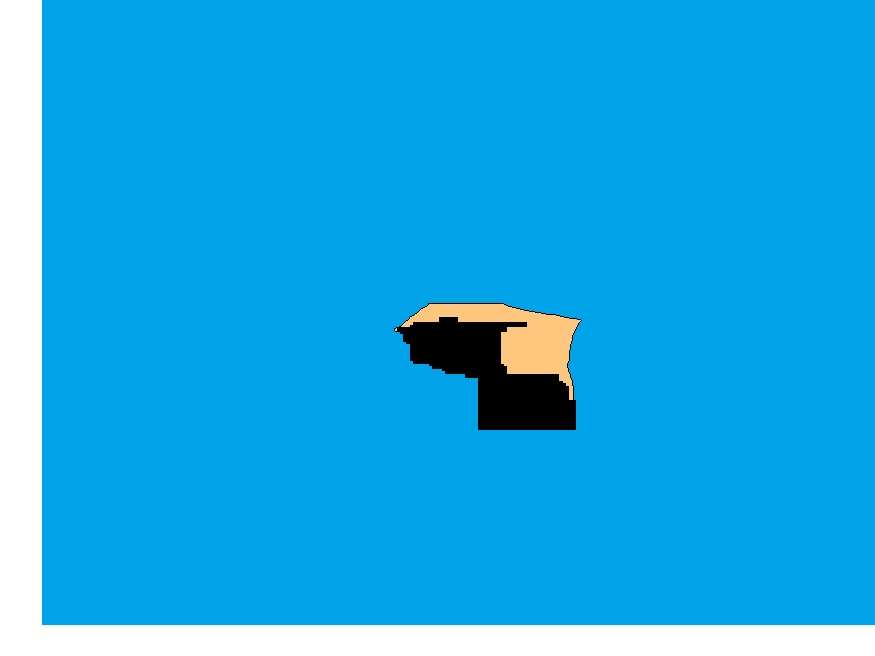}
\includegraphics[trim=16pt 8pt 0pt 0pt, clip=true, width=0.285\columnwidth]{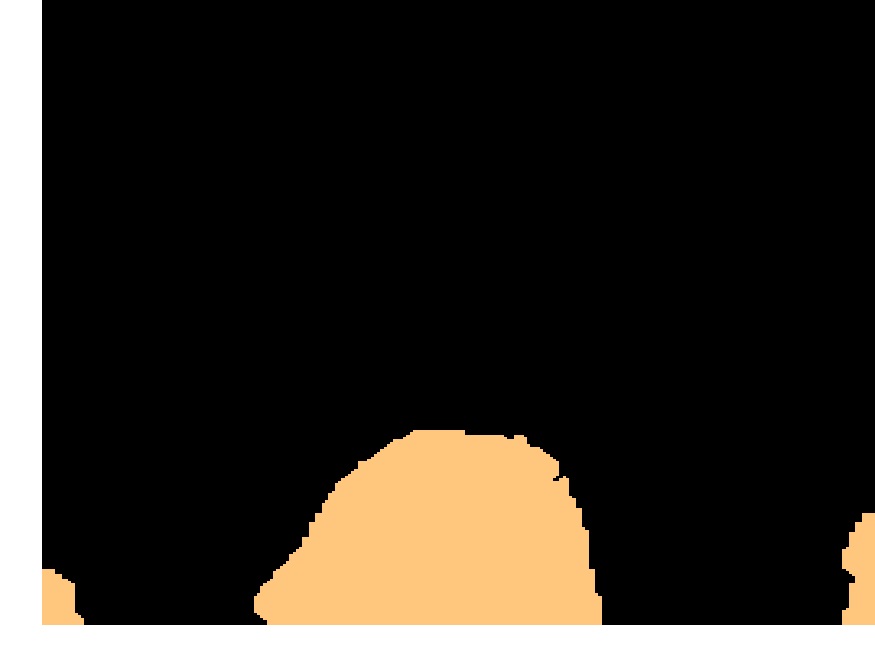}
\includegraphics[trim=16pt 8pt 0pt 0pt, clip=true, width=0.285\columnwidth]{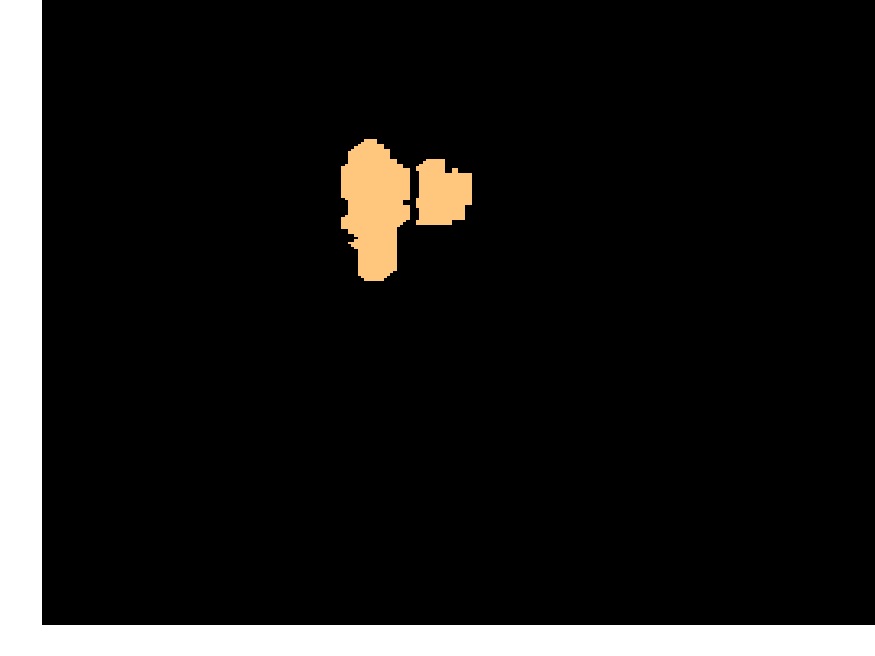}
\includegraphics[trim=16pt 8pt 0pt 0pt, clip=true, width=0.285\columnwidth]{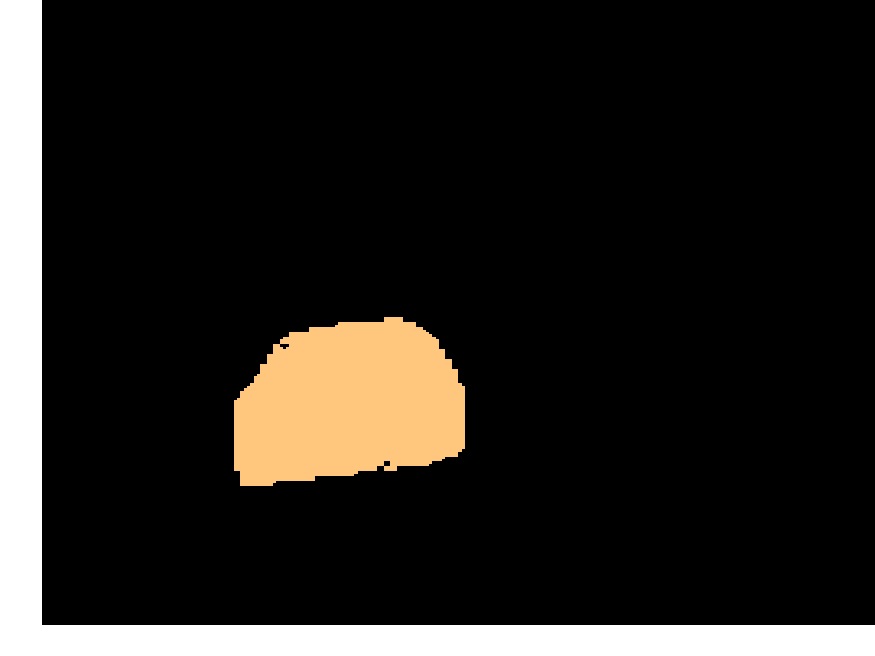}
\includegraphics[trim=16pt 8pt 0pt 0pt, clip=true, width=0.285\columnwidth]{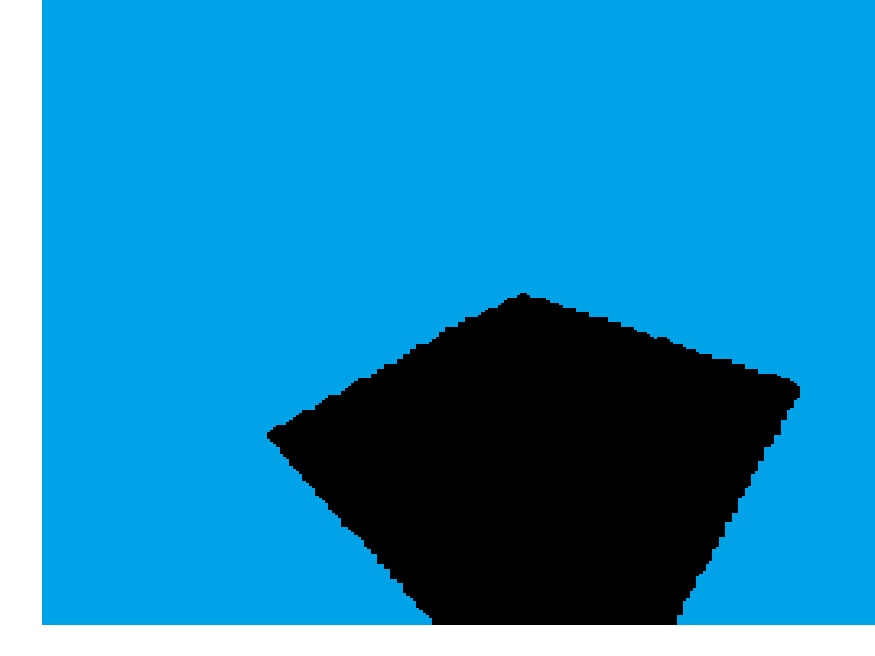}
\includegraphics[trim=16pt 8pt 0pt 0pt, clip=true, width=0.285\columnwidth]{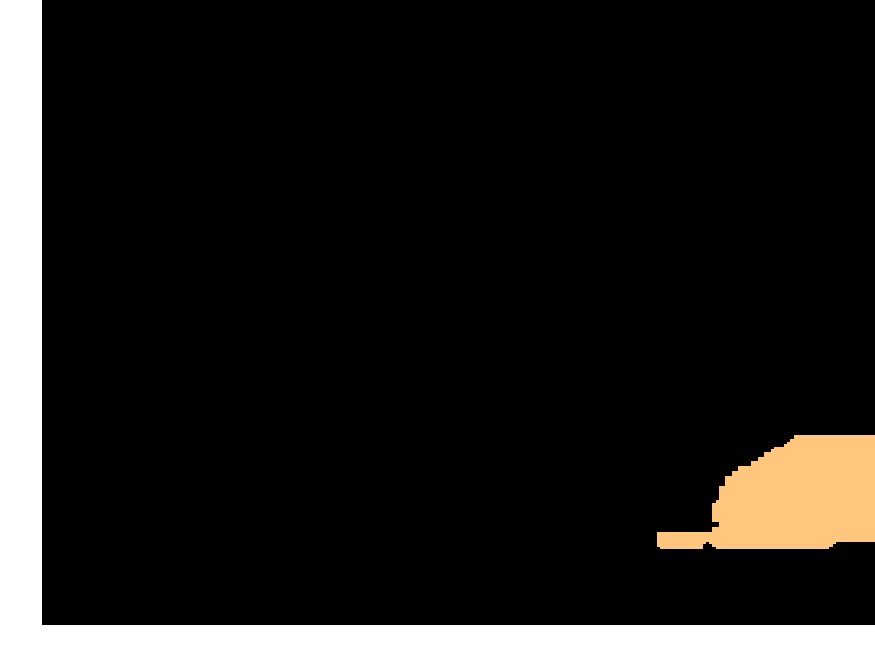}
\includegraphics[trim=16pt 8pt 0pt 0pt, clip=true,width=0.285\columnwidth]{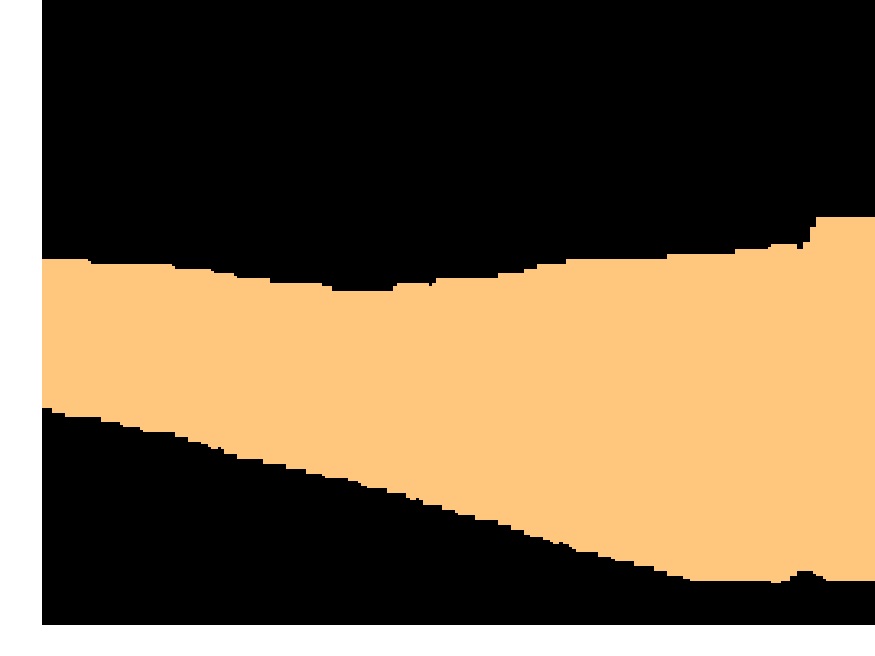}
\includegraphics[trim=16pt 8pt 0pt 0pt, clip=true,width=0.285\columnwidth]{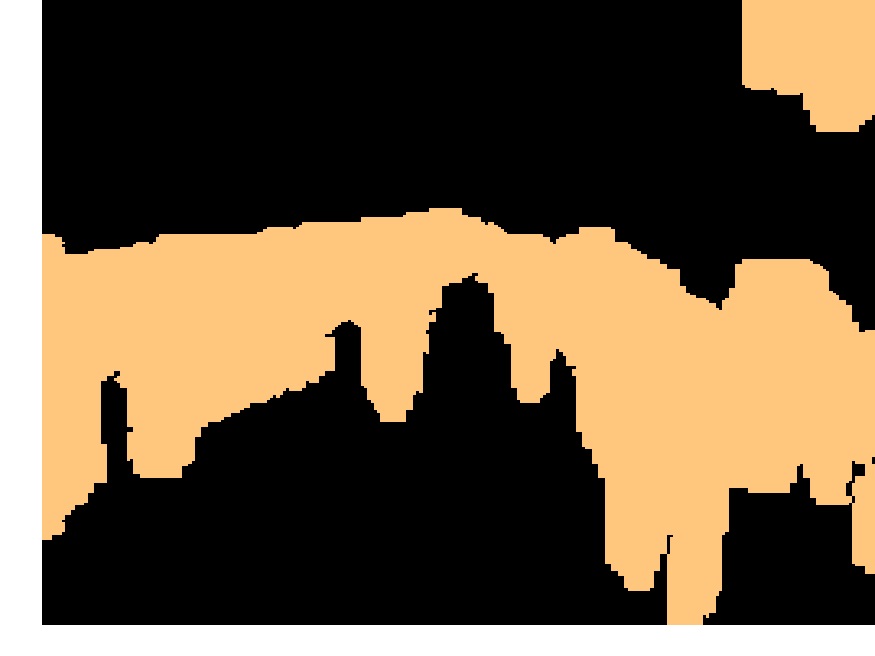}
}
\vspace{-0.5em}
\caption{\textbf{Qualitative results on the CDnet-2014, GASI-2015 and PCD-2015 datasets:} Rows $1-2$ and $5-6$ show image pairs, our results are shown in rows $3$ and $7$ and the ground-truths are shown in rows $4$ and $8$. No-change pixels are shown in \emph{black}, regions outside the ROIs are shown in \emph{sky-blue} while the changes are shown in \emph{coral-pink}.}
\label{fig:qualRes}
\vspace{-0.5em}
\end{figure*}

{The unary energies of our CRF model} are defined using the CNN activations, while the Gaussian edge potentials proposed by  \cite{krahenbuhl2011efficient} are used as pairwise terms.
Note that changes of interest can occur in any of the two paired images, and therefore it is not desirable to remain restricted to the detection of changes in only one of the images (w.r.t the other image).
For this purpose, the ground-truth with which we compare our final segmentation results include the changes in both images (see Fig.~\ref{fig:qualRes}).  
During the mean-field inference step, we find the segmentation map of both images using their respective edge potentials.
Subsequently, the two output maps are combined to get the final estimate of hidden variables. 
The resulting segmentation map is used as ground-truth during the CNN training (M step).

\vspace{-0.2em}
\subsection{Datasets and Protocols}\label{sec:dataNpro}
{We evaluate our method on the following four datasets. All of them include pixel level change ground truth, from which we derive the image-level annotations for weakly supervised learning. The pixel-level labels are not used in the training of our deep network.}

\vspace{-0.5em}
\paragraph{CDnet 2014 Dataset:}
The original video database consists of 53 videos with frame-by-frame ground-truths available for $\sim90,000$ frames in specified regions-of-interests (ROIs). 
Various types of changes (e.g., shadow, object motion and motion blur) under different conditions (e.g., challenging weather, air turbulence and dynamic background) are included in this database \cite{wang2014cdnet}. 
It is also important to note that the paired images are not registered and therefore background can change across paired images \cite{wang2014cdnet}. 
A total of $91,595$ distinct image pairs are generated at random from the video sequences. 
In each pair, both images belong to the same video but they are captured at different time instances. 


\vspace{-0.5em}
\paragraph{AICD 2012 Dataset:}
Aerial Image Change Detection (AICD) dataset \cite{bourdis2011constrained} consists of $1000$ pairs of large sized images ($800\times600$).  
It is a synthetic dataset in which the images are generated using a realistic rendering engine of a computer game (Battle Station 2). 
A total of $100$ scenes are included in this dataset containing several real-world objects including buildings, trees and vehicles. 
The scenes are generated under varying conditions with significant changes in viewpoint, shadows and types of changes.
Because the change regions are very small in satellite/aerial images, we work on the patch level and extract $48$ patches of size $122\times122$ from each image with minimal overlap. 
This provides a total of $24,000$ paired images, facilitating the training of a model with a large number of parameters.

\vspace{-0.5em}
\paragraph{GASI 2015 Dataset:}
Geoscience Australia Satellite Image (GASI) dataset is a custom built dataset based on the changes occurred during $1999-2015$ in a $\sim100\times100~km^2 $ area in the east of city of Melbourne in Victoria, Australia \cite{khan2017forest}. 
For each region of interest, we have a time lapse sequence (between 1999-2015) of surface reflectance data and the corresponding pixel quality maps. 
Due to the severe artifacts caused due to clouds and band saturation, the modelling of the  temporal trends is very challenging.
In contrast, the acquisition of paired images captured at different times is much easier. 

The annotations for two types of changes are provided in the GASI dataset namely: fire and harvests. We generate pairs of image patches for $67$ distinct regions of interest which were identified by experts. In total, $\sim 300$ pairs are generated for each region of interest which makes a total of $\sim20,000$ pairs. Since the raw data contains artifacts, we improved it's quality by filling data across different time instances.

There exists a large disparity among the sizes of change regions in the GASI dataset.
For very large sized regions, we cropped the region bounded by a tightly fit bounding box. 
For small regions (mostly changes due to forest harvesting) with area $<5\%$ of the total image area, we crop a bounding box with dimensions equal to three times that of a tightly bounded box. 
Since, there are large variations between the size of changes in the identified change regions, we converted all the regions to a uniform size of $224\times 224$ to get a consistent segmentation map.

\vspace{-0.5em}
\paragraph{PCD 2015 Dataset:}
Panoramic change detection dataset \cite{Sakurada2015change} consists of 200 pairs of panoramic images of street scenes and tsunami-hit areas.
The image size is $224\times 1024$, from which we extract $122\times 122$ patches with a minimal overlap for training and testing. 
This gives us a total of $3,600$ pairs ($18/$panoramic image).
It is important to mention that the two images are not perfectly registered. 
As a result, there are temporal differences in camera viewpoints, illumination and acquisition conditions.

\subsection{Results}\label{sec:results}

The change detection results of our approach on the four CD datasets are shown in Table~\ref{tab:det_res}. 
As a baseline, we only consider the classification branch of the CNN network initialized with the pre-trained VGG-net (configuration D, see $2-3$ columns of Table~\ref{tab:det_res}). 
Paired images are fed to this network architecture and $4096$ dimensional feature vectors are extracted from the FC2$'$ layer.
A linear SVM classifier is then trained for classification using the lib-linear package \cite{fan2008liblinear}. 
On both datasets, the average precision (AP) and the overall accuracy of our approach was significantly higher  than that of the baseline procedure (specifically 6.8\% and 5.3\% boost in AP for the CDnet and GASI datasets, respectively). 
{As a stronger baseline, we also report performance of the network when only the fine-tuned classification branch was used (columns $4-5$, Table~\ref{tab:det_res}). We note that our full model outperformed the results from the fine-tuned classification branch. }

\begin{table}
\centering
\scalebox{0.8}{
\begin{tabular}{l c c c c c c}
\toprule[1.3pt]
Dataset & \multicolumn{2}{c}{Classification}  & \multicolumn{2}{c}{Fine-tuned}  & \multicolumn{2}{c}{This Paper} \\
&   \multicolumn{2}{c}{Branch} & \multicolumn{2}{c}{Classification Branch}  & \\
\cmidrule{2-7}
 & AP & Acc. & AP & Acc.  & AP & Acc. \\
\midrule
CDnet-2014 & 88.8 & 92.0 & 94.0 & 96.6 & \textbf{95.6} & \textbf{98.7} \\
AICD-2012  & 92.7 & 95.4 & 95.7 & 98.0 & \textbf{97.3} & \textbf{99.1} \\
GASI-2015  & 80.3 & 82.5 & 83.0 & 84.4 & \textbf{85.6} & \textbf{86.5} \\
PCD-2015   & 67.1 & 73.5 & 72.9 & 81.1 & \textbf{74.9} & \textbf{84.2} \\
\bottomrule[1.3pt]
\end{tabular}}
\vspace{-2pt}
\caption{\textbf{Detection results} in terms of average precision (\%) and overall accuracy (\%) are listed above. Our approach clearly outperforms the baseline networks with only the classification branch.}
\label{tab:det_res}
\vspace{-0.5em}
\end{table}


\begin{figure*}
\centering
\begin{minipage}{0.7\textwidth}
\centering
\scalebox{0.82}{
\begin{tabular}{l c c c c c c c}
\toprule[1.3pt]
Dataset & \multicolumn{5}{c}{Baseline Approaches (mIOU-\%)}  & \multicolumn{1}{c}{This Paper} & \multicolumn{1}{c}{Fully-supervised}  \\
\cmidrule{2-6}
  & RS & DT & PN & Th. & GC\cite{boykov2001fast} & (mIOU-\%) & (mIOU-\%) \\
\midrule
CDnet-2014 & 16.4 & 36.8 & 37.4 & 35.9 & 37.1 & \textbf{46.2} &  59.2\\
AICD-2012 & 16.8 & 55.0 & 48.1  & 59.5 & 60.7 & \textbf{64.9} & 71.0 \\
GASI-2015 & 18.3 & 40.5 & 40.7 &  41.6 & 42.2 &  \textbf{55.3} & 62.4 \\
PCD-2015 & 16.5 & 41.7 & 39.3 & 35.9 & 39.5 & \textbf{47.7} & 58.8\\
\bottomrule[1.3pt]
\end{tabular}}
\vspace{-2pt}
\captionof{table}{\textbf{Segmentation Results} and comparisons with baseline methods. Note that all results (except the last column) are reported for the weakly-supervised setting. 
}
\label{tab:seg_res}
\end{minipage}
\hfill
\begin{minipage}{0.27\textwidth}
\centering
\includegraphics[width=0.8\columnwidth]{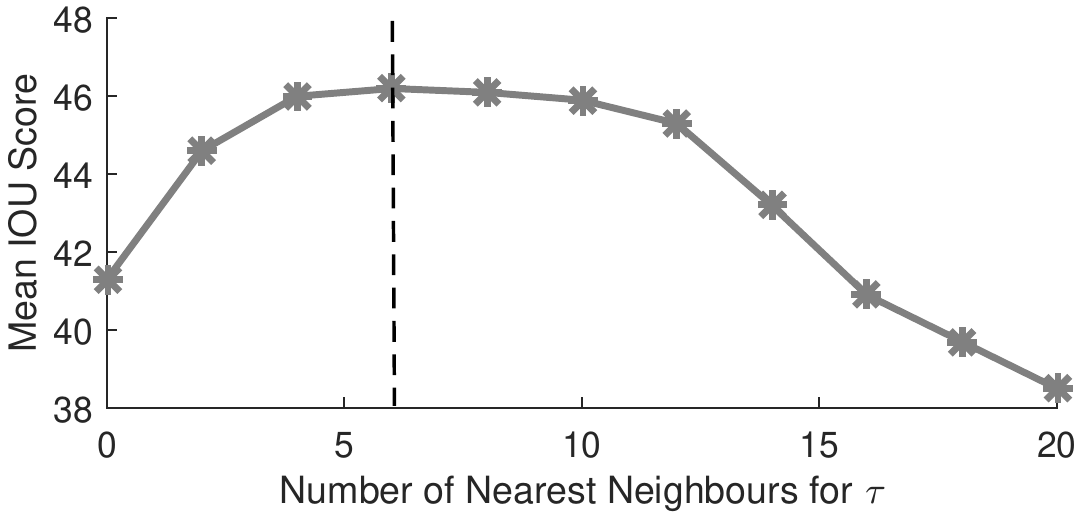}
\vspace{-0.5em}
\caption{\textbf{Sensitivity analysis} on the number of nearest neighbours used to estimate $\tau$ (CDnet-2014).}
\label{fig:sen_analysis}
\end{minipage}
\vspace{-0.5em}
\end{figure*}

We report the segmentation performance of our approach in Table~\ref{tab:seg_res} in terms of the mean intersection over union (mIOU) score.
To compare our change localization results, we report four baseline procedures. 
Specifically, we compare against random segmentation masks ($2^{nd}$ column-RS), thresholding applied to a difference map obtained from the pair of images ($3^{rd}$ column-DT), thresholding applied to the output from a pre-trained network (weights initialized for segmentation branch using VGG-net, $4^{th}$ column-PN), thresholding applied to the output from the fine-tuned network ($5^{th}$ column-Th.) and the graph-cuts inference \cite{boykov2001fast} using CNN outputs as unaries and contrast based pairwise potentials with a Potts model ($6^{th}$ column-GC). 
We note that random segmentation provides a lower baseline, while our results after training with ground-truths (shown in last column, Table~\ref{tab:seg_res}) sets an upper bar on the performance.
Another important trend is that the thresholding approach and graph-cuts performances do not differ by a large margin.
However, our weakly supervised approach was able to achieve significantly higher mIOU scores due to the additional potentials and constraints (Sec.~\ref{sec:model}). 

We also report segmentation results  on two additional baselines which use cardinality based pattern potentials (Table~\ref{tab:add_SegRes}). These baselines include the higher order potential (HOP) based dense and grid CRF models of Vineet $et~al.$ \cite{vineet2014filter} and Kohli $et~al.$ \cite{kohli2007p3} respectively. For both these baselines, we define HOPs on segments generated using mean-shift segmentation. Due to absence of pixel level supervision, we use the parameters from \cite{kohli2007p3}. We note that the dense CRF model with $P^{n}$ HOP \cite{vineet2014filter} performs better than the grid CRF model \cite{kohli2007p3}, however our deep structured prediction model outperforms both these strong baselines by a fair margin of $\sim 4-8\%$ in terms of mIOU score.

\begin{table}[t]
\centering
\scalebox{0.8}{
\begin{tabular}{l c c c}
\toprule[1pt]
Method & Dense CRF + $P^{n}$ HOP  & Grid CRF + $P^{n}$ HOP  & This Paper \\ 
 & \cite{vineet2014filter} & \cite{kohli2007p3} & \\
\midrule
mIOU\% &  42.0 &  38.3 & 46.2 \\
\bottomrule[1pt]
\end{tabular}}
\vspace{2pt}
\caption{Comparisons for \textbf{segmentation performance} with methods using cardinality potentials on the CDnet-2014.}
\label{tab:add_SegRes}
\end{table}

\begin{table}
\centering
\scalebox{0.9}{
\begin{tabular}{l c}
\toprule[1.3pt]
Method &  \multicolumn{1}{c}{Segmentation Results (mIOU)} \\
\midrule
with only segmentation branch &  40.7 \\
w/o CD fine-tuning &  37.4 \\
\cmidrule{2-2}
w/o difference term &  41.5 \\
w/o proportion constraint & 41.3 \\
\bottomrule[1.3pt]
\end{tabular}}
\vspace{-2pt}
\caption{\textbf{Ablative Analysis} on the CDnet-2014 Dataset. Change localization performance decreases without our full model. Both the CNN architecture and the proposed CRF model contribute towards the accurate change detection. }
\label{tab:ablative}
\vspace{-0.5em}
\end{table}

The qualitative results for change localization on the CDnet-2014, GASI-2015 and PCD-2015 datasets are shown in Fig.~\ref{fig:qualRes}.
The proposed approach performed well in localizing small as well as large sized changes (e.g., $1^{st}$ col, Fig.~\ref{fig:qualRes}).
Moreover, it showed good results for images acquired in varying conditions (e.g., night, snow, rainfall, dynamic background) and with different capturing devices (e.g., thermal camera, PTZ).
For the CDnet-2014 dataset, it is interesting to note that our method localized several changes in the regions outside the ROIs (shown in \emph{blue} color in the ground-truth).
Similarly, the qualitative results indicate the good performance of our method for satellite image based change detection. 

We performed an ablative study on the CDnet-2014 dataset for the change segmentation task (Table~\ref{tab:ablative}). The experimental results show that the localisation performance decreases without the feedback from the classification branch (whose predictions are more accurate). Moreover, since the pre-trained network is not trained to detect changes from multichannel inputs, the performance is considerably lower than that of the fine-tuned network. 
The difference term in the unary potential of the dense CRF and the global proportion constraint on the foreground probability mass also contributes a fair share in the final mIOU score. 

\begin{SCfigure}[][!th]
\centering
\parbox{0.65\linewidth}{
\includegraphics[trim=16pt 8pt 0pt 0pt, clip=true,width=0.15\columnwidth]{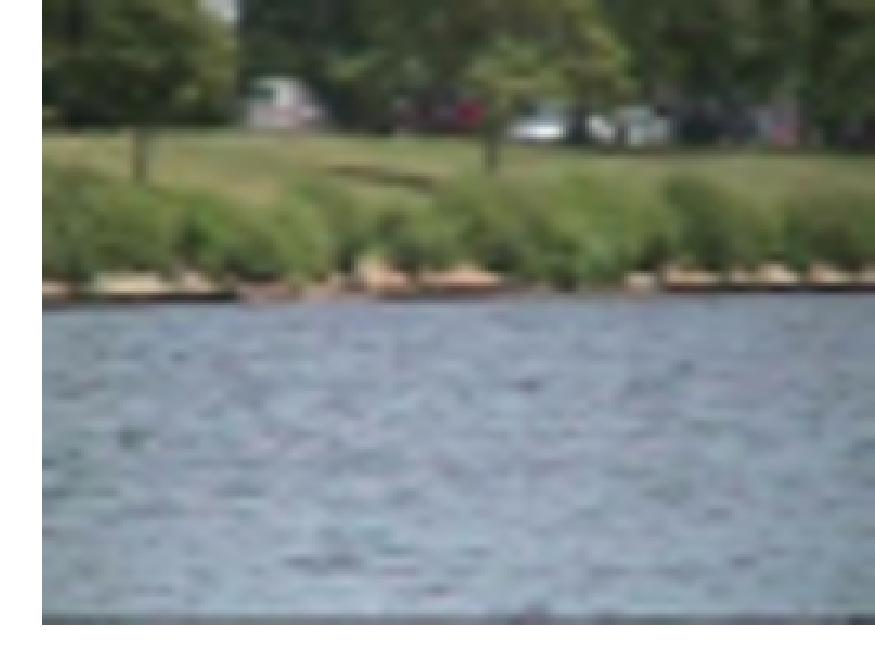}
\includegraphics[trim=16pt 8pt 0pt 0pt, clip=true,width=0.15\columnwidth]{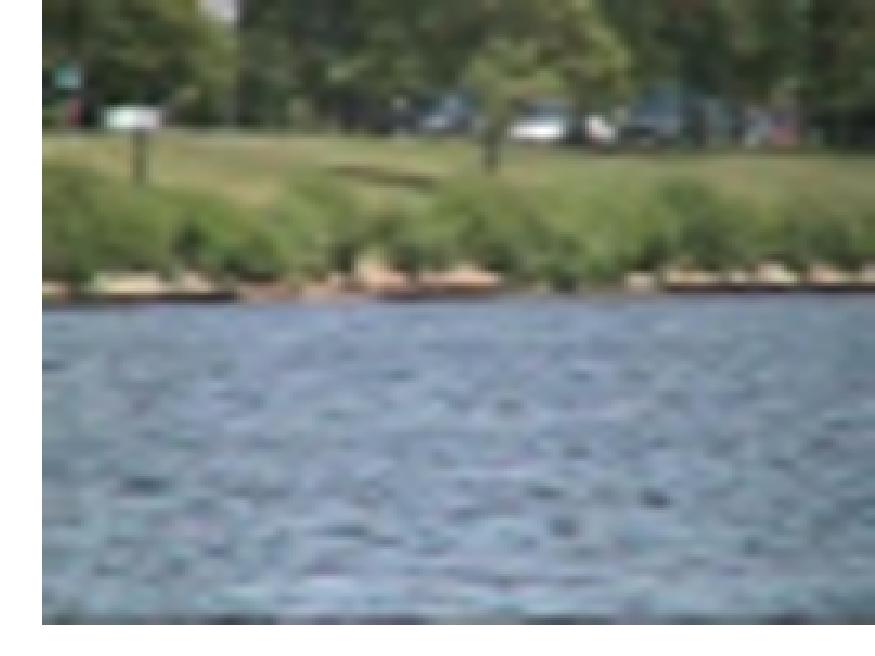}
\includegraphics[trim=16pt 8pt 0pt 0pt, clip=true,width=0.15\columnwidth]{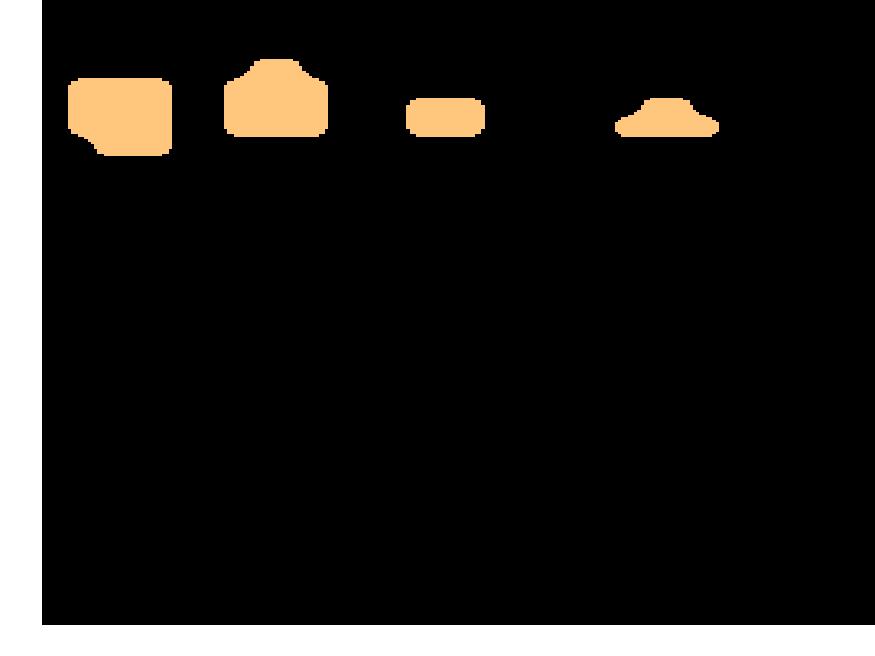}
\includegraphics[trim=16pt 8pt 0pt 0pt, clip=true,width=0.15\columnwidth]{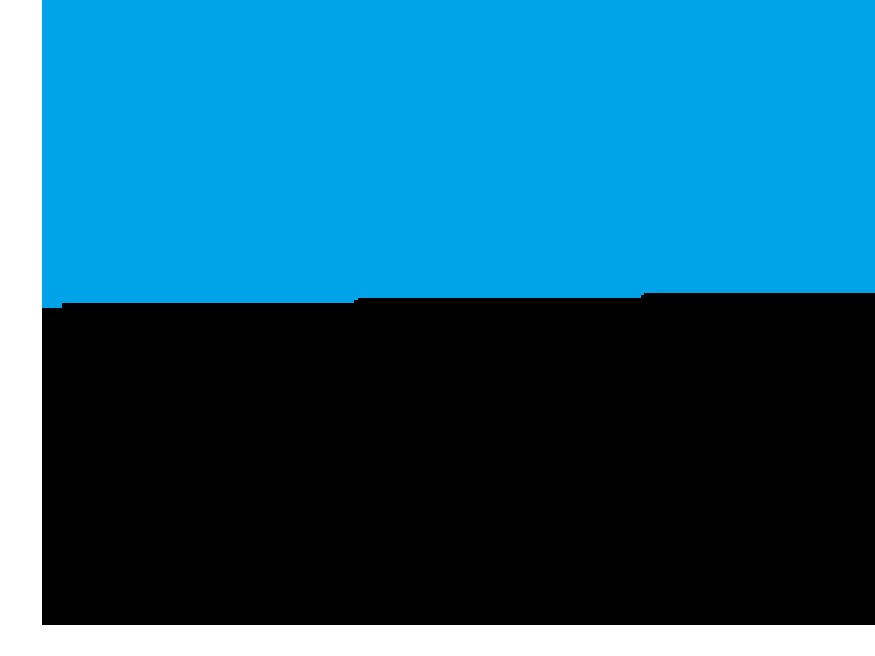}
\newline
\includegraphics[trim=16pt 8pt 0pt 0pt, clip=true,width=0.15\columnwidth]{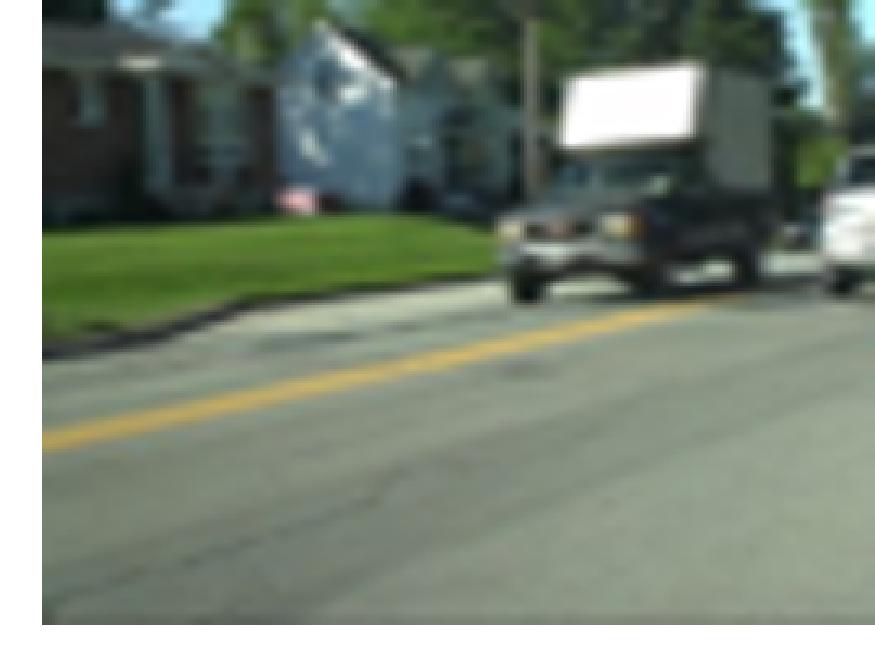}
\includegraphics[trim=16pt 8pt 0pt 0pt, clip=true,width=0.15\columnwidth]{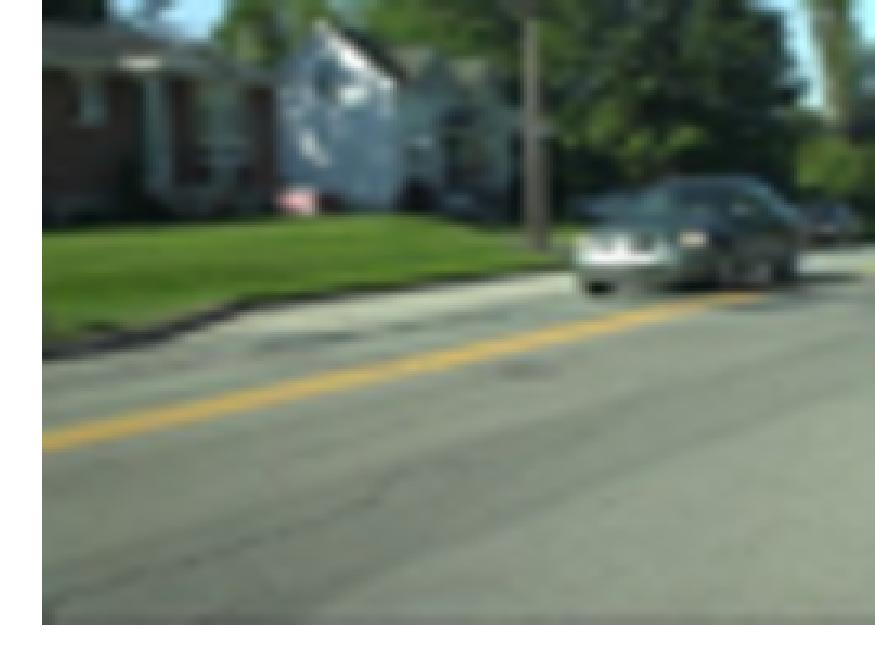}
\includegraphics[trim=16pt 8pt 0pt 0pt, clip=true,width=0.15\columnwidth]{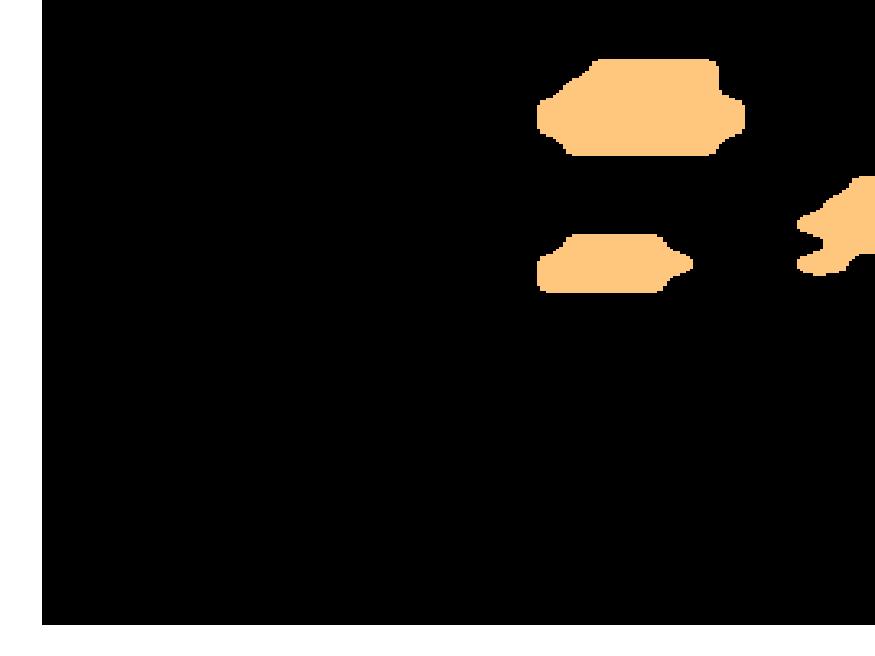}
\includegraphics[trim=16pt 8pt 0pt 0pt, clip=true,width=0.15\columnwidth]{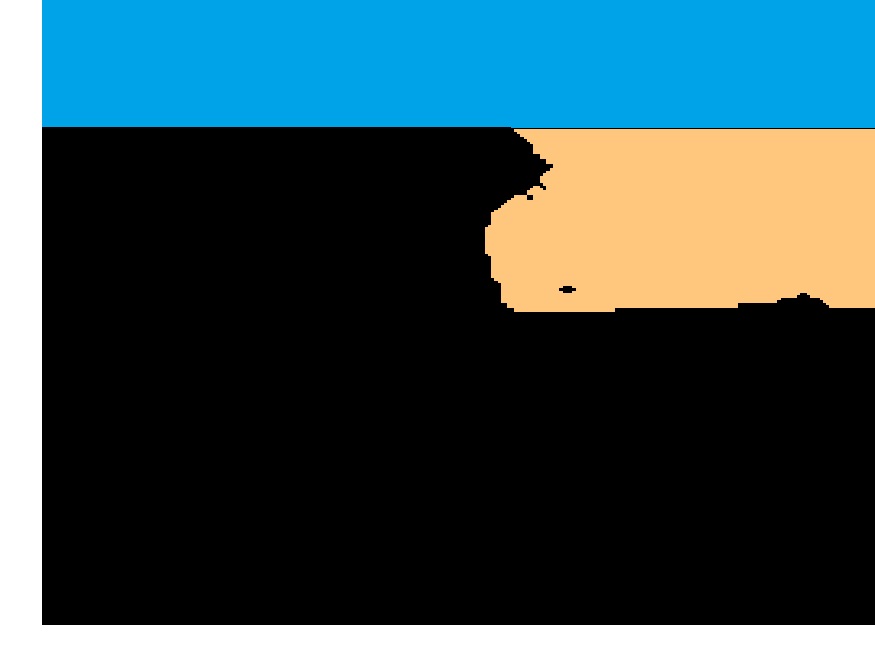}
\newline
\includegraphics[trim=16pt 8pt 0pt 0pt, clip=true,width=0.15\columnwidth]{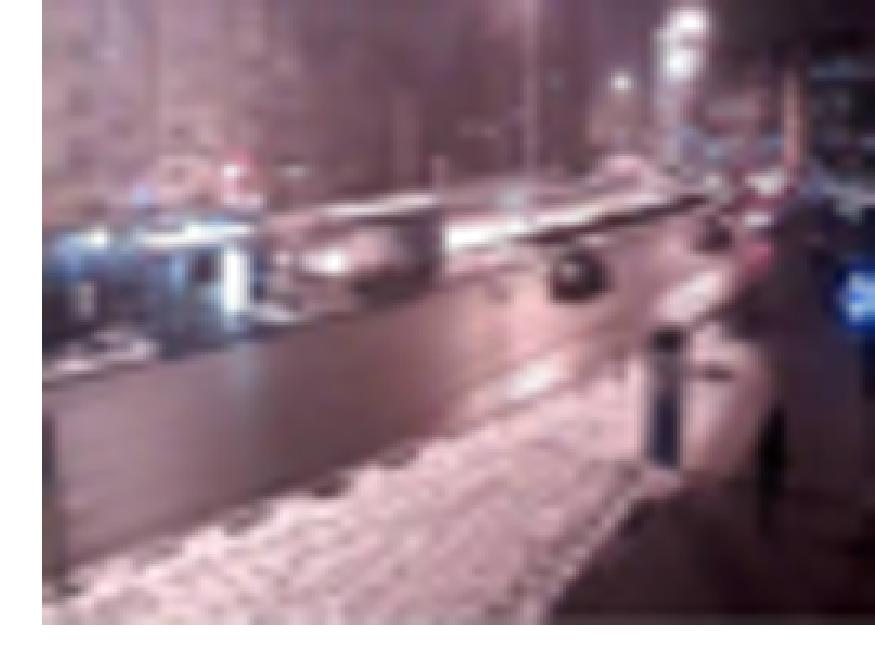}
\includegraphics[trim=16pt 8pt 0pt 0pt, clip=true,width=0.15\columnwidth]{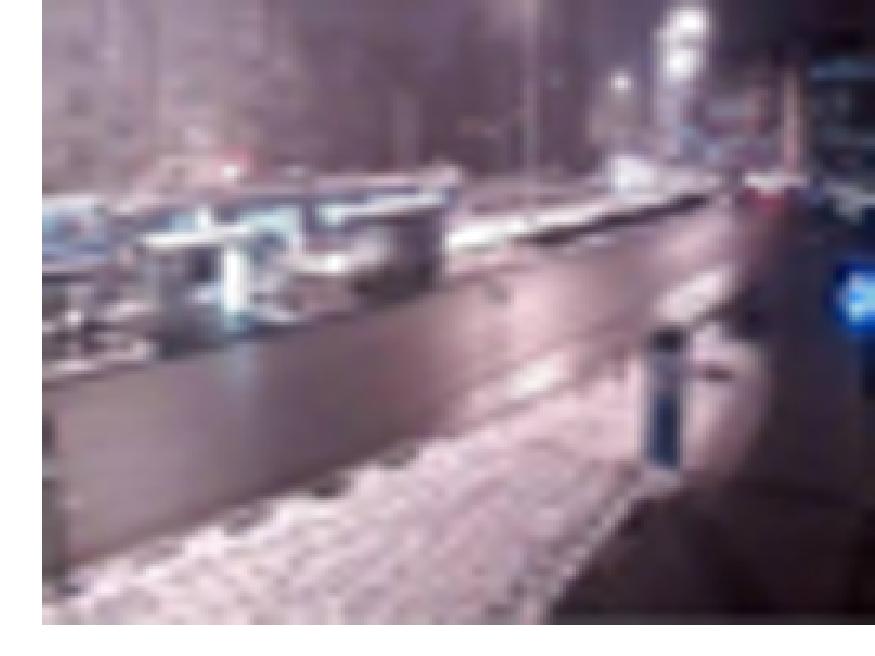}
\includegraphics[trim=16pt 8pt 0pt 0pt, clip=true,width=0.15\columnwidth]{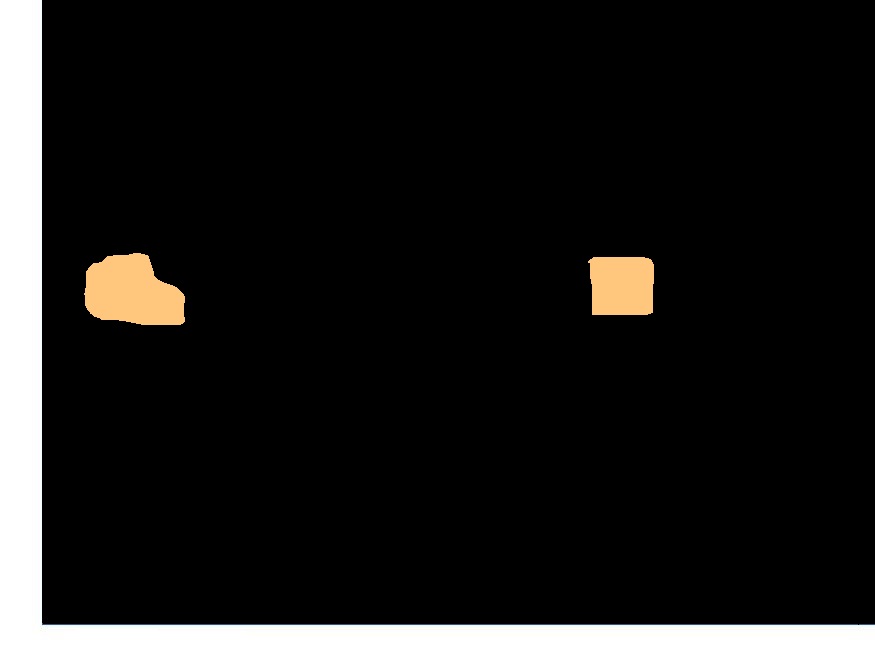}
\includegraphics[trim=16pt 8pt 0pt 0pt, clip=true,width=0.15\columnwidth]{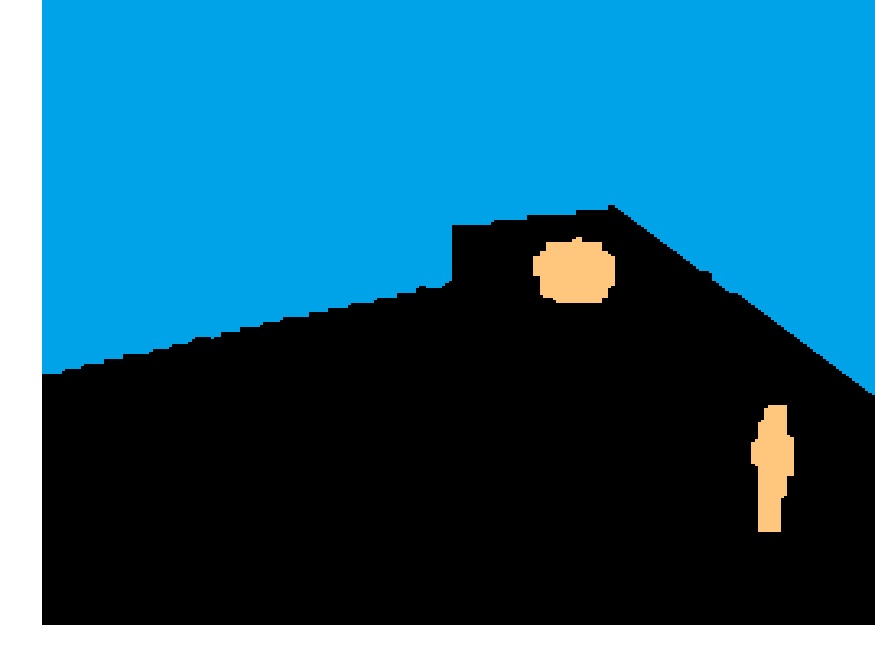}
}
\caption{\textbf{Error Cases:} We present example cases, for which the ground-truths didn't exactly match with the generated results. 
}
\label{fig:ambiguous_cases}
\end{SCfigure}

\begin{table}[t]
\centering
\scalebox{0.9}{
\begin{tabular}{l c c c c c  c }
\toprule[1pt]
 Normalized $\tau$  & 0.1 & 0.2 & 0.3  & 0.4 & 0.5 & 0.6 \\ 
\midrule
mIOU\% (CDnet-14) & 30.8 & 28.4 & 36.7 & 41.1 & 32.5 & 25.4  \\
\bottomrule[1pt]
\end{tabular}}
\vspace{-2pt}
\caption{\textbf{Segmentation performance} for different $\tau$ values.}
\label{tab:tau_trend}
\vspace{-0.5em}
\end{table}



{
During test, we use KNN to estimate an image-adaptive $\tau$, which gives better estimate of foreground proportion and performance. 
{We perform the KNN search using Euclidean distance on the features from the FC1 layer of the CNN model (classification branch)}. We use the fast approximate KNN search method based on KD-tree which has an average complexity of at most {O(log(n))}. Note that for the training, we validate a fixed-value $\tau$, which is faster than using KNN. As the pixel-level labels are unavailable in training, we set $\tau$ to a value which gives coverage of at least $15\%$ of each image on a validation set. 
To compare the performance of image-adaptive $\tau$ with a fixed-value $\tau$, we include the test segmentation scores  on the CDnet-14 dataset with different fixed values of $\tau$ in Table~\ref{tab:tau_trend}. 
Furthermore, we evaluate the segmentation performance with different numbers of nearest neighbours used to estimate $\tau$ and notice that the best performance is achieved when the $K$ is set to 6 (for KNN) to estimate the normalized foreground probability mass (Fig.~\ref{fig:sen_analysis}). Finally, we present some error cases of our approach in Fig.~\ref{fig:ambiguous_cases}.
}

\vspace{-0.3em}
\section{Conclusion}\label{sec:Conclusion}
This paper tackles the problem of weakly supervised change detection in paired images. We developed a novel CNN based model, which predicts change events and their location. Our approach defines a dense CRF model on top of the CNN activations and uses a modified mean-field inference procedure to enforce the compatibility between image and pixel level predictions.  The proposed algorithm achieved a significant boost both in the case of detection and localisation of change events compared to strong baseline procedures. Our work is the first effort in the area of weakly supervised change detection using paired images and will find possible applications in damage detection, structural monitoring and automatic 3D model updating systems. In future, we will explore the possibility of multi-class change detection in pair of images/videos. 

\bibliographystyle{named}
\bibliography{egbib}

\begin{thebibliography}{}

\bibitem[\protect\citeauthoryear{Adams \bgroup \em et al.\egroup
  }{2010}]{adams2010fast}
Andrew Adams, Jongmin Baek, and Myers~Abraham Davis.
\newblock Fast high-dimensional filtering using the permutohedral lattice.
\newblock In {\em Computer Graphics Forum}, volume~29, pages 753--762. Wiley
  Online Library, 2010.

\bibitem[\protect\citeauthoryear{Badrinarayanan \bgroup \em et al.\egroup
  }{2013}]{badrinarayanan2013semi}
Vijay Badrinarayanan, Ignas Budvytis, and Roberto Cipolla.
\newblock Semi-supervised video segmentation using tree structured graphical
  models.
\newblock {\em IEEE Transactions on Pattern Analysis and Machine Intelligence},
  35(11):2751--2764, 2013.

\bibitem[\protect\citeauthoryear{Bourdis \bgroup \em et al.\egroup
  }{2011}]{bourdis2011constrained}
Nicolas Bourdis, Denis Marraud, and Hichem Sahbi.
\newblock Constrained optical flow for aerial image change detection.
\newblock In {\em IGARSS}, pages 4176--4179. IEEE, 2011.

\bibitem[\protect\citeauthoryear{Boykov \bgroup \em et al.\egroup
  }{2001}]{boykov2001fast}
Yuri Boykov, Olga Veksler, and Ramin Zabih.
\newblock Fast approximate energy minimization via graph cuts.
\newblock {\em IEEE Transactions on Pattern Analysis and Machine Intelligence},
  23(11):1222--1239, 2001.

\bibitem[\protect\citeauthoryear{Fan \bgroup \em et al.\egroup
  }{2008}]{fan2008liblinear}
Rong-En Fan, Kai-Wei Chang, Cho-Jui Hsieh, Xiang-Rui Wang, and Chih-Jen Lin.
\newblock Liblinear: A library for large linear classification.
\newblock {\em The Journal of Machine Learning Research}, 9:1871--1874, 2008.

\bibitem[\protect\citeauthoryear{Girshick \bgroup \em et al.\egroup
  }{2014}]{girshick2014rich}
Ross Girshick, Jeff Donahue, Trevor Darrell, and Jagannath Malik.
\newblock Rich feature hierarchies for accurate object detection and semantic
  segmentation.
\newblock In {\em CVPR}, pages 580--587. IEEE, 2014.

\bibitem[\protect\citeauthoryear{Gueguen and Hamid}{2015}]{gueguen2015large}
Lionel Gueguen and Raffay Hamid.
\newblock Large-scale damage detection using satellite imagery.
\newblock {\em CVPR}, 2(2):3, 2015.

\bibitem[\protect\citeauthoryear{Jain and Grauman}{2013}]{jain2013predicting}
Suyog~Dutt Jain and Kristen Grauman.
\newblock Predicting sufficient annotation strength for interactive foreground
  segmentation.
\newblock In {\em ICCV}, pages 1313--1320. IEEE, 2013.

\bibitem[\protect\citeauthoryear{Khan \bgroup \em et al.\egroup
  }{2015}]{khan2015cost}
Salman~H Khan, Mohammed Bennamoun, Ferdous Sohel, and Roberto Togneri.
\newblock Cost sensitive learning of deep feature representations from
  imbalanced data.
\newblock {\em arXiv preprint arXiv:1508.03422}, 2015.

\bibitem[\protect\citeauthoryear{Khan \bgroup \em et al.\egroup
  }{2016}]{khan2016automatic}
Salman~H Khan, Mohammed Bennamoun, Ferdous Sohel, and Roberto Togneri.
\newblock Automatic shadow detection and removal from a single image.
\newblock {\em IEEE transactions on pattern analysis and machine intelligence},
  38(3):431--446, 2016.

\bibitem[\protect\citeauthoryear{Khan \bgroup \em et al.\egroup
  }{2017}]{khan2017forest}
Salman~H Khan, Xuming He, Fatih Porikli, and Mohammed Bennamoun.
\newblock Forest change detection in incomplete satellite images with deep
  neural networks.
\newblock {\em IEEE Transactions on Geosciences and Remote Sensing}, 2017.

\bibitem[\protect\citeauthoryear{Kohli \bgroup \em et al.\egroup
  }{2007}]{kohli2007p3}
Pushmeet Kohli, M~Pawan Kumar, and Philip~HS Torr.
\newblock P3 \& beyond: Solving energies with higher order cliques.
\newblock In {\em CVPR}, pages 1--8. IEEE, 2007.

\bibitem[\protect\citeauthoryear{Kr{\"a}henb{\"u}hl and
  Koltun}{2011}]{krahenbuhl2011efficient}
Philipp Kr{\"a}henb{\"u}hl and Vladlen Koltun.
\newblock Efficient inference in fully connected crfs with gaussian edge
  potentials.
\newblock In {\em NIPS}, pages 109--117, 2011.

\bibitem[\protect\citeauthoryear{Kr{\"a}henb{\"u}hl and
  Koltun}{2013}]{krahenbuhl2013parameter}
Philipp Kr{\"a}henb{\"u}hl and Vladlen Koltun.
\newblock Parameter learning and convergent inference for dense random fields.
\newblock In {\em ICML}, pages 513--521, 2013.

\bibitem[\protect\citeauthoryear{Lin \bgroup \em et al.\egroup
  }{2014}]{lin2014microsoft}
Tsung-Yi Lin, Michael Maire, Serge Belongie, James Hays, Pietro Perona, Deva
  Ramanan, Piotr Doll{\'a}r, and C~Lawrence Zitnick.
\newblock Microsoft coco: Common objects in context.
\newblock In {\em ECCV}, pages 740--755. Springer, 2014.

\bibitem[\protect\citeauthoryear{Long \bgroup \em et al.\egroup
  }{2015}]{long2015fully}
Jonathan Long, Evan Shelhamer, and Trevor Darrell.
\newblock Fully convolutional networks for semantic segmentation.
\newblock In {\em CVPR}, pages 3431--3440. IEEE, 2015.

\bibitem[\protect\citeauthoryear{Papandreou \bgroup \em et al.\egroup
  }{2015}]{papandreou15weak}
George Papandreou, Liang-Chieh Chen, Kevin~P Murphy, and Alan~L Yuille.
\newblock Weakly-and semi-supervised learning of a deep convolutional network
  for semantic image segmentation.
\newblock In {\em CVPR}, pages 1742--1750. IEEE, 2015.

\bibitem[\protect\citeauthoryear{Pinheiro and
  Collobert}{2015}]{pinheiro2015image}
Pedro~O Pinheiro and Ronan Collobert.
\newblock From image-level to pixel-level labeling with convolutional networks.
\newblock In {\em CVPR}, pages 1713--1721. IEEE, 2015.

\bibitem[\protect\citeauthoryear{Russakovsky \bgroup \em et al.\egroup
  }{2015}]{russakovsky2015s}
Olga Russakovsky, Amy~L Bearman, Vittorio Ferrari, and Fei-Fei Li.
\newblock What's the point: Semantic segmentation with point supervision.
\newblock {\em arXiv preprint arXiv:1506.02106}, 2015.

\bibitem[\protect\citeauthoryear{Sakurada and
  Okatani}{2015}]{Sakurada2015change}
Ken Sakurada and Takayuki Okatani.
\newblock Change detection from a street image pair using cnn features and
  superpixel segmentation.
\newblock In {\em BMVC}, 2015.

\bibitem[\protect\citeauthoryear{Simonyan and
  Zisserman}{2014}]{simonyan2014very}
Karen Simonyan and Andrew Zisserman.
\newblock Very deep convolutional networks for large-scale image recognition.
\newblock {\em arXiv preprint arXiv:1409.1556}, 2014.

\bibitem[\protect\citeauthoryear{Song \bgroup \em et al.\egroup
  }{2015}]{song2015sun}
Shuran Song, Samuel~P Lichtenberg, and Jianxiong Xiao.
\newblock Sun rgb-d: A rgb-d scene understanding benchmark suite.
\newblock In {\em CVPR}, pages 567--576, 2015.

\bibitem[\protect\citeauthoryear{Vineet \bgroup \em et al.\egroup
  }{2014}]{vineet2014filter}
Vibhav Vineet, Jonathan Warrell, and Philip~HS Torr.
\newblock Filter-based mean-field inference for random fields with higher-order
  terms and product label-spaces.
\newblock {\em International Journal of Computer Vision}, 110(3):290--307,
  2014.

\bibitem[\protect\citeauthoryear{Wang \bgroup \em et al.\egroup
  }{2014}]{wang2014cdnet}
Yi~Wang, Pierre-Marc Jodoin, Fatih Porikli, Janusz Konrad, Yannick Benezeth,
  and Prakash Ishwar.
\newblock Cdnet 2014: An expanded change detection benchmark dataset.
\newblock In {\em CVPR Workshops}, pages 393--400. IEEE, 2014.

\bibitem[\protect\citeauthoryear{Yuille and
  Rangarajan}{2003}]{yuille2003concave}
Alan~L Yuille and Anand Rangarajan.
\newblock The concave-convex procedure.
\newblock {\em Neural computation}, 15(4):915--936, 2003.

\bibitem[\protect\citeauthoryear{Zagoruyko and
  Komodakis}{2015}]{zagoruyko2015learning}
Sergey Zagoruyko and Nikos Komodakis.
\newblock Learning to compare image patches via convolutional neural networks.
\newblock {\em arXiv preprint arXiv:1504.03641}, 2015.

\end{thebibliography}

\end{document}